\documentclass[preprint,12pt,3p]{elsarticle}

\usepackage{graphicx}
\usepackage{caption}
\usepackage{subcaption}
\usepackage{psfrag}
\usepackage{array}
\usepackage{multirow}
\usepackage{comment}
\usepackage{amsmath}%
\newcommand{\norm}[1]{\left\lVert#1\right\rVert}
\DeclareMathOperator*{\argmin}{argmin}
\usepackage{xcolor}
\usepackage{amssymb}
\usepackage{amsfonts}%
\usepackage{amssymb}
\usepackage{algorithm}
\usepackage{algpseudocode}
\usepackage{setspace}
\usepackage{setspace}
\usepackage{textcomp}

\journal{Journal of Computational Physics}

\begin{document}
\sloppy
\begin{frontmatter}

\title{A deep-learning-based surrogate model for data assimilation in dynamic subsurface flow problems}

\author[label1,label5]{Meng Tang\corref{cor1}}
\address[label1]{367 Panama Street, Stanford, CA, 94305}
\address[label5]{Department of Energy Resources Engineering, Stanford University}

\cortext[cor1]{Corresponding author}

\ead{mengtang@stanford.edu}

\author[label1,label5]{Yimin Liu}

\ead{yiminliu@stanford.edu}

\author[label1,label5]{Louis J. Durlofsky}
\ead{lou@stanford.edu}

\begin{abstract}
A deep-learning-based surrogate model is developed and applied for predicting dynamic subsurface flow in channelized geological models. The surrogate model is based on deep convolutional and recurrent neural network architectures, specifically a residual U-Net and a convolutional long short term memory recurrent network. Training samples entail global pressure and saturation maps, at a series of time steps, generated by simulating oil-water flow in many (1500 in our case) realizations of a 2D channelized system. After training, the `recurrent R-U-Net' surrogate model is shown to be capable of accurately predicting dynamic pressure and saturation maps and well rates (e.g., time-varying oil and water rates at production wells) for new geological realizations. Assessments demonstrating high surrogate-model accuracy are presented for an individual geological realization and for an ensemble of 500 test geomodels. The surrogate model is then used for the challenging problem of data assimilation (history matching) in a channelized system. For this study, posterior reservoir models are generated using the randomized maximum likelihood method, with the permeability field represented using the recently developed CNN-PCA parameterization. The flow responses required during the data assimilation procedure are provided by the recurrent R-U-Net. The overall approach is shown to lead to substantial reduction in prediction uncertainty. High-fidelity numerical simulation results for the posterior geomodels (generated by the surrogate-based data assimilation procedure) are shown to be in essential agreement with the recurrent R-U-Net predictions. The accuracy and dramatic speedup provided by the surrogate model suggest that it may eventually enable the application of more formal posterior sampling methods in realistic problems.

\end{abstract}

\end{frontmatter}
\section{Introduction}
\label{sect:intro}

Reliable subsurface flow forecasts are essential for the effective management of oil, gas and groundwater resources. The intrinsic uncertainty in subsurface characterizations can, however, lead to substantial uncertainty in subsurface flow predictions. Inverse modeling, also referred to in this context as data assimilation or history matching, entails the calibration of geological models based on observed data of various types. The resulting posterior (history matched) models generally provide predictions with narrower uncertainty ranges, and are thus more useful for reservoir/aquifer management. Inverse modeling algorithms can, however, be very computationally intensive, particularly when the forward model entails complex physical processes and contains a large number of grid blocks. For such cases it would be very beneficial to have access to an accurate surrogate model that can be used in place of the original model for the majority of the required function evaluations.
 
In this paper, we introduce a new deep-learning-based surrogate model that can accurately capture the evolution of high-dimensional (global) pressure and saturation fields, along with well phase flow rate data, given the subsurface geological characterization. The surrogate model employs convolutional neural networks (CNNs) to capture the nonlinear relationship between the geological parameter map (permeability in our case) and subsurface flow state maps. A type of recurrent neural network (RNN) is applied to capture the temporal evolution of the system. Given a sufficient number of training samples (1500 in our case), this surrogate model can provide flow predictions in close agreement with the underlying flow simulator, but with a significant reduction in computational cost. Thus this approach enables the application of accurate but computationally demanding inverse modeling procedures.  
 
There has been extensive research on constructing surrogate models for subsurface flow prediction. These can be generally classified, based on the mathematical formulation, into physics-based and data-driven procedures (though these categories are not mutually exclusive). The physics-based methods typically neglect or simplify physical or numerical aspects of the problem, through, for example, reduced-physics modeling, coarse-grid modeling, or proper orthogonal decomposition (POD) based reduced-order modeling (ROM). A variety of POD-based ROMs, in which the state variables and the system of equations are projected into low-dimensional space and then solved, have been applied for a range of subsurface flow problems  \citep{van2006reduced, cardoso2009development, he2014reduced, yang2016fast, jin2018reduced}. These ROMs can be effective, although they are generally only accurate when new (test) runs are sufficiently `close' to training runs. In addition, the application of POD-based ROMs for inverse modeling has been somewhat limited, though a few studies have shown promise in this area \citep{he2013reduced,xiao2018non}.








Data-driven approaches, on the other hand, rely purely on simulation data to train a statistical model to approximate the input-output relationship of interest. Along these lines, Hamdi et al.~\citep{hamdi2017gaussian} applied a Gaussian process for the surrogate modeling of a 20-parameter unconventional-gas reservoir system in the context of history matching. A polynomial chaos expansion surrogate model was constructed by Bazargan et al.~\citep{bazargan2015surrogate} for inverse modeling with a 40-parameter representation of a 2D fluvial channelized reservoir undergoing waterflooding. Artificial neural networks (ANNs) were applied by Costa et al.~\citep{costa2014application} to build a 16-parameter surrogate model to assist in the history matching of an oil-water system. Although these data-driven surrogates each have their own advantages and drawbacks, they share the limitation of applicability only for relatively low-dimensional problems.
 
Recent advances in deep neural networks, and their successful application for high-dimensional data regression in image recognition \citep{baltrusaitis2013constrained, liu2015deep, isola2017image} and natural language processing \citep{wu2016google, devlin2018bert}, have stimulated research on deep-learning-based surrogate modeling for high-dimensional nonlinear systems.  In contrast to shallow ANNs, carefully designed deep neural networks can capture complex high-dimensional nonlinearities, with relatively limited training data, while avoiding overfitting \citep{dziugaite2017computing, arora2018stronger}. Within a subsurface flow setting, Zhu and Zabaras~\citep{zhu2018bayesian} first introduced a fully convolutional encoder-decoder network to approximate flow quantities. They considered  single-phase steady-state flow in models characterized by Gaussian permeability fields and demonstrated that their deep convolutional neural network, trained with a limited amount of data, was able to predict high-dimensional pressure maps. Subsequent applications, again with Gaussian permeability fields, have included the prediction of CO$_2$ saturation plumes~\citep{mo2018deepCO2} and groundwater contaminant concentration~\citep{mo2018deep} in the context of uncertainty quantification and inverse modeling, respectively. Jin et al.~\citep{jin2019deep} presented a deep-learning-based embed-to-control framework. This surrogate model was shown to provide very fast predictions for well responses and dynamic reservoir states for two-phase systems under varying well controls. These studies demonstrate the ability of deep convolutional neural networks to capture high-dimensional relationships in subsurface flow systems. 

Our interest here, however, is in the use of surrogate modeling for nonlinear oil-water systems in formations characterized by channelized (non-Gaussian) permeability fields. Our numerical experiments indicate that, for such problems, existing deep-CNN-based surrogate models may not provide the level of accuracy required. Specifically, the autoregressive strategy proposed in~\citep{mo2018deep} can lead to error accumulation in time, while the strategy to encode time as an additional input channel in~\citep{mo2018deepCO2} still treats the dynamic system (essentially) as a steady problem. In addition, the temporal evolution between time-dependent state maps was not considered. Another limitation, for our application, is that the wells in these studies were all specified to operate under rate control (fixed injection and production rates). Here we intend to operate wells under pressure control, in which case well rates, and thus the time evolution of the saturation field, can vary significantly from realization to realization. These variations are much larger under pressure control than rate control, since with rate control the amount of injected fluid, at a given time, is the same in all training and test runs.

In order to facilitate fast inverse modeling for oil-water flow in channelized systems, with multiple wells operating under pressure control, we introduce a new deep-learning-based surrogate model. This model is trained by simulating flow through a number of different geological realizations (drawn from a single geological scenario). The surrogate model then provides very fast predictions of pressure and saturation and well flow rate data, which can be used in the inverse modeling procedure. The deep-learning-based surrogate model developed here uses a convolutional U-Net~\citep{ronneberger2015u} architecture to approximate the state responses from the (input) permeability field. Above this U-Net, a recurrent architecture, specifically long short term memory~(LSTM)~\citep{hochreiter1997long,xingjian2015convolutional}, is incorporated to capture the time-dependent evolution of the global pressure and saturation state maps. 

 
This paper proceeds as follows. In Section~\ref{sect:method}, we provide the underlying flow equations and then describe the surrogate model, in which a residual U-Net and a recurrent architecture LSTM are combined to capture both spatial and temporal information. In Section~\ref{sect:flowstats}, the surrogate model is applied for oil-water flow involving multiple realizations of a channelized system, with flow driven by 25 wells under pressure control. A detailed assessment of model accuracy, in terms of global states and well-rate quantities, is presented. Then, in Section~\ref{sect:hm}, we apply the surrogate model to history match a channelized geomodel. A randomized maximum likelihood framework is used to generate multiple posterior realizations, and posterior (surrogate-based) flow predictions are verified through comparison to numerical flow simulations. In Section~\ref{sect:concl}, we summarize this work and provide suggestions for future investigations. In the Appendix we provide architecture details for the deep-learning-based surrogate model used in this work.


\section{Methodology}
\label{sect:method}

In this section, we present the governing flow equations and then describe our  deep-learning-based surrogate model for dynamic two-phase subsurface flow problems. The key aspects of the surrogate model, including model architecture, the training process, and data pre-processing, are discussed.

\subsection{Governing equations for two-phase flow}
\label{sect:goveqns}

In this work, we consider 2D immiscible oil-water flow problems. Combining mass conservation and Darcy's law, which relates Darcy velocity to pressure gradient and other quantities, we arrive at:
\begin{equation} 
    \nabla \cdot (\rho_j \lambda_j \mathbf{k}\nabla p_j) + q_j^w = \frac{\partial}{\partial{t}}(\phi \rho_j S_j), \ \ \  j=o,~w.
\label{eq:governing-eq}
\end{equation}
Here $j$ denotes phase/component, with $j=o$ for oil and $w$ for water, $\rho_j$ is phase density, $\lambda_j = \frac{k_{rj}}{\mu_j}$ is the phase mobility (here $k_{rj}(S_j)$ is the relative permeability, a prescribed function of phase saturation $S_j$ that is usually derived from laboratory measurements, and $\mu_j$ is phase viscosity), $\mathbf{k}$ is the absolute permeability tensor, $p_j$ is the phase pressure, $q_j^w$ is the source/sink term (superscript $w$ indicates well), $t$ is time, and $\phi$ is the rock porosity. The governing equations are completed by noting that the phase saturations sum to unity, and that the phase pressures are related through the prescribed capillary pressure $P_c$; i.e., $p_o - p_w = P_c (S_w)$. In this work we neglect capillary pressure effects (as is common in large-scale reservoir simulation), so $p_o=p_w=p$. Note also that Eq.~\ref{eq:governing-eq} is written for horizontal ($x-y$) systems, so gravitational effects do not appear.

Eq.~\ref{eq:governing-eq} is discretized using a fully implicit finite volume method, as is standard in oil reservoir simulation. The primary variables are $S_w$ and $p$. Fluid is introduced and removed from the system via wells, and the source term, for a well in grid block $i$, is modeled using the Peaceman representation~\citep{peaceman1983interpretation}:
%
\begin{equation}
\label{eq:well_flow}
    \left(q_j^w\right)_i = WI_i  (\lambda_j  \rho_j)_i (p_i - p^w).
\end{equation}
Here $p_i$ and $p^w$ denote the well block and wellbore pressure, respectively, and $WI_i$ denotes the well index, given by
\begin{equation}
\label{eq:wi}
    WI_i = \frac{2\pi k_i \Delta z}{\ln(r_0 / r_w)},
\end{equation}
where $k_i$ is the (isotropic) permeability in grid block $i$, $\Delta z$ is the grid block thickness, $r_w$ is the wellbore radius, and $r_0 = 0.14\sqrt{(\Delta x)^2 + (\Delta y)^2}$, where $\Delta x$ and $\Delta y$ are grid block dimensions in the $x$ and $y$ directions. Note that Eq.~\ref{eq:wi} applies for a fully penetrating vertical well, centered in the grid block, and isotropic permeability $k_i$. Analogous expressions have been developed for a wide range of more general cases. We see from Eqs.~\ref{eq:well_flow} and \ref{eq:wi} that, when wellbore pressure $p^w$ is specified (as it is in our case), well injection and phase production rates depend strongly on the well block states $p$ and $S_w$. Thus the key well-rate quantities are not specified but must be computed from the dynamic solution. 


\subsection{Data-driven surrogate modeling in reservoir simulation}

In applications such as inverse modeling and uncertainty quantification, we need to solve the discretized versions of Eq.~\ref{eq:governing-eq} hundreds or thousands of times, for different reservoir models, but under identical initial and boundary conditions. A single simulation run can be expressed as
\begin{align}
    \mathbf{x} = f(\mathbf{m}, \mathbf{u}),
    \label{eq-f}
\end{align}
where $f$ indicates the reservoir simulation process, $\mathbf{m} \in \mathbb{R}^{n_b}$ denotes the geological model (taken to be the permeability value in every grid block in the model), $\mathbf{u}$ represents the well controls, taken to be well bottom-hole pressures (BHPs), and $\mathbf{x} \in \mathbb{R}^{2n_bn_t}$ denotes the state maps ($p$ and $S_w$ in every grid block) at all $n_t$ time steps in the simulation. Here $n_b = n_x n_y n_z$ is the total number of grid blocks in the model, with $n_x$, $n_y$ and $n_z$ the number of  blocks in the $x$, $y$ and $z$ directions.

Data-driven surrogate modeling entails an inexpensive and nonintrusive replacement of the numerical simulator. It applies statistical or machine learning tools to approximate the relationship between state responses $\mathbf{x}$ and rock properties $\mathbf{m}$, for a given set of well controls $\mathbf{u}$, by learning from the training dataset $\{(\mathbf{m}_1, \mathbf{x}_1), \dots, (\mathbf{m}_{n_s}, \mathbf{x}_{n_s})\}$, where $n_s$ is the number of training samples. Traditional machine learning algorithms such as support vector machines and random forest rely on hand-designed kernels to extract useful features. They are not applicable for mapping sets of high-dimensional input ($\mathbf{m}$) to high-dimensional output ($\mathbf{x}$).

A key capability of recent deep-learning-based methods is to simultaneously detect useful features from data and to approximate input-to-output mappings. With such an approach, we approximate the reservoir simulation process as   
\begin{align}
    \mathbf{x} \approx \mathbf{\hat{x}} = \hat{f}(\mathbf{m}, \mathbf{u}; \mathbf{\boldsymbol\theta}),
    \label{eq-g}
\end{align}
where $\hat{f}$ indicates the surrogate model, $\mathbf{\hat{x}} \in \mathbb{R}^{2n_bn_t}$ denotes the approximate state responses, which are expected to be close to the simulated $\mathbf{x}$, and $\boldsymbol \theta$ are the deep neural network parameters determined during the training procedure. Consistent with Eq.~\ref{eq-g}, our goal in this work is to develop a surrogate model to provide the time-dependent states $\mathbf{\hat{x}}$ given a permeability map $\mathbf{m}$, for a fixed set of well controls $\mathbf{u}$.

\subsection{R-U-Net architecture}
\label{sect:r-u-net}

The multiscale spatial correlations that characterize the permeability maps determine the spatial variations of the resulting state maps. Convolutional neural networks (CNNs) are specifically designed to capture this type of spatial information. 
In a CNN, the lower (earlier) layers generally capture more local features, while high (later) layers capture more global information \citep{zeiler2014visualizing}. A general CNN architecture can be formulated recursively through the following expression:
\begin{equation}
    \mathbf{F}_l = \sigma(\mathbf{W}^{\intercal}_{l} * \mathbf{F}_{l-1}(\mathbf{m}) + \mathbf{b}),
\end{equation}
where $\mathbf{F}_l$ and $\mathbf{F}_{l-1}$ denote feature maps at layers $l$ and $l-1$, $\sigma$ represents a nonlinear activation function, $\mathbf{W}^{\intercal}_{l}$ designates a kernel matrix, $*$ denotes the convolution operation, and $\mathbf{b}$ is the bias \citep{dumoulin2016guide}. Feature maps at different layers are also functions of the input map $\mathbf{m}$, and we define $\mathbf{F}_{0} = \mathbf{m}$.


Among different CNN architectures, the U-Net architecture~\citep{ronneberger2015u}, with contracting (encoding) and symmetric expanding (decoding) paths, can efficiently capture hierarchical spatial features and approximate complex nonlinearities between input and output maps. The U-Net is built upon CNNs that do not include any fully connected layers and can take input maps of arbitrary size without the need for architecture modification \citep{long2015fully}. Compared to the DenseED architecture applied for similar problems in \citep{zhu2018bayesian} and \citep{mo2018deep}, the U-Net architecture demonstrated superior performance in capturing flow responses in our experiments. This may be because it facilitates the flow of multiscale information between the encoding and decoding paths in network training, while DenseED only (separately) improves local information flow within the encoding and decoding components.

\begin{figure}[htbp]
  \centering
  \includegraphics[trim={0 5cm 0 3cm}, clip, width=\linewidth]{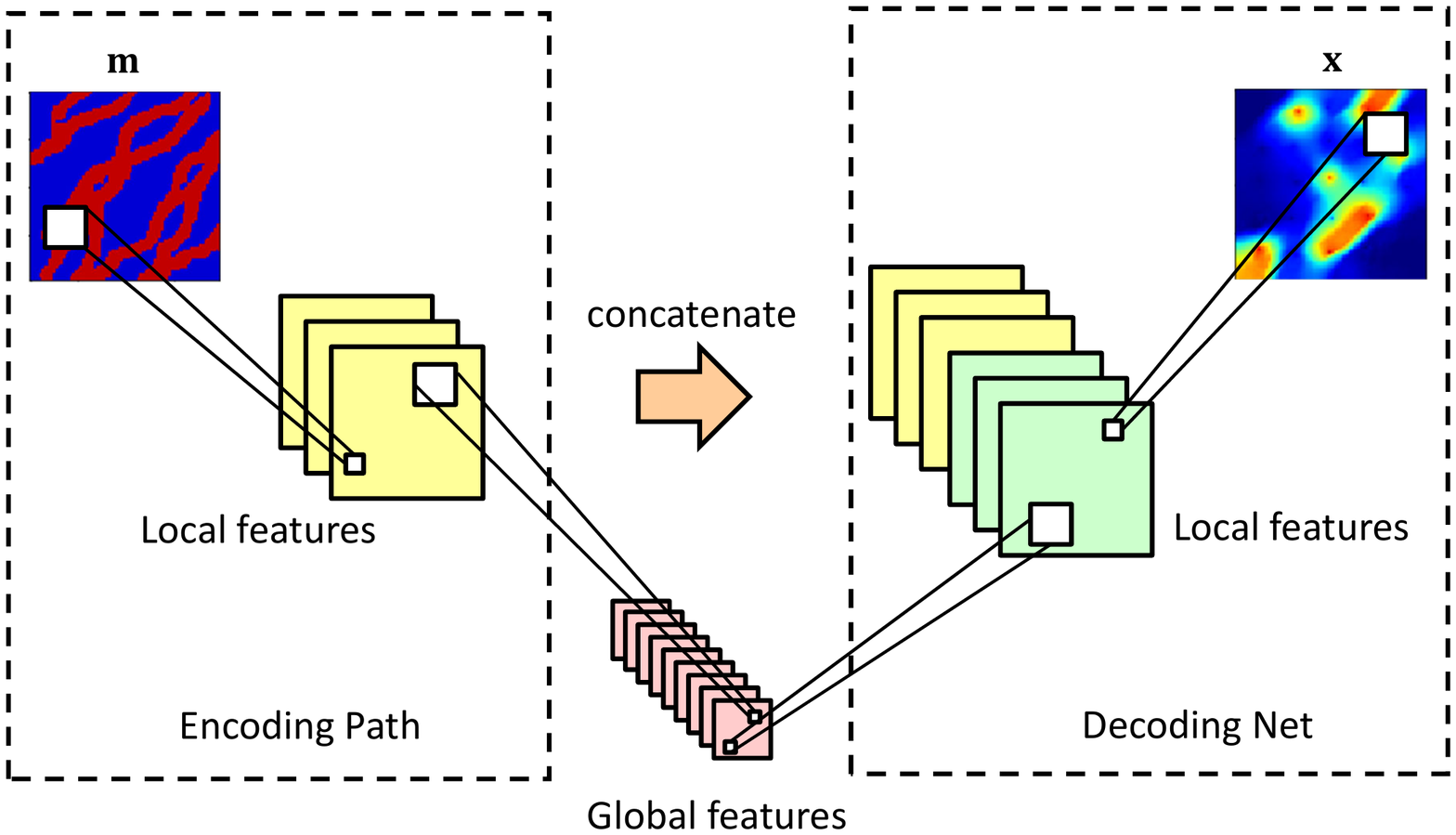}
  \caption{Schematic illustration of R-U-Net architecture (detailed architecture is provided in the Appendix).  R-U-Net consists of encoding and decoding paths, where the local features extracted in the encoding path are concatenated with the upsampled features in the decoding path to produce the state map prediction.}
  \label{fig:r-u-net}
\end{figure}

To further enhance local information flow, we introduce a residual U-Net (R-U-Net), in which residual CNN modules are added into the U-Net architecture \citep{he2016deep}. A schematic diagram of the U-shaped R-U-Net architecture is shown in Fig.~\ref{fig:r-u-net}, where the extracted features in the encoding path are copied and concatenated onto the upsampled features in the decoding path. This enables the multiscale features extracted in the encoding path to be propagated to the corresponding decoding path. 

\begin{figure}[H]
  \centering
  \includegraphics[trim={0 5cm 0 5cm}, clip, width=\linewidth]{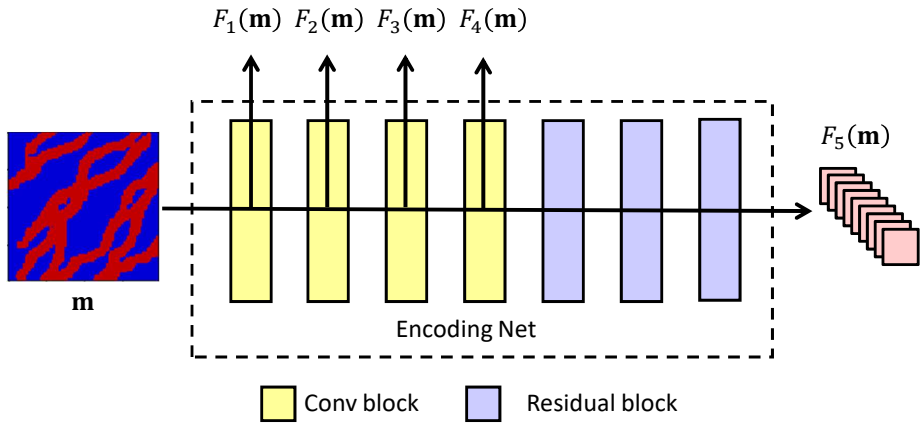}
  \caption{Encoding net consisting of convolutional and residual blocks. The encoding net accepts the permeability map as input. The extracted multiscale features $\mathbf{F}_k(\mathbf{m}) \in \mathbb{R}^{N_{x,k}\times N_{y,k} \times N_{z,k}}$~$(k=1,\dots, 5)$ are input to the decoding net.}
  \label{fig:encoding-net}
\end{figure}

\begin{figure}[H]
  \centering
  \includegraphics[trim={0 5cm 0 5cm},clip,width=\linewidth]{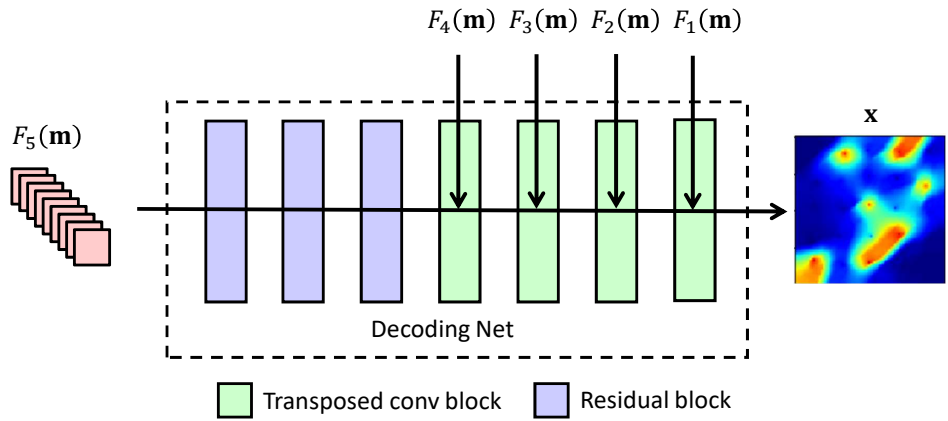}
  \caption{Decoding net consisting of transposed convolutional (upsampling) and residual blocks. The decoding net utilizes the multiscale features $\mathbf{F}_k(\mathbf{m}) \in \mathbb{R}^{N_{x,k}\times N_{y,k} \times N_{z,k}}$ $(k=1, \dots, 5)$ extracted by the encoding net to predict the state map.}
  \label{fig:decoding-net}
\end{figure}

We illustrate the encoding and decoding network architectures in Figs.~\ref{fig:encoding-net} and \ref{fig:decoding-net}. The encoding net shown in Fig.~\ref{fig:encoding-net} takes the permeability map as input. The extracted feature maps $\mathbf{F}_k(\mathbf{m}) \in \mathbb{R}^{N_{x,k}\times N_{y,k} \times N_{z,k}}$~$(k=1,\dots, 5)$ from different encoding blocks will later be copied and fed to the decoding net. Here, $N_{x,k}$ and $N_{y,k}$ denote the dimensions of feature map $\mathbf{F}_k(\mathbf{m})$ along the $x$ and $y$ directions, and $N_{z,k}$ indicates the number of filters in convolutional block $k$. From $\mathbf{F}_1(\mathbf{m})$ to $\mathbf{F}_5(\mathbf{m})$, the extracted features grow from simple and local to complex and global. Residual blocks are applied to produce feature map $\mathbf{F}_5(\mathbf{m})$, which is the most complex and compressed feature map. This map will be fed to the decoding net.

The decoding net illustrated in Fig.~\ref{fig:decoding-net} upsamples the global feature map $\mathbf{F}_5(\mathbf{m})$ to different smaller-scale feature maps, and combines it with the corresponding smaller-scale feature maps $\mathbf{F}_k(\mathbf{m})$ $(k = 1, \dots, 4)$ extracted in the symmetric encoding path. Through this procedure the decoding net provides the target state map. The transposed convolutional block~\citep{dumoulin2016guide} is applied here for upsampling. Similar to the convolutional blocks, this block has tunable weights that must be learned from the training process to achieve optimal upsampling results.

We found the R-U-Net described thus far to be capable of accurately mapping from input permeability fields to output steady-state pressure fields. However, in our inverse modeling scenarios, we are interested in reservoir dynamics, which require the surrogate model to capture the relationship between the input property map $\mathbf{m}$ and the dynamic state maps $\mathbf{x}=[\mathbf{x}^1,\dots, \mathbf{x}^{n_t}]$ over $n_t$ time steps. The R-U-Net as described fails to provide acceptable approximations for complex dynamic systems because the time-dependent information is not encoded and captured by the feed-forward architecture. This motivates us to investigate a recurrent R-U-Net architecture, which we now describe.

\subsection{Recurrent R-U-Net architecture}
\label{sect:recurrent_arch}

To capture temporal dynamics, we apply a recurrent R-U-Net architecture. The ability of recurrent neural networks (RNNs) to capture temporal dynamics stems from the fact that the composite RNN input at the current time step contains historical information  \citep{mikolov2010recurrent}. In practice, long short term memory (LSTM) \citep{hochreiter1997long}, which is a variant of the standard RNN architecture, is often applied to treat long-term temporal dependency. This is because the set of gates used in LSTM improves information flow and solves the vanishing gradient problem that is common in standard RNNs \citep{bengio1994learning}.

\begin{figure}[H]
  \centering
  \includegraphics[trim={0 1cm 0 2cm},clip, scale=0.6]{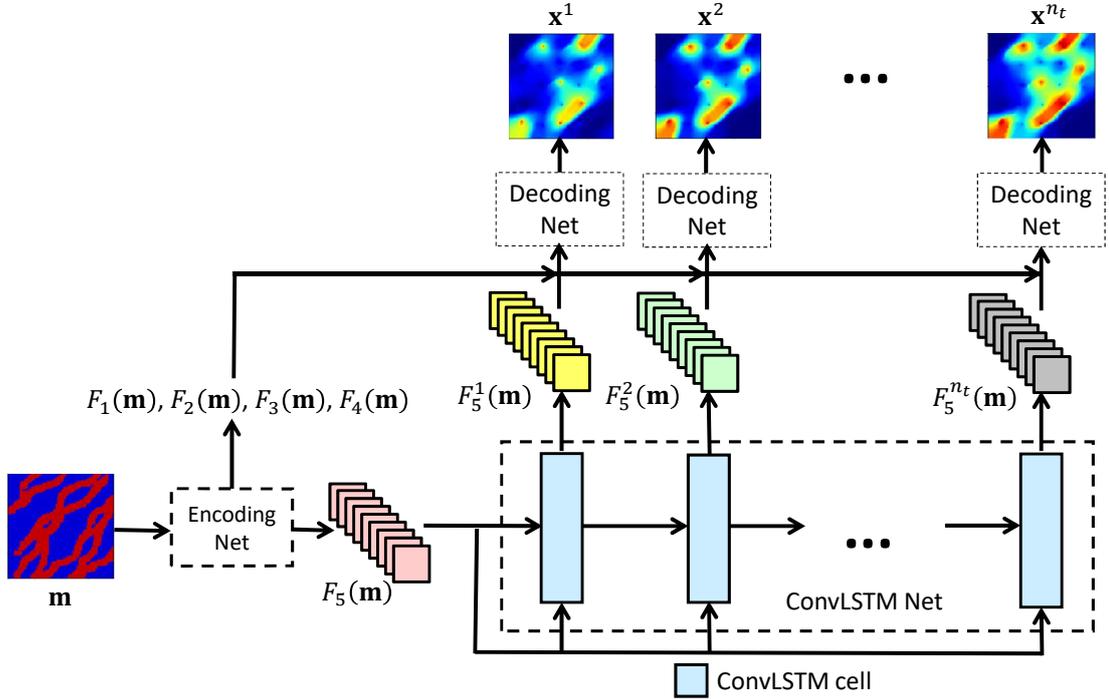}
  \caption{Recurrent R-U-Net architecture incorporating convLSTM into the R-U-Net. The convLSTM net takes the global feature map $\mathbf{F}_5(\mathbf{m})$ from the encoding net and generates a sequence of feature maps $\mathbf{F}_5^t(\mathbf{m})$ $(t=1,\dots, n_t)$ that will be decoded into a sequence of state maps $\mathbf{x}^t$ $(t=1,\dots, n_t)$ separately, using the same decoding net.}
  \label{fig:recurrent-r-u-net}
\end{figure}

It is desirable to incorporate an LSTM architecture into the R-U-Net architecture to capture temporal dynamics. To achieve a more compact architecture, which may significantly benefit training efficiency, the LSTM architecture is incorporated only on feature map $\mathbf{F}_5(\mathbf{m})$, as illustrated in Fig.~\ref{fig:recurrent-r-u-net}. This is because this feature map carries the most compressed representation of the input property map $\mathbf{m}$, and the temporal evolution of state maps can be expressed by evolving $\mathbf{F}_5(\mathbf{m})$. In addition, in order to maintain the encoded spatial information while avoiding a large number of extra parameters, as would be introduced by a conventional fully connected LSTM, the convolutional LSTM (convLSTM) \citep{xingjian2015convolutional} is adopted. 
 
The convLSTM net is composed of a chain of repeating convLSTM cells, which share the same set of (trainable) weights. In the convLSTM cell capturing information at time~$t$, the input~$\boldsymbol\chi^t$, output state (also referred to as the hidden state) $\mathbf{H}^t$, cell state $\mathbf{C}^t$, and the different gates are all 3D tensors. The cell state $\mathbf{C}^t$ serves as the memory of the convLSTM net and is updated via
   \begin{equation}
     \mathbf{C}^t = \mathbf{f}^t \circ \mathbf{C}^{t-1} + \mathbf{i}^t \circ \tilde{\mathbf{C}}^t,
 \end{equation}
where $\circ$ denotes the Hadamard product, $\mathbf{C}^{t-1}$ is the cell state at the previous time step, $\tilde{\mathbf{C}}^t$ is the new candidate cell state, $\mathbf{f}^t$ is the `forget gate' that controls what information to eliminate from the previous cell state $\mathbf{C}^{t-1}$, and $\mathbf{i}^t$ is the input gate that determines what information to update from the proposed cell state $\tilde{\mathbf{C}}^t$. The output state $\mathbf{H}^t$ is updated based on $\mathbf{C}^t$ filtered by the output gate $\mathbf{o}^t$ and is given by
   \begin{equation}
     \mathbf{H}^t = \mathbf{o}^t \circ \tanh(\mathbf{C}^t).
 \end{equation}
Here $\mathbf{o}^t$ determines which information in the cell state $\mathbf{C}^t$ is transferred to the output state $\mathbf{H}^t$.
 
The values of the different gates $\mathbf{f}^t$, $\mathbf{i}^t$, $\mathbf{o}^t$ and the proposed cell state~$\tilde{\mathbf{C}}^t$ are determined from the previous output state $\mathbf{H}^{t-1}$ and the current input $\boldsymbol\chi^t$. The specific expressions for these quantities are
 \begin{equation}
     \mathbf{f}^t = \sigma(\mathbf{W}_{xf} * {\boldsymbol\chi}^t + \mathbf{W}_{hf}*\mathbf{H}^{t-1} + \mathbf{b}_f),
 \end{equation}
  \begin{equation}
     \mathbf{i}^t = \sigma(\mathbf{W}_{xi} * {\boldsymbol\chi}^t + \mathbf{W}_{hi}*\mathbf{H}^{t-1} + \mathbf{b}_i),
 \end{equation}
  \begin{equation}
     \mathbf{o}^t = \sigma(\mathbf{W}_{xo} * {\boldsymbol\chi}^t + \mathbf{W}_{ho}*\mathbf{H}^{t-1} + \mathbf{b}_o),
 \end{equation}
  \begin{equation}
     \tilde{\mathbf{C}}^t = \tanh(\mathbf{W}_{xc} * {\boldsymbol\chi}^t + \mathbf{W}_{hc}*\mathbf{H}^{t-1} + \mathbf{b}_c).
 \end{equation}
In the above expressions, $\mathbf{W}$ and $\mathbf{b}$ are convolution filter weights and bias terms, which are both shared across convLSTM cells. The parameters associated with these quantities are tuned during the training process.

The convLSTM net and its variants have been used in a range of application areas, including precipitation forecasting~\citep{xingjian2015convolutional}, 
video gesture recognition \citep{zhu2017multimodal} and MRI cardiac segmentation \citep{poudel2016recurrent}, where they have been shown to be effective in capturing both spatial and temporal information. The integration of the convLSTM net into the R-U-Net provides the recurrent R-U-Net developed in this study. As illustrated in Fig.~\ref{fig:recurrent-r-u-net}, the recurrent R-U-Net takes the property map $\mathbf{m}$ as input, and the corresponding multiscale feature maps $\mathbf{F}_k(\mathbf{m})$ $(k=1,\dots, 5)$ are extracted by the encoding net. Then the convLSTM net takes the most compressed feature map $\mathbf{F}_5(\mathbf{m})$ and generates a sequence of feature maps $\mathbf{F}_5^t(\mathbf{m})$ $(t=1,\dots, n_t)$. These are then decoded separately, by the same decoding net, into a sequence of state maps $\mathbf{x}^t$ $(t=1,\dots, n_t)$. After proper training, our recurrent R-U-Net can produce a sequence of state maps $[\mathbf{x}^1, \dots, \mathbf{x}^{n_t}]$ that describe reservoir dynamics for an input property map $\mathbf{m}$ (and a fixed set of controls $\mathbf{u}$).

\subsection{Training procedure}
\label{sect:training}

\begin{figure}[htbp]
  \centering
  \includegraphics[trim={1cm 3cm 0 3cm},clip,scale=0.7]{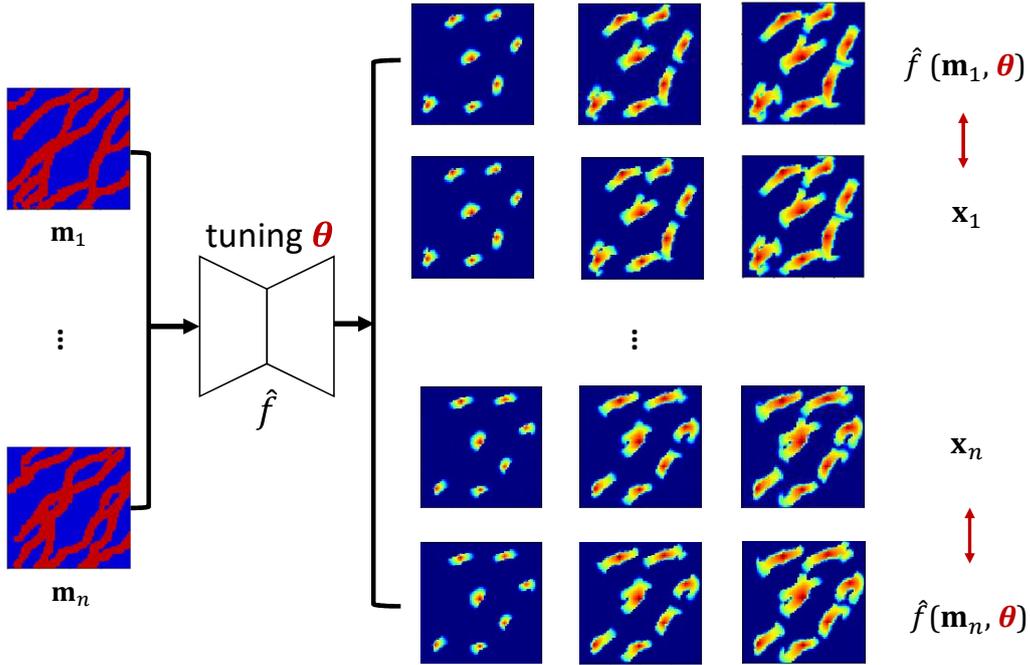}
  \caption{Schematic of the training procedure. The parameters $\boldsymbol\theta$ in the recurrent R-U-Net are determined by minimizing the objective function in Eq.~\ref{eq:loss-function} using backpropagation.} 
  \label{fig:training-procedure}
\end{figure}

During training, in order to allow the recurrent R-U-Net to learn the temporal dynamics of the system given a permeability map $\mathbf{m}$, we minimize the difference between the sequence of target state maps $\mathbf{x}^t_i$, generated from a set of high-fidelity simulations, and the sequence of state maps $\hat{\mathbf{x}}^t_i$, found by the recurrent R-U-Net (i.e., through application of $\hat{f}(\mathbf{m}_i, \mathbf{u}_i; \boldsymbol\theta)$). This training set includes a sequence of states generated for each geomodel $\mathbf{m}_i$, $i=1, \dots, n_s$, where $n_s$ is the total number of geomodels in the training set. The training process is illustrated in Fig.~\ref{fig:training-procedure}.

The training objective is to minimize the $L^p$ norm of the difference between the $\mathbf{x}^t_i$ and the ${\hat {\mathbf x}}^t_i$. Extra weight is placed on the states in blocks containing wells in order to improve the accuracy of the well flow rates (computed through application of Eq.~\ref{eq:well_flow}), which are usually the key data we seek to match in history matching studies. The specific minimization is as follows: 
\begin{equation}
\argmin_{\boldsymbol\theta} \frac{1}{n_s}\frac{1}{n_t} \sum_{i=1}^{n_s} \sum_{t=1}^{n_t} ||{\hat {\mathbf x}}^t_i - \mathbf{{x}}_{i}^{t}||_{p}^p + \lambda\frac{1}{n_s}\frac{1}{n_t}\frac{1}{n_w} \sum_{i=1}^{n_s} \sum_{t=1}^{n_t} \sum_{w=1}^{n_w} ||{\hat {\mathbf x}}^t_{i,w} - \mathbf{{x}}_{i,w}^{t}||_{p}^p.
\label{eq:loss-function}
\end{equation}
Here $\lambda$ is the additional weighting for the well states, which is applied at $n_w$ well locations. Our numerical experiments showed that the use of the 
$L^2$ norm results in better predictions of the saturation maps, while the use of the $L^1$ norm leads to slightly more accurate pressure maps. Therefore, we use two separate recurrent R-U-Nets, which are trained with an $L^2$ norm loss for saturation and an $L^1$ norm loss for pressure.

In the training process, the loss function is minimized by tuning the network parameters~$\boldsymbol\theta$. The gradient of the loss function with respect to $\boldsymbol\theta$ is automatically computed by backpropagation \citep{hecht1992theory} through the recurrent R-U-Net. In this work, we use the adaptive moment estimation (ADAM) \citep{kingma2014adam} optimization algorithm, which is an extension of stochastic gradient descent (SGD). This has been found to be an effective procedure for the training of many deep neural network architectures. 

The training of the recurrent R-U-Net can be accomplished efficiently, though the specific training time depends on many factors. These include the training set size, training batch size, optimizer setup and learning rate, as well as the graphics processing unit (GPU) performance. Although the training time can vary by case, it is very small compared to the time that would be required if we were to perform high-fidelity simulations in the history matching procedure. The training process converges in 80~minutes or less using a Nvidia Tesla V100 GPU for the cases considered in this study. Our example involves a 2D geomodel defined on an $80 \times 80$ grid, and we predict pressure and saturation states at 10 time steps. After training, given a new geomodel, the recurrent R-U-Net can provide predictions for the states, at 10~time steps, in an elapsed time of about 0.01 seconds using GPU. 

There are about 2.6~million trainable parameters in our recurrent R-U-Net. As is the case with other well-designed deep neural networks, this over-parameterized network does not appear to suffer from over-fitting. A rigorous explanation for this has yet to be provided. Recent studies explain this general observation in terms of intrinsic dimension~\citep{li2018measuring} and the lottery-ticket hypothesis~\citep{frankle2018lottery}, and they suggest that the number of parameters does not represent the true complexity of deep neural networks. In addition, as suggested in \citep{zhu2018bayesian}, the high-dimensional rock property and reservoir state data embed essential physical and dynamical information, and this acts to regularize the problem and thus reduce over-fitting.

\subsection{Data processing}
\label{sect:data_proc}

Data pre-processing is important for the effective training of deep neural networks. Constraining training data values to be near zero, for example, can enhance the deep neural network training in many cases. In this study, the input property map $\mathbf{m}$ is a binary geological facies map, and the output state maps $\mathbf{x}$ are water saturation  and pressure in each grid block. The binary facies map is naturally represented by 0 (denoting shale/mud) and 1 (denoting sand/channel) block values, and saturation map values are physically between 0 and 1. Thus these two maps do not require any pre-processing.
 
Pressure map values, however, typically display large ranges, and the mean is far from 0. 
Previous (related) studies \citep{zhu2018bayesian, mo2018deep} considered steady-state flow, and a simple min-max normalization was used for the pressure data. This treatment can lead to significant prediction error in dynamic cases. 
 
Based on the observation that the training samples generally provide reliable statistics for pressure at different time steps, we detrend the original pressure map data by performing data normalization at each time step $t$ in the training set. Specifically, we compute the mean pressure map $\mathbf{\bar{p}}^{t}$ at each time step as
  \begin{equation}
     \mathbf{\bar{p}}^{t} = \frac{1}{n_s}\sum_{i}^{n_s} \mathbf{p}^{t}_i \quad (t = 1, \dots, n_t).
     \label{eq:p-0}
\end{equation}
At each time step $t$, we subtract this mean pressure map from each training pressure map to give the difference map $\mathbf{\hat{p}}^{t}_i$, i.e.,
 \begin{equation}
     \mathbf{\hat{p}}^{t}_i = \mathbf{p}^{t}_i - \mathbf{\bar{p}}^{t}, \quad i=1,\dots, n_s, \ \ t = 1, \dots, n_t.
     \label{eq:p-1}
\end{equation}
Finally, we perform min-max normalization over the difference map $\hat{\mathbf{p}}^t_i$ at each time step $t$ via application of
\begin{equation}   
     \mathbf{\tilde{p}}^{t}_i = \frac{\mathbf{\hat{p}}^{t}_i - \text{min} ([\mathbf{\hat{p}}^{t}_1, \dots, \mathbf{\hat{p}}^{t}_{n_s}])}{\text{max} ([\mathbf{\hat{p}}^{t}_1, \dots, \mathbf{\hat{p}}^{t}_{n_s}]) - \text{min} ([\mathbf{\hat{p}}^{t}_1, \dots, \mathbf{\hat{p}}^{t}_{n_s}])},  \quad i=1,\dots, n_s, \ \ t = 1, \dots, n_t,
     \label{eq:p-2}
 \end{equation}
where the `max' and `min' operations here find the maximum and minimum scalar values for the entire model over all training pressure map samples at time step $t$. 


The maximum and minimum values, along with all of the $\mathbf{\bar{p}}^t$ maps, are saved. These are then used for the inverse transform of the predicted pressure maps; i.e., to transform from the predicted $\mathbf{\tilde{p}}^{t}$ map to the physical pressure map $\mathbf{p}^{t}$. Numerical experimentation demonstrated that this treatment acts to clearly enhance the accuracy of the recurrent R-U-Net pressure predictions.


\section{Surrogate Model Evaluation}
\label{sect:flowstats}
In this section, we describe a 2D binary facies channelized model and define a specific oil-water flow problem. We then evaluate the performance of our recurrent R-U-Net surrogate model for this system. Pressure and saturation field evolution and well responses are considered. These quantities are assessed for individual realizations and for an ensemble of models.

\subsection{Flow problem setup}
\label{sect:setup}

Binary channelized models, such as those considered here, are more demanding to treat than Gaussian models in many respects. They are, for example, much more difficult to parameterize than Gaussian models, and they are less suited for surrogate modeling since the range of flow responses can be very large. Thus the use of channelized geomodels provides a challenging test for our R-U-Net surrogate model.

\begin{figure}[htbp]
  \centering
  \includegraphics[scale=0.7]{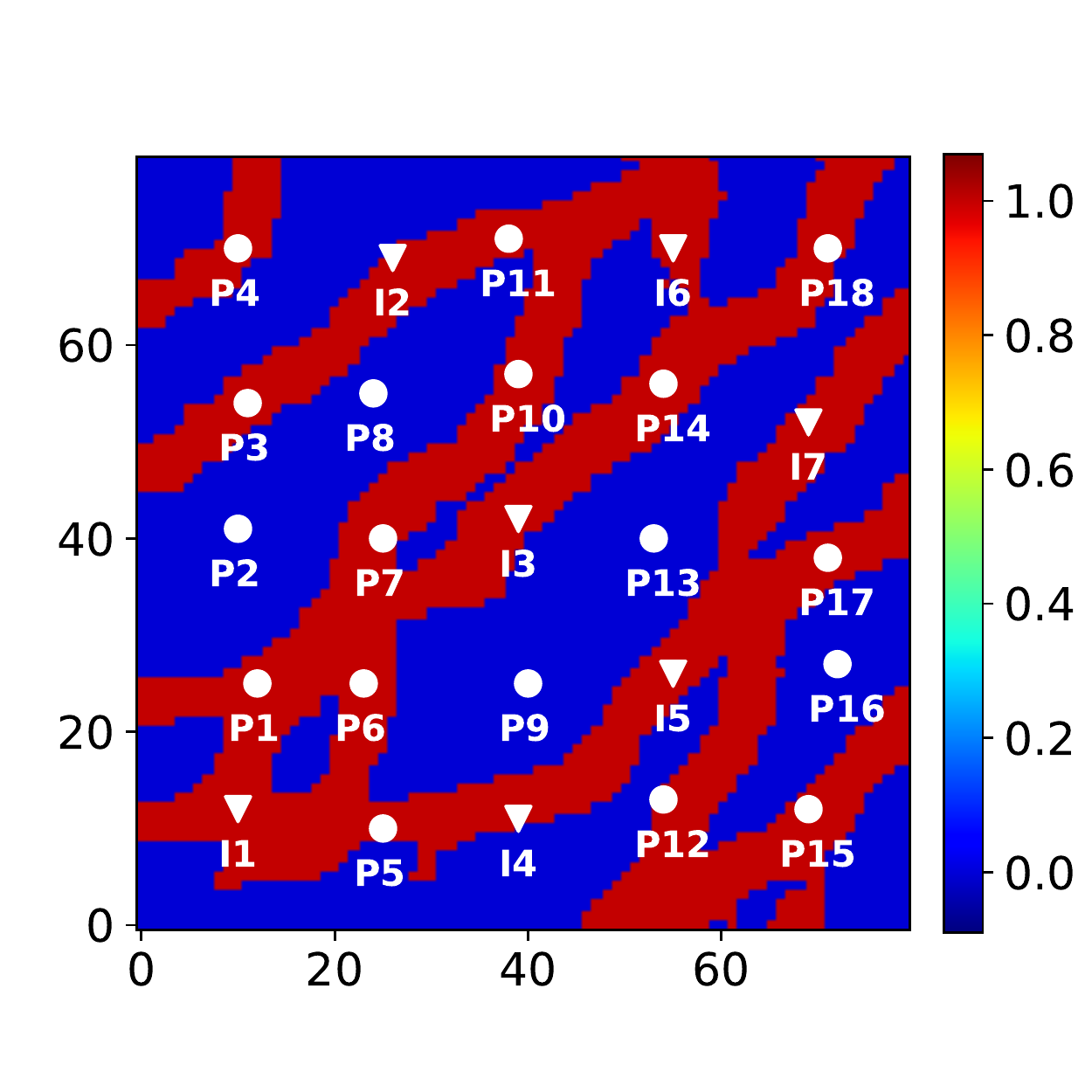}
  \caption{Channelized $80\times80$ facies map, conditioned to facies type at 25 wells.}
  \label{fig:chap3-model-set-up}
\end{figure}

One realization of the channelized system considered here, in terms of a binary facies (rock type) map, is shown in Fig.~\ref{fig:chap3-model-set-up}. The geomodel is defined on an $80\times80$ grid, with each grid block of size 50~m $\times$ 50~m $\times$ 10~m (in the $x$, $y$ and $z$ directions respectively). The model contains 25 wells (seven water injection wells and 18 production wells), and the geomodels are conditioned to facies type at all wells. For a particular grid block ($i$), $m_i=1$ indicates channel (sand), and $m_i = 0$ indicates shale (mud). The permeability $k_i$ is related to the facies type via the expression $k_i = a\exp(bm_i)$, with $a = 30$~md and $b = \ln(\frac{2000}{30})$. This results in $k_i=2000$~md for blocks with sand, and $k_i=30$~md for blocks with shale.

\begin{figure}[htbp]
  \centering
  \includegraphics[scale=0.5]{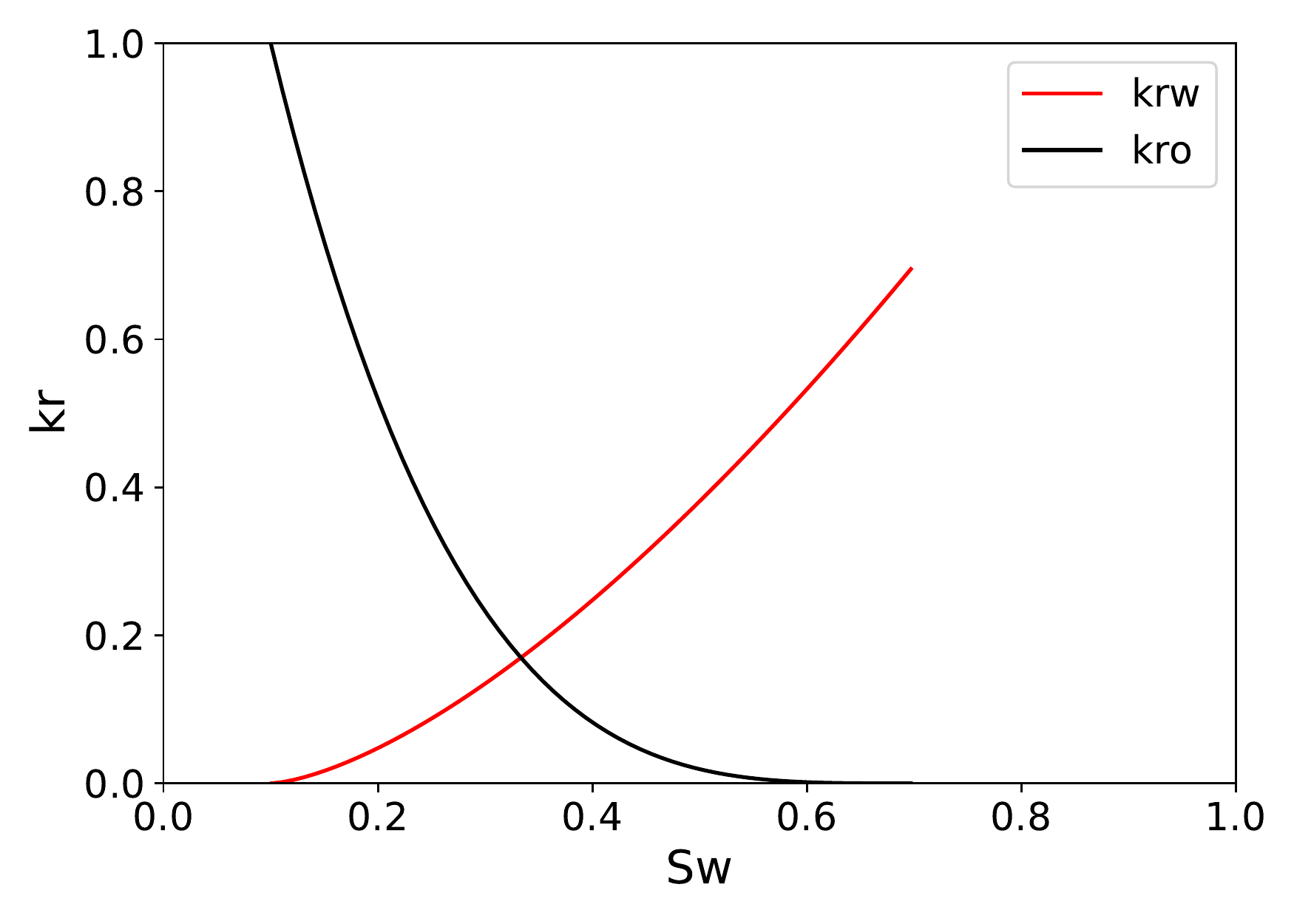}
  \caption{Oil-water relative permeability curves.}
  \label{fig:chap3-rel-perm}
\end{figure}

All wells are specified to operate are under bottom-hole pressure (BHP) control. Injection well BHPs are set to 330~bar, and production well BHPs to 320~bar. The initial oil and water saturations are 0.9 and 0.1. The oil viscosity changes with reservoir pressure and is 0.29~cp at the initial reservoir pressure of 325~bar. Water viscosity is constant at 0.31~cp. The oil-water relative permeability curves are shown in Fig.~\ref{fig:chap3-rel-perm}. Porosity is set to a constant value of 0.2. 

\subsection{Geomodel generation and training simulations}

The recurrent R-U-Net surrogate model requires a number of training flow simulations to learn the correct mapping from the input permeability field to the dynamic output states (pressure and saturation). Geomodel realizations for this training step can be generated by a geological modeling package such as SGeMS \citep{remy2009applied}. SGeMS realizations are constructed to honor the features (i.e., multipoint spatial statistics) that exist in a prescribed geological `training image.' Data measured at wells, referred to as hard data, are also honored in the resulting realizations.

In this work, rather than apply SGeMS directly, we use a parameterized representation for geological realizations. This parameterization is denoted  CNN-PCA (PCA here indicates principal component analysis) \citep{liu2019deep, liu2019multilevel}. This representation entails the use of a CNN to post-process PCA-based realizations (which capture two-point spatial statistics but not multipoint statistics) into geomodels with the requisite channel structure and continuity. CNN-PCA models were shown to provide flow results in close agreement with those from SGeMS realizations \citep{liu2019deep}. The key advantage of the CNN-PCA representation is that it enables us to represent the geomodel, which in this case contains 6400 grid blocks, in terms of, e.g., $O(100)$ parameters. 
This is very beneficial in history matching applications, since many fewer parameters need to be determined.

\begin{figure}
     \centering
     \begin{subfigure}[b]{0.32\textwidth}
         \centering
         \includegraphics[width=\textwidth]{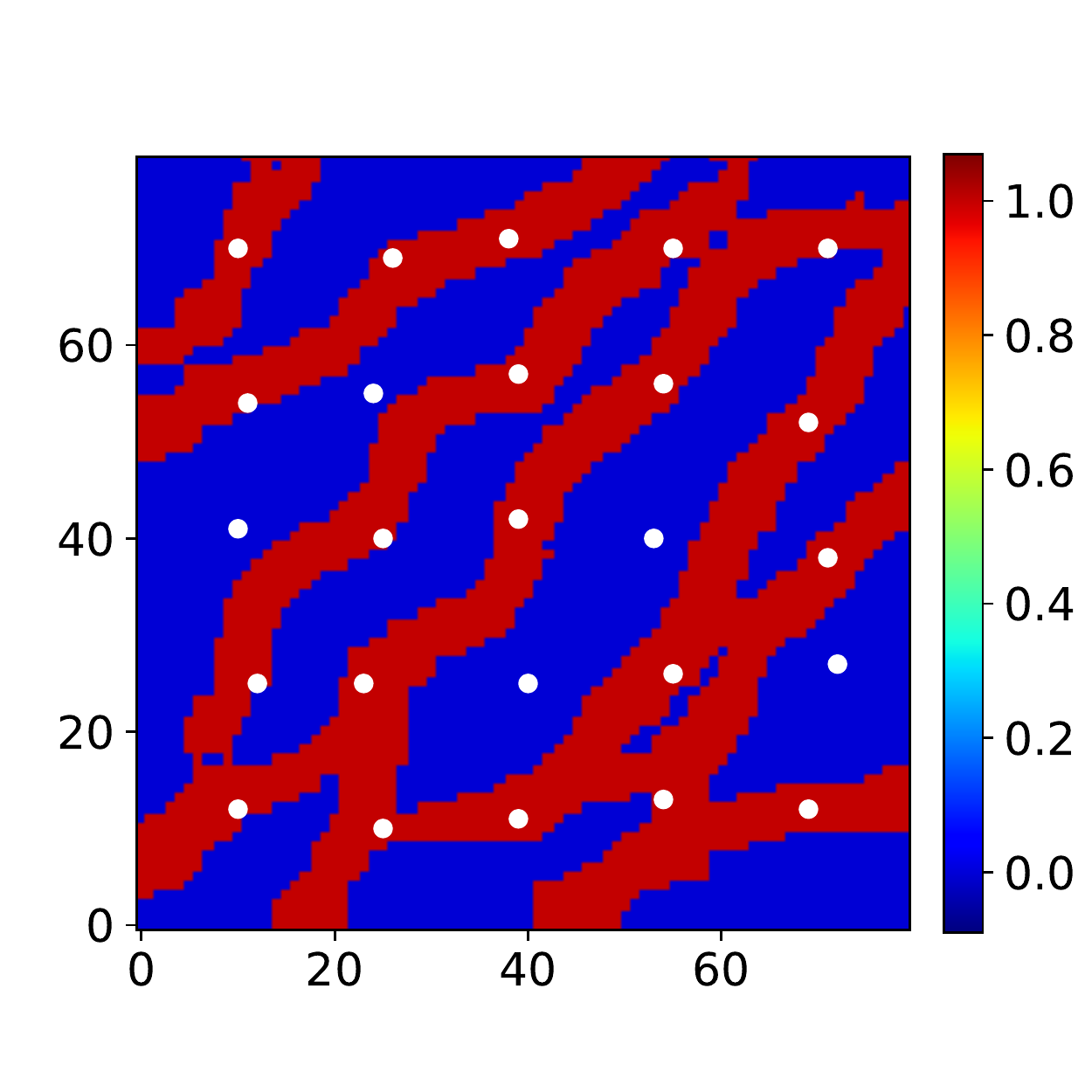}
         \caption{}
         \label{real-3}
     \end{subfigure}
     \begin{subfigure}[b]{0.32\textwidth}
         \centering
         \includegraphics[width=\textwidth]{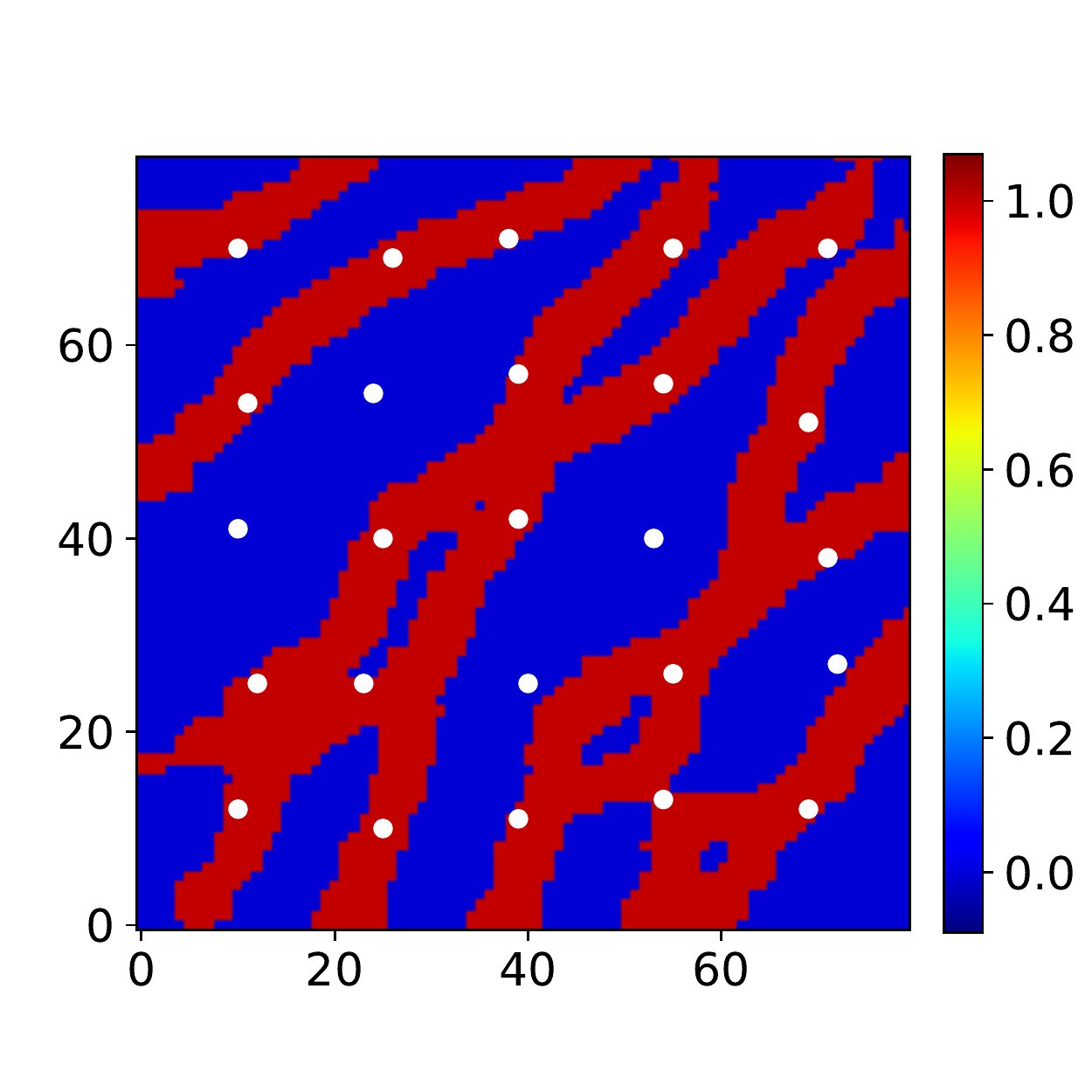}
         \caption{}
         \label{real-4}
     \end{subfigure}
     \begin{subfigure}[b]{0.32\textwidth}
         \centering
         \includegraphics[width=\textwidth]{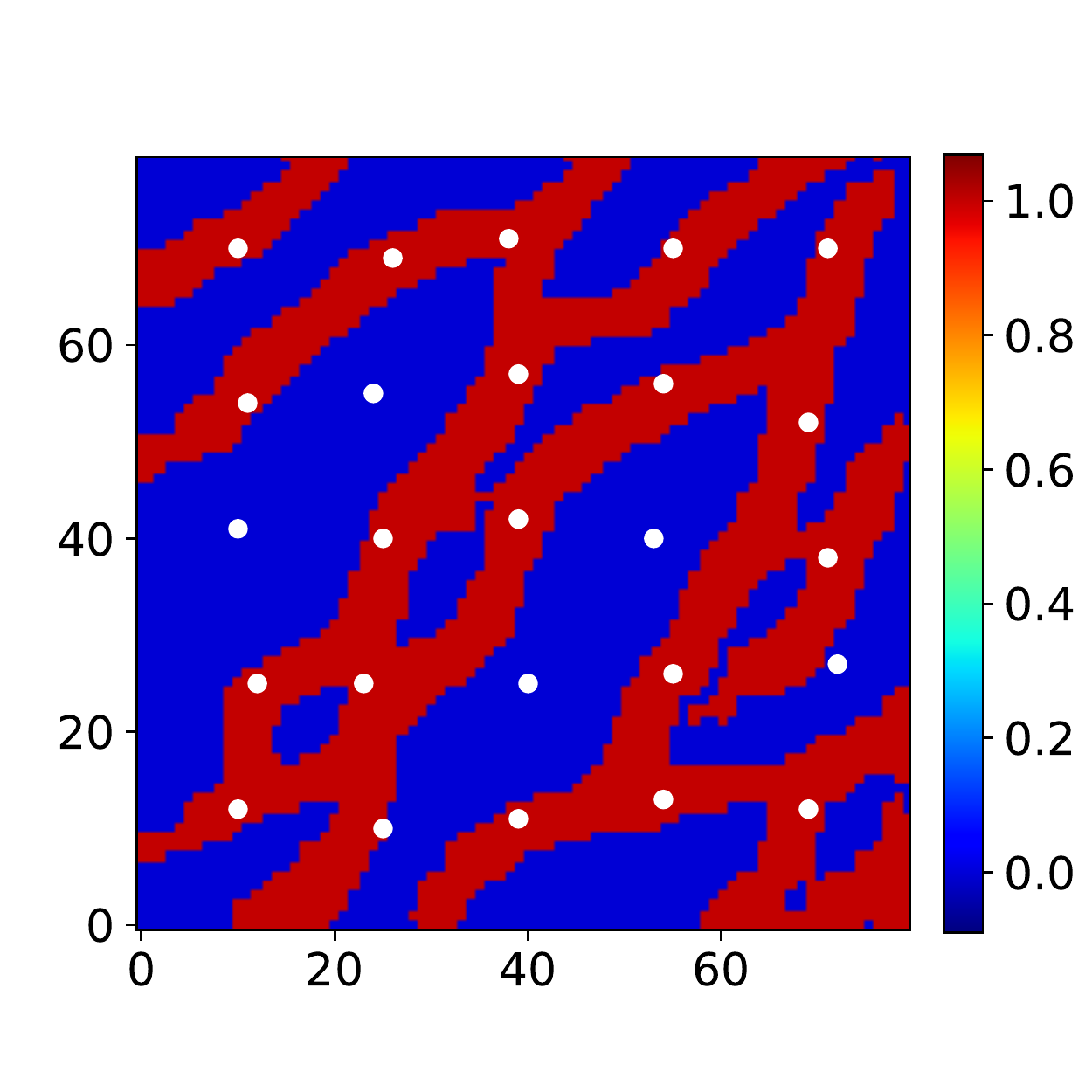}
         \caption{}
         \label{real-5}
     \end{subfigure}
     
     \begin{subfigure}[b]{0.32\textwidth}
         \centering
         \includegraphics[width=\textwidth]{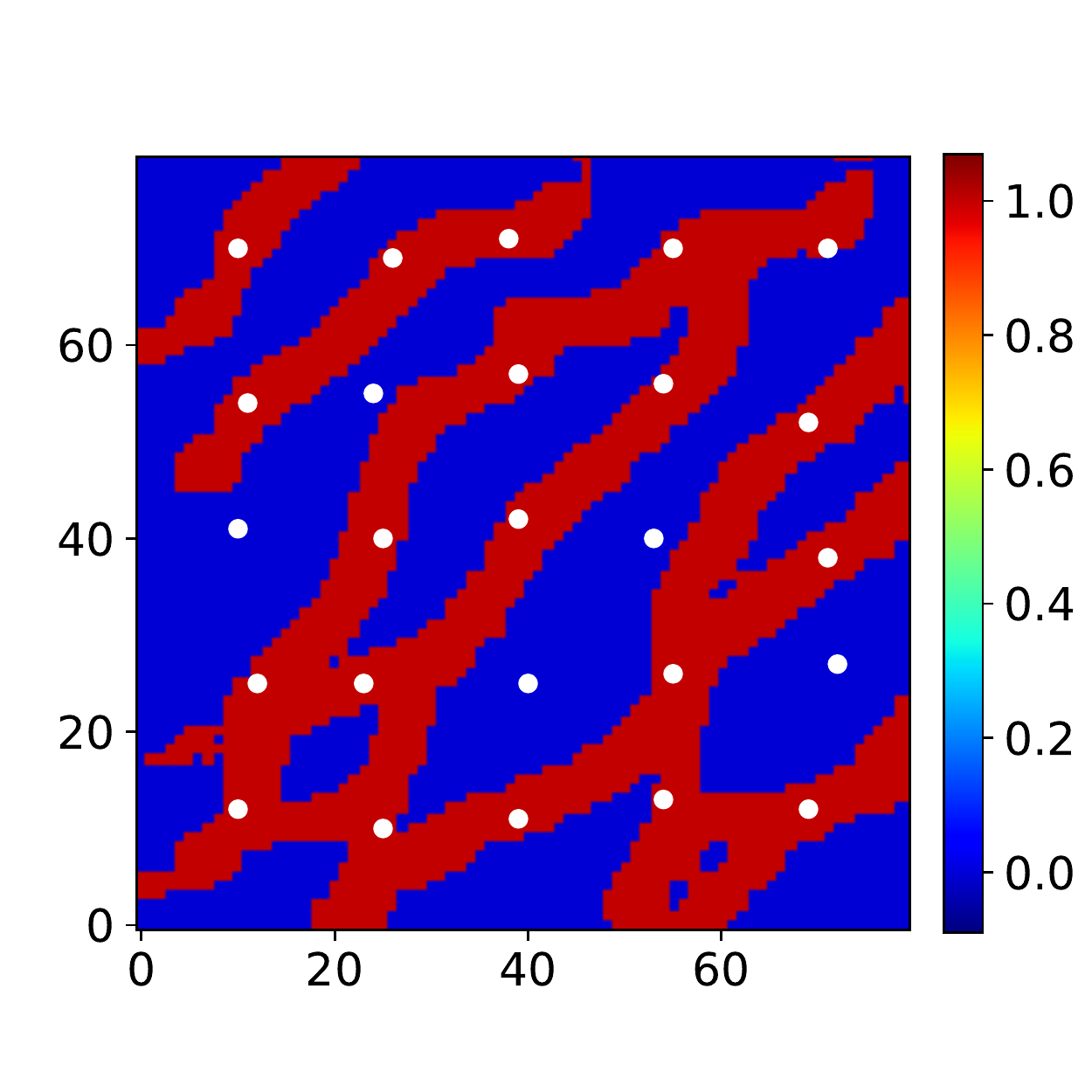}
         \caption{}
         \label{real-6}
     \end{subfigure}
          \begin{subfigure}[b]{0.32\textwidth}
         \centering
         \includegraphics[width=\textwidth]{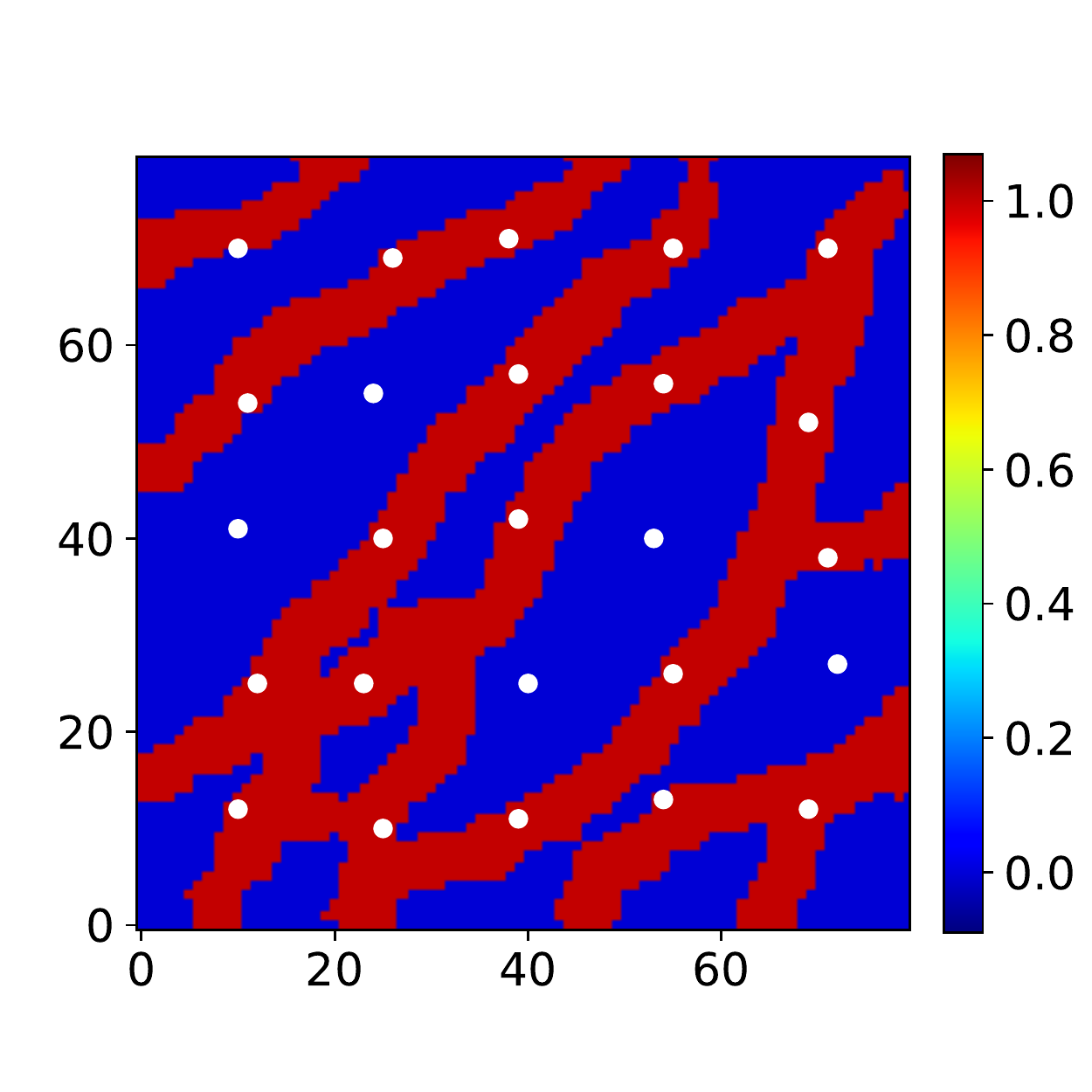}
         \caption{}
         \label{real-7}
     \end{subfigure}
          \begin{subfigure}[b]{0.32\textwidth}
         \centering
         \includegraphics[width=\textwidth]{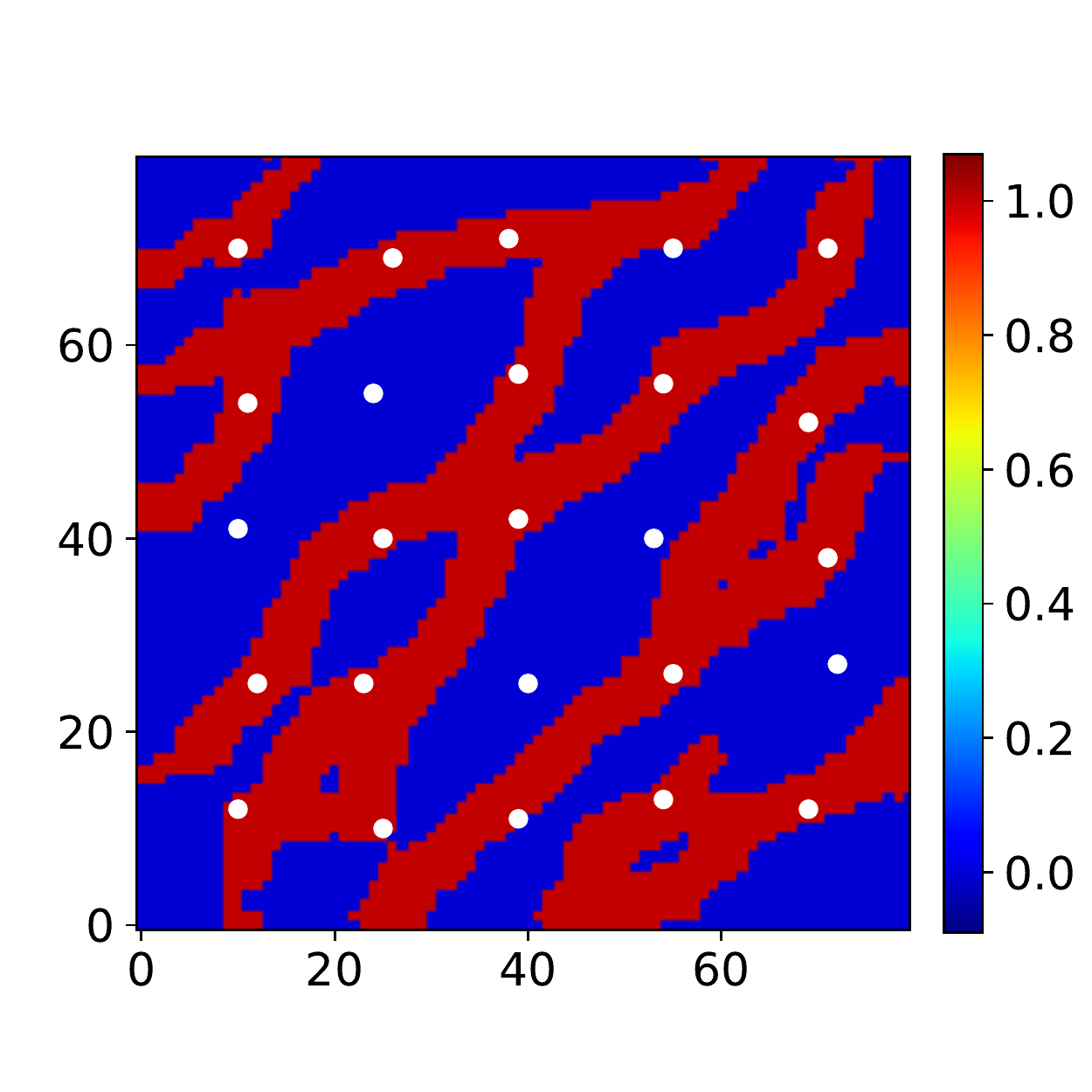}
         \caption{}
         \label{real-8}
     \end{subfigure}
    
    \caption{Six random channelized realizations generated by CNN-PCA. All models honor facies data at the 25 wells. Model in (a) used in the assessments in Sections~\ref{sect:map_pred} and \ref{sect:well_pred}.}
    \label{fig:cnn-pca-reals}
\end{figure}

CNN-PCA generates high-dimensional models by first sampling a lower-dimensional variable $\boldsymbol\xi$ from the standard Gaussian distribution. In this study, the dimension of $\boldsymbol\xi$, $n_{\xi}$, is set to 100, which is generally consistent with the values used in \citep{liu2019deep, liu2019multilevel}. Fig.~\ref{fig:cnn-pca-reals} displays six random channelized facies models generated by CNN-PCA with $n_{\xi} = 100$. A final hard-thresholding step is applied to provide strictly binary fields. The white points in the figures depict the well locations, where the $m_i$ are conditioned to honor facies data. Specifically, there are five wells drilled in mud, and 20 wells drilled in sand. Given the facies model, we construct permeability in each grid block through application of $k_i=a\exp(bm_i)$, with $a$ and $b$ as given previously.

Once the geomodels are constructed, we simulate flow using Stanford's Automatic Differentiation-based General Purpose Research Simulator, AD-GPRS \citep{zhou2012parallel}. We simulate each model over a time frame of 1000~days. We collect training data (pressure and saturation maps) at 10 time steps, from 50~days to 1000~days. More variation in the states occurs at earlier times in the simulations, and the data collection is skewed to capture this.

\subsection{Training procedure}
\label{sect:training_proc}

As noted earlier, we found that the use of the $L^1$ norm loss results in more accurate recurrent R-U-Net pressure predictions, while the $L^2$ norm loss provides better saturation  predictions. Therefore, we train two separate recurrent R-U-Nets with exactly the same architectures but with different training sets (one with pressure and one with saturation).

There are several important hyperparameters that require specification before training, including learning rate, batch size, number of epochs, and the weight $\lambda$ defined in the loss function (given in Eq.~\ref{eq:loss-function}). The two recurrent R-U-Nets share the same hyperparameter setup, with initial learning rate $l_r= 0.003$ for the ADAM optimization algorithm, batch size $N_b = 8$, and weight $\lambda = 1000$. The training of both nets converges within 200~epochs, though we observed that the training for saturation usually converges faster than that for pressure. The optimal hyperparameter values can usually be found by conducting grid search or random search \citep{bergstra2012random} over the specified value ranges. In our numerical experiments, we found the training to not be very sensitive to hyperparameter values after appropriate data pre-processing. Therefore, the same set of hyperparameters can be used as the initial setup for a new training set.

In this study, we use a training sample size of 1500, which means we have 1500 random channelized permeability fields (generated by CNN-PCA) and corresponding sequences of state maps (generated by AD-GPRS). Although more training data will usually lead to higher prediction accuracy, it also corresponds to higher pre-processing cost, so there is a tradeoff between these two objectives. As noted earlier, not counting the 1500 training AD-GPRS simulations, it takes about 80~minutes to train each of the recurrent R-U-Nets applied in this study. Since we train networks (separately) for pressure and saturation, this corresponds to a total of 160~minutes of training. These trainings can however be performed in parallel when a single GPU is available, in which case elapsed time is only 80~minutes.

The way in which the training time scales with problem size, number of training runs, number of time steps considered, etc., is important for practical applications. This is a complicated issue because some aspects of the training are expected to scale linearly with problem size, while others scale sub-linearly (since the number of training parameters will stay the same). These scalings should be investigated and quantified in future work. We do expect, however, that with larger models training time will continue to be small compared to the time required for data assimilation using full numerical simulation.


\subsection{Saturation and pressure map predictions}
\label{sect:map_pred}

We now evaluate the performance of the trained recurrent R-U-Nets. A total of 500 test cases are considered. The geomodels in the test set are new (random) CNN-PCA realizations. AD-GPRS is applied to produce the reference state maps. The recurrent R-U-Net state maps are generated at 10 time steps, though here we present results at just three (representative) time steps, 50, 400 and 850~days, for a single realization. The realization considered here is shown in Fig.~\ref{fig:cnn-pca-reals}a. This particular geomodel provides saturation and pressure results that correspond to errors (quantified below) that are slightly larger than the median errors over the 500 test cases. Thus the saturation and pressure results for the majority of test cases are of higher accuracy than the results that follow.


\begin{figure}
     \centering
     \begin{subfigure}[b]{0.325\textwidth}
         \centering
         \includegraphics[width=\textwidth]{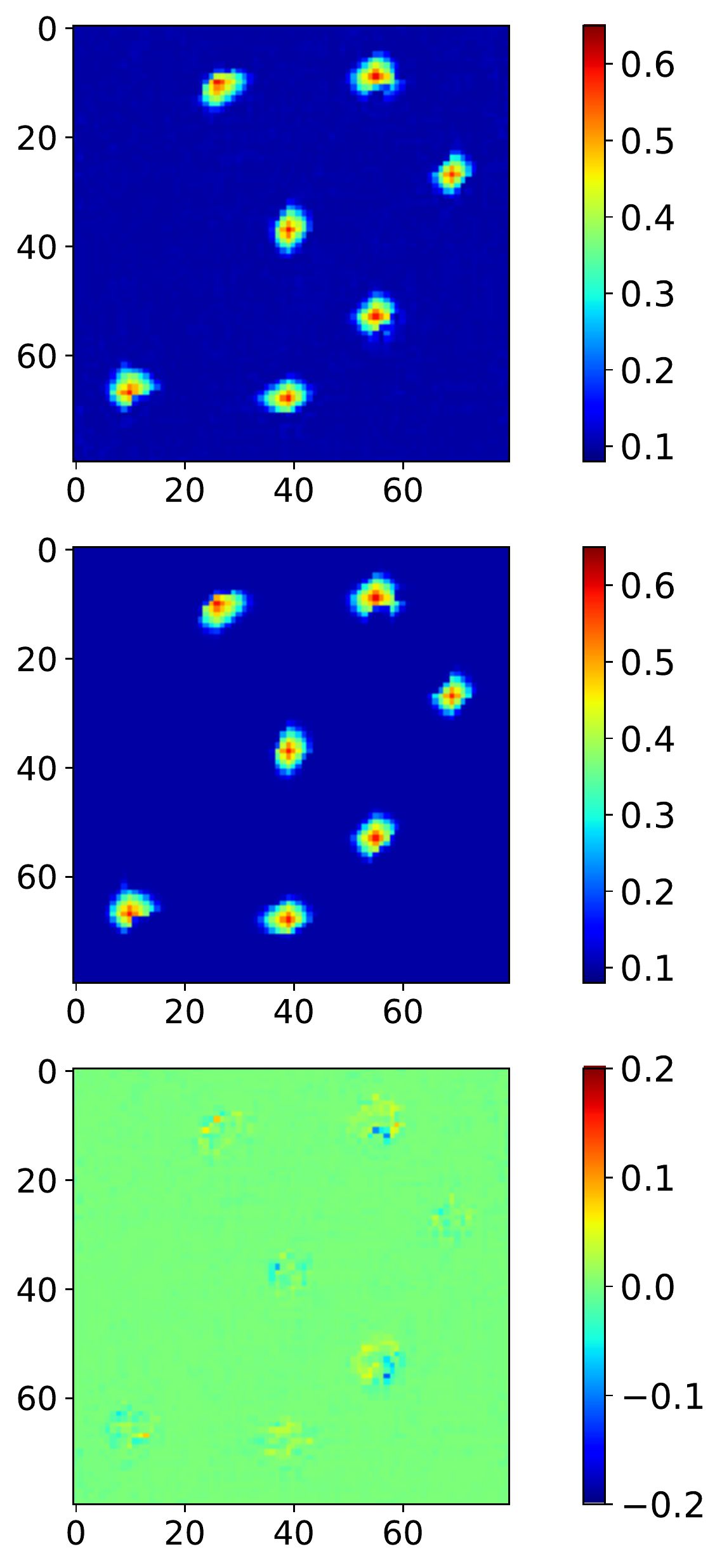}
         \caption{50~days}
         \label{fig:sat-map-single-step1}
     \end{subfigure}
     \begin{subfigure}[b]{0.325\textwidth}
         \centering
         \includegraphics[width=\textwidth]{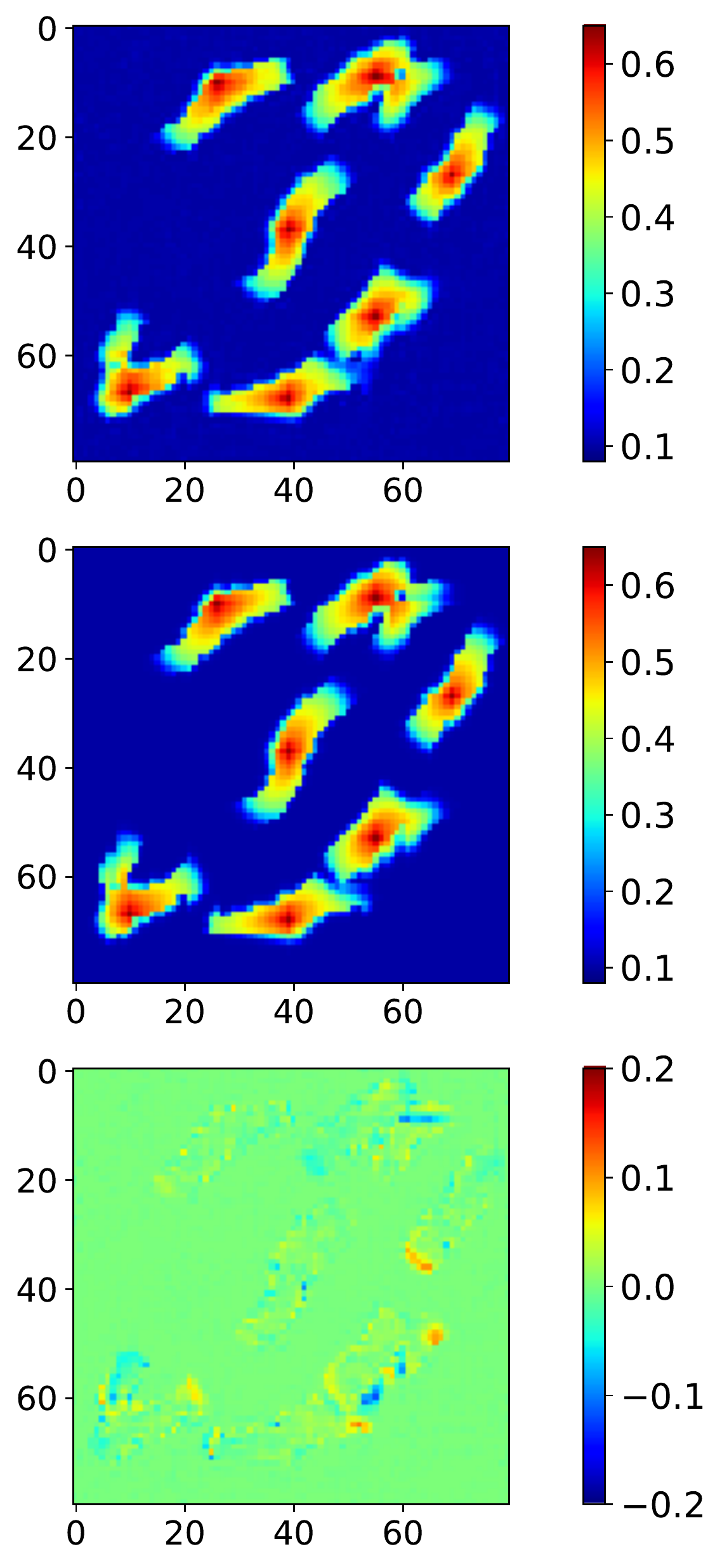}
         \caption{400~days}
         \label{fig:sat-map-single-step4}
     \end{subfigure}
     \begin{subfigure}[b]{0.325\textwidth}
         \centering
         \includegraphics[width=\textwidth]{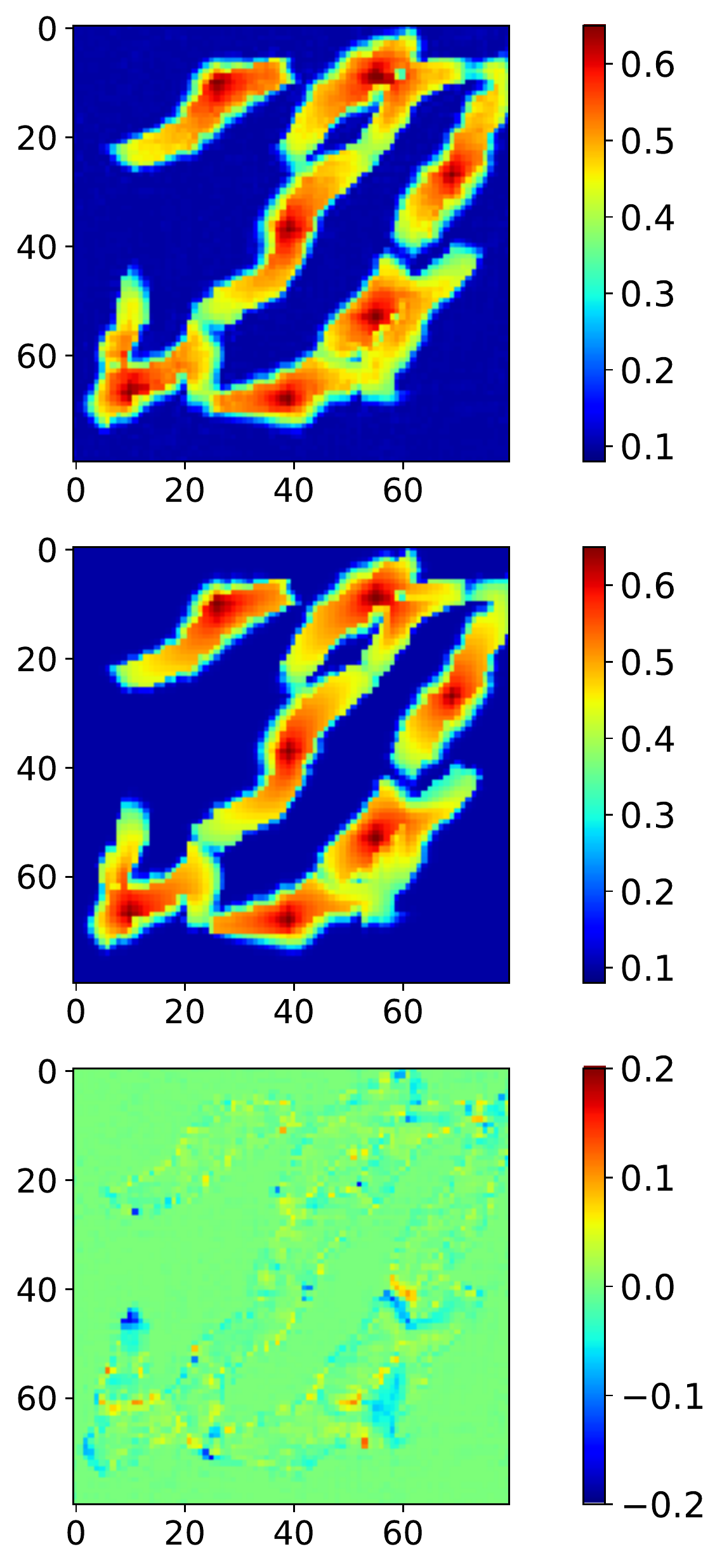}
         \caption{850~days}
         \label{fig:sat-map-single-step9}
     \end{subfigure}
    \caption{Saturation maps from recurrent R-U-Net surrogate model (top row) and high-fidelity simulator (middle row), along with difference maps (bottom row), at three different times.}
    \label{fig:sat-map-single}
\end{figure}

\begin{figure}
     \centering
     \begin{subfigure}[b]{0.325\textwidth}
         \centering
         \includegraphics[width=\textwidth]{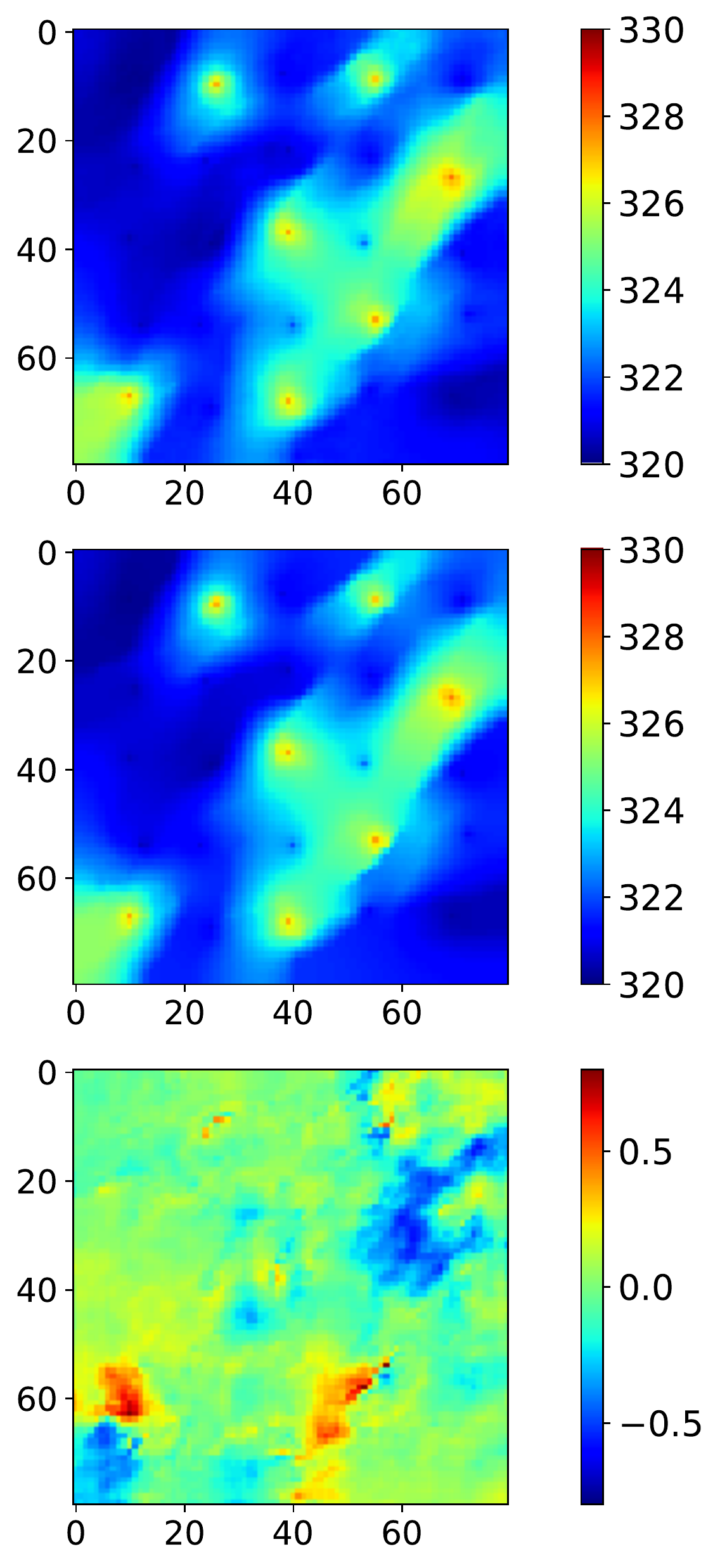}
         \caption{50~days}
         \label{fig:pressure-map-single-step1}
     \end{subfigure}
     \begin{subfigure}[b]{0.325\textwidth}
         \centering
         \includegraphics[width=\textwidth]{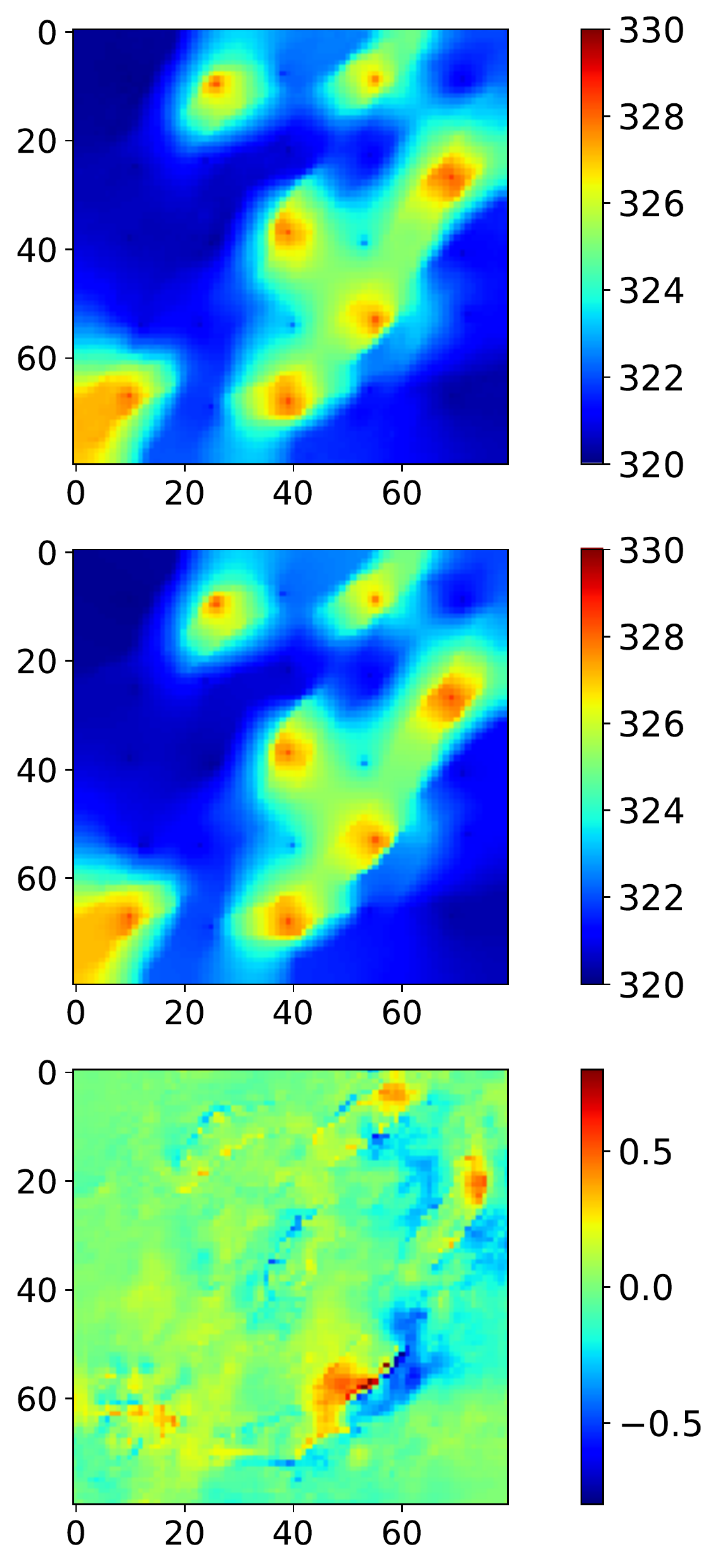}
         \caption{400~days}
         \label{fig:pressure-map-single-step4}
     \end{subfigure}
     \begin{subfigure}[b]{0.325\textwidth}
         \centering
         \includegraphics[width=\textwidth]{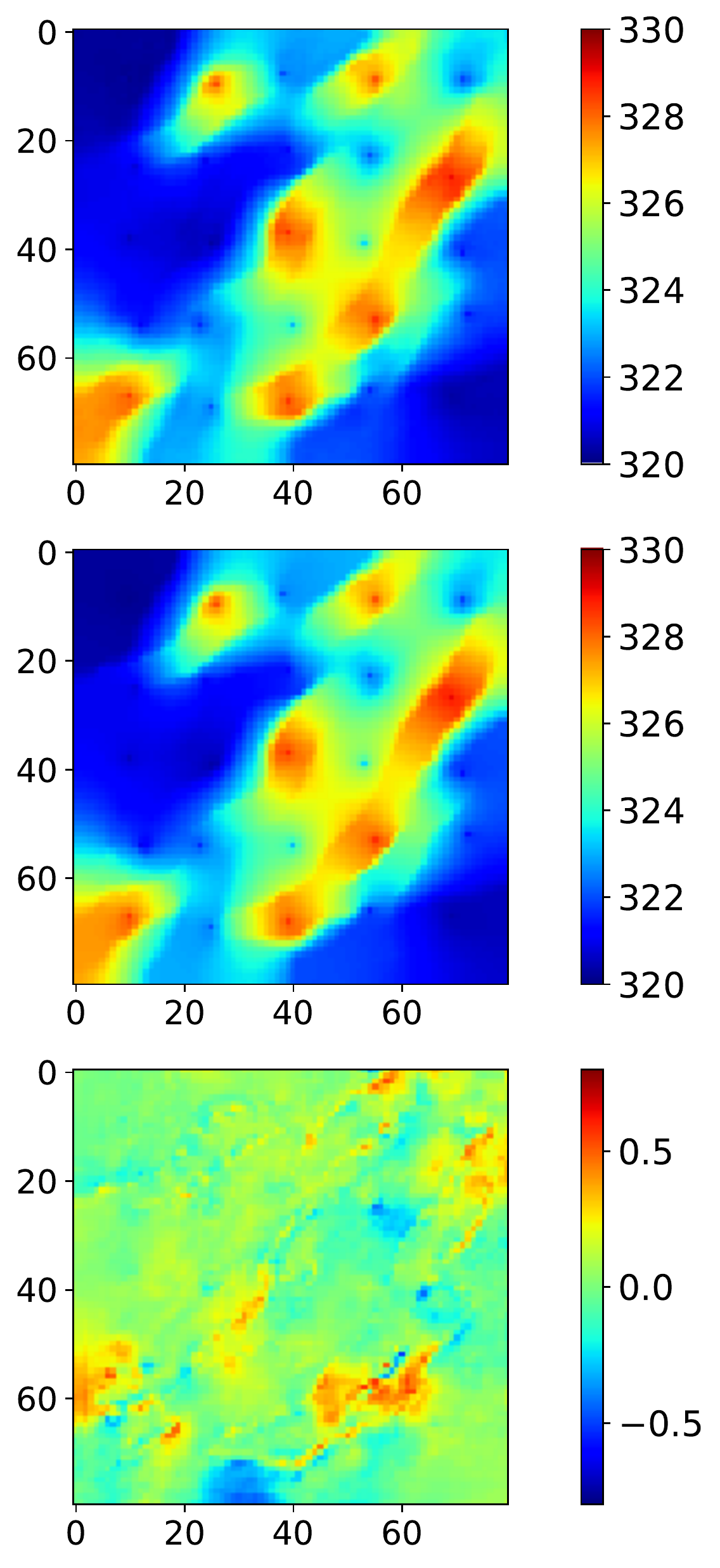}
         \caption{850~days}
         \label{fig:pressure-map-single-step9}
     \end{subfigure}
    \caption{Pressure maps from recurrent R-U-Net surrogate model (top row) and high-fidelity simulator (middle row), along with difference maps (bottom row), at three different times.}
    \label{fig:pressure-map-single}
\end{figure}


The saturation and pressure predictions provided by the recurrent R-U-Nets, for the geomodel in Fig.~\ref{fig:cnn-pca-reals}a, are displayed in Figs.~\ref{fig:sat-map-single} and \ref{fig:pressure-map-single}. In both plots the top row shows the recurrent R-U-Net surrogate model predictions, the middle row displays the high-fidelity  simulation (AD-GPRS) results, and the bottom row shows difference maps between the AD-GPRS and surrogate results. It is evident in Fig.~\ref{fig:sat-map-single} that the progression of the saturation field with time is strongly impacted by the channelized permeability field, with transport occurring along the high-permeability channels. This type of saturation distribution is quite different from that typically observed with multi-Gaussian permeability fields. We see that the recurrent R-U-Net provides generally accurate predictions for the saturation field at the three times. Small errors are, however, noticeable near saturation fronts. The pressure results in Fig.~\ref{fig:pressure-map-single} similarly demonstrate the high degree of accuracy of the recurrent R-U-Net for this case. Surrogate model error for pressure is small, though unlike saturation error, pressure error is not confined to fluid fronts.

It is useful to quantify the relative error in the recurrent R-U-Net predictions for saturation and pressure. The relative error in saturation, $\delta_S$, for the full set of $n_{e} = 500$ test samples, is given by
\begin{equation}
\delta_S = \frac{1}{n_{e}n_bn_{t}}\sum_{i=1}^{n_{e}}\sum_{j=1}^{n_b}\sum_{t=1}^{n_t} \frac{\norm{S_{i,j,t}^{surr} - S_{i,j,t}^{sim}}}{S_{i,j,t}^{sim}},
\end{equation}
where $S_{i,j,t}^{surr}$ and $S_{i,j,t}^{sim}$ denote the saturation value produced by the surrogate model and the simulator for test sample $i$, in grid block $j$, at time step $t$. Here we have $n_b = 80 \times 80= 6400$ grid blocks and $n_t = 10$ time steps. The initial water saturation is 0.1, which means $S_{i,j,t}^{sim} \geq 0.1$ (so the the denominator is well behaved). Evaluated over the 500 random test samples, we find $\delta_S = 2.8\%$, which indicates a high degree of accuracy in the surrogate dynamic saturation field. We note that the corresponding relative saturation error for the geomodel in Fig.~\ref{fig:cnn-pca-reals}a (over all 10 time steps) is 3.0\%. 

The relative pressure error, $\delta_p$, is given by
\begin{equation}
\delta_p = \frac{1}{n_{e}n_bn_{t}}\sum_{i=1}^{n_{e}}\sum_{j=1}^{n_b}\sum_{t=1}^{n_t} \frac{\norm{p_{i,j,t}^{surr} - p_{i,j,t}^{sim}}}{p_{max}^{sim} - p_{min}^{sim}},
\end{equation}
where $p_{i,j,t}^{surr}$ and $p_{i,j,t}^{sim}$ are defined analogously to $S_{i,j,t}^{surr}$ and $S_{i,j,t}^{sim}$, and difference between the maximum pressure $p_{max}^{sim}$ and minimum pressure $p_{min}^{sim}$ is used to normalize the error. Evaluated over the same 500 random test samples, we find $\delta_p = 1.2\%$, which again indicates high accuracy.
The relative pressure error for the geomodel in Fig.~\ref{fig:cnn-pca-reals}a (over all 10 time steps) is 1.4\%.




\subsection{Well flow rate predictions}
\label{sect:well_pred}

Well production and injection rates are often key quantities of interest, and data of this type are commonly assimilated in history matching studies. We now assess the accuracy of our recurrent R-U-Nets for well-rate data. This entails the application of Eq.~\ref{eq:well_flow}, which in turn requires the R-U-Net estimates for $p$ and $S_w$ in well blocks, along with the well index given in Eq.~\ref{eq:wi}.

\begin{figure}
     \centering
     \begin{subfigure}[b]{0.45\textwidth}
         \centering
         \includegraphics[width=\textwidth]{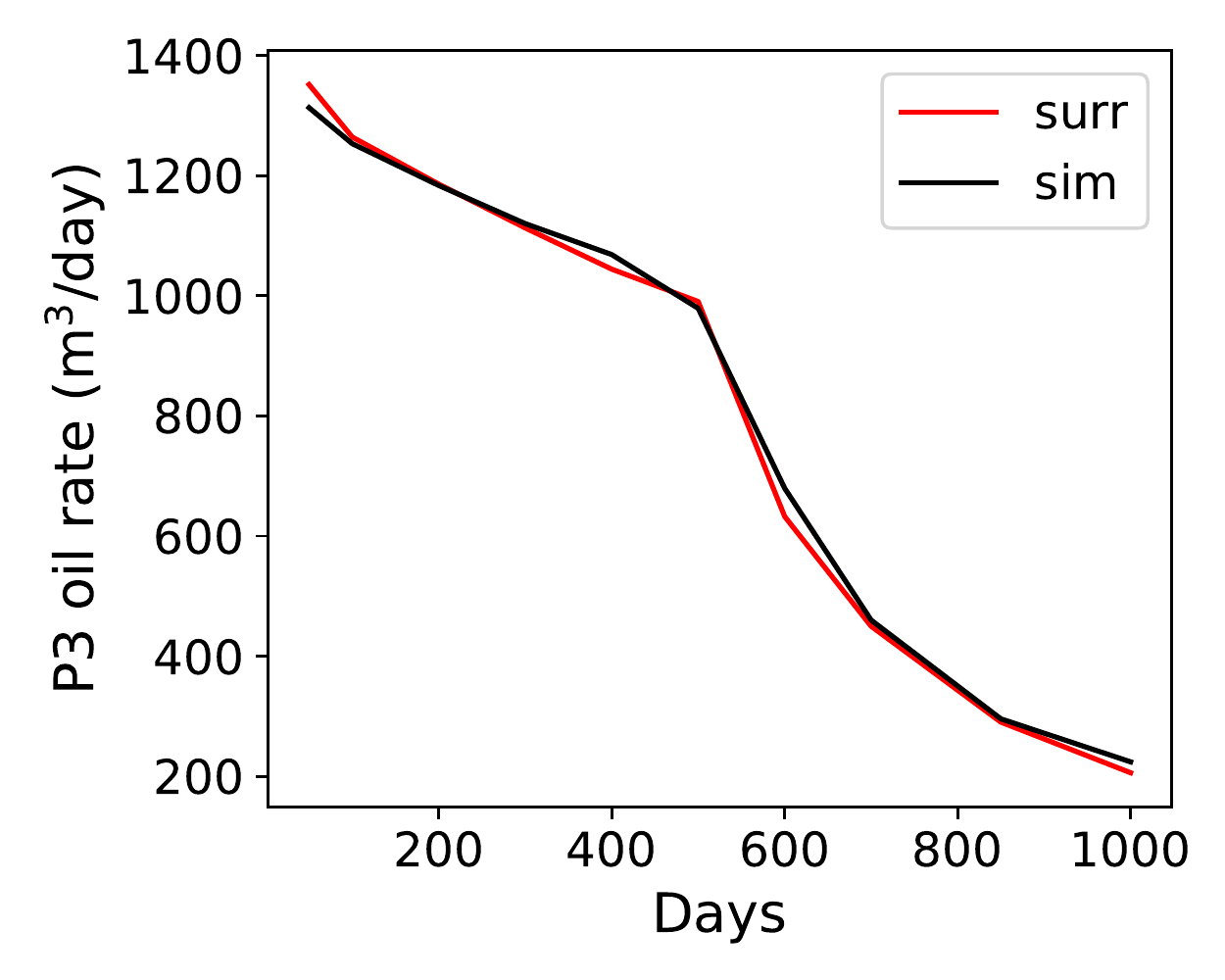}
         \caption{P3 oil rate}
         \label{pfs-case3-orate-w3}
     \end{subfigure}
     \begin{subfigure}[b]{0.45\textwidth}
         \centering
         \includegraphics[width=\textwidth]{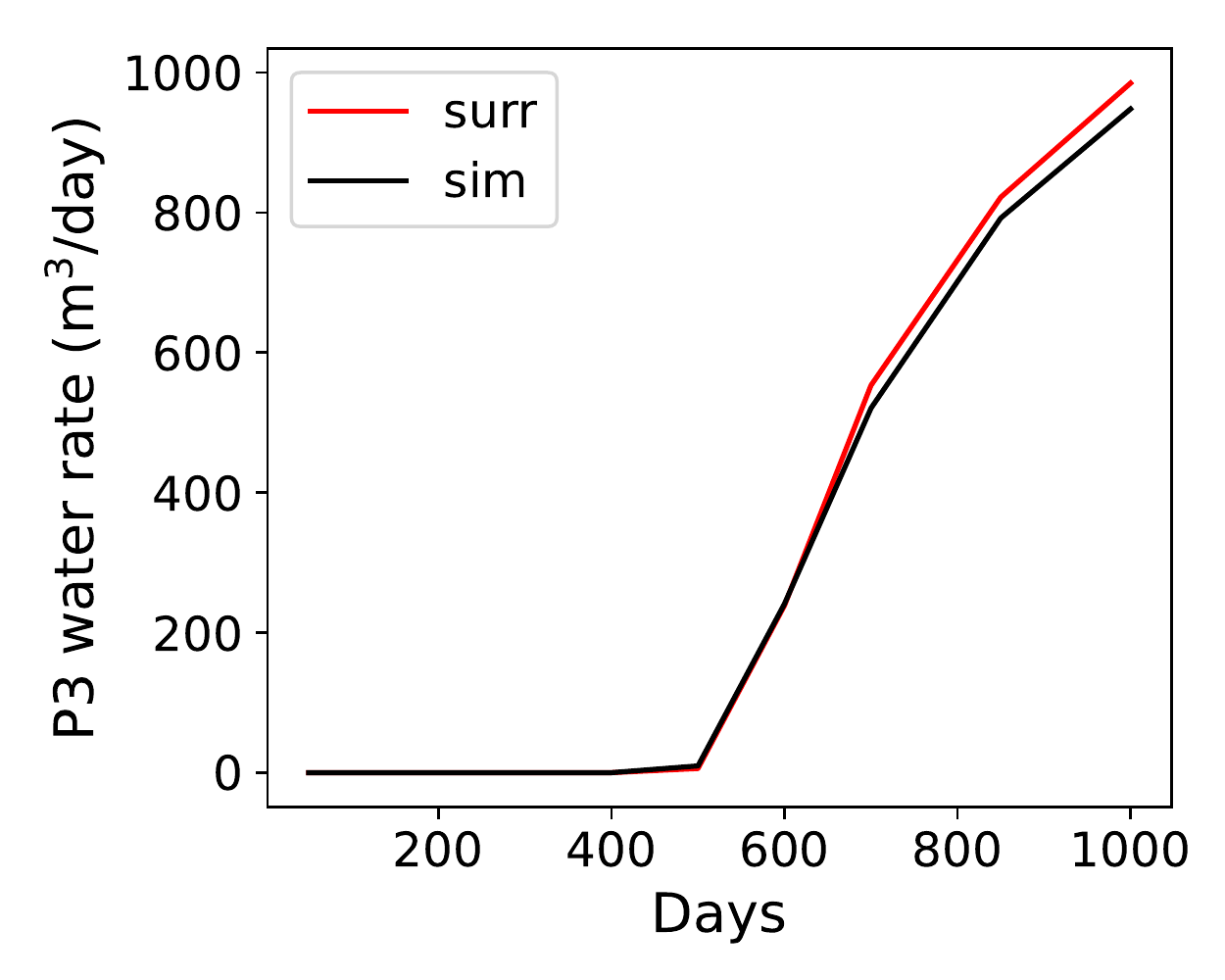}
         \caption{P3 water rate}
         \label{pfs-case3-wrate-w3}
     \end{subfigure}
     
     \begin{subfigure}[b]{0.45\textwidth}
         \centering
         \includegraphics[width=\textwidth]{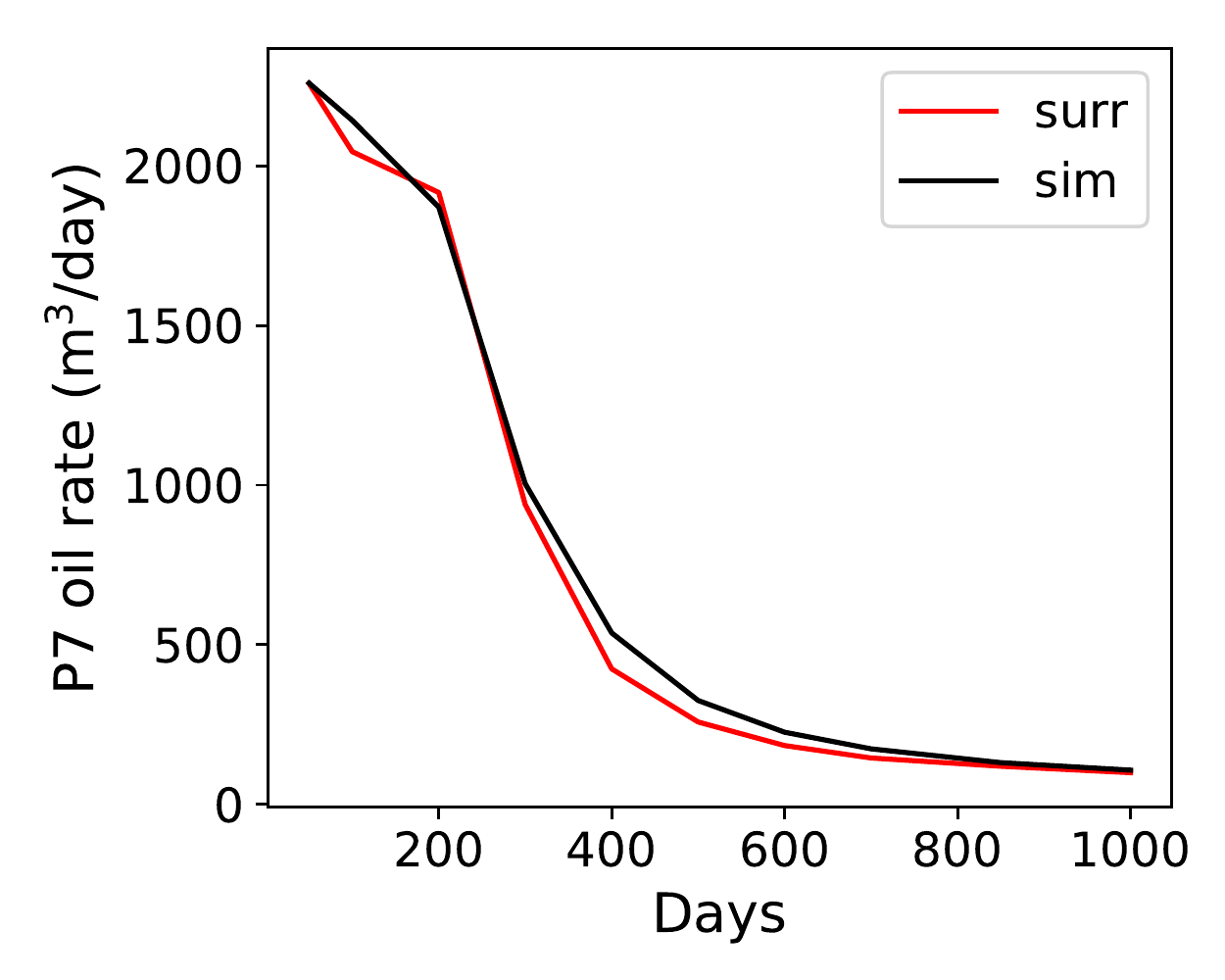}
         \caption{P7 oil rate}
         \label{pfs-case3-orate-w5}
     \end{subfigure}
     \begin{subfigure}[b]{0.45\textwidth}
         \centering
         \includegraphics[width=\textwidth]{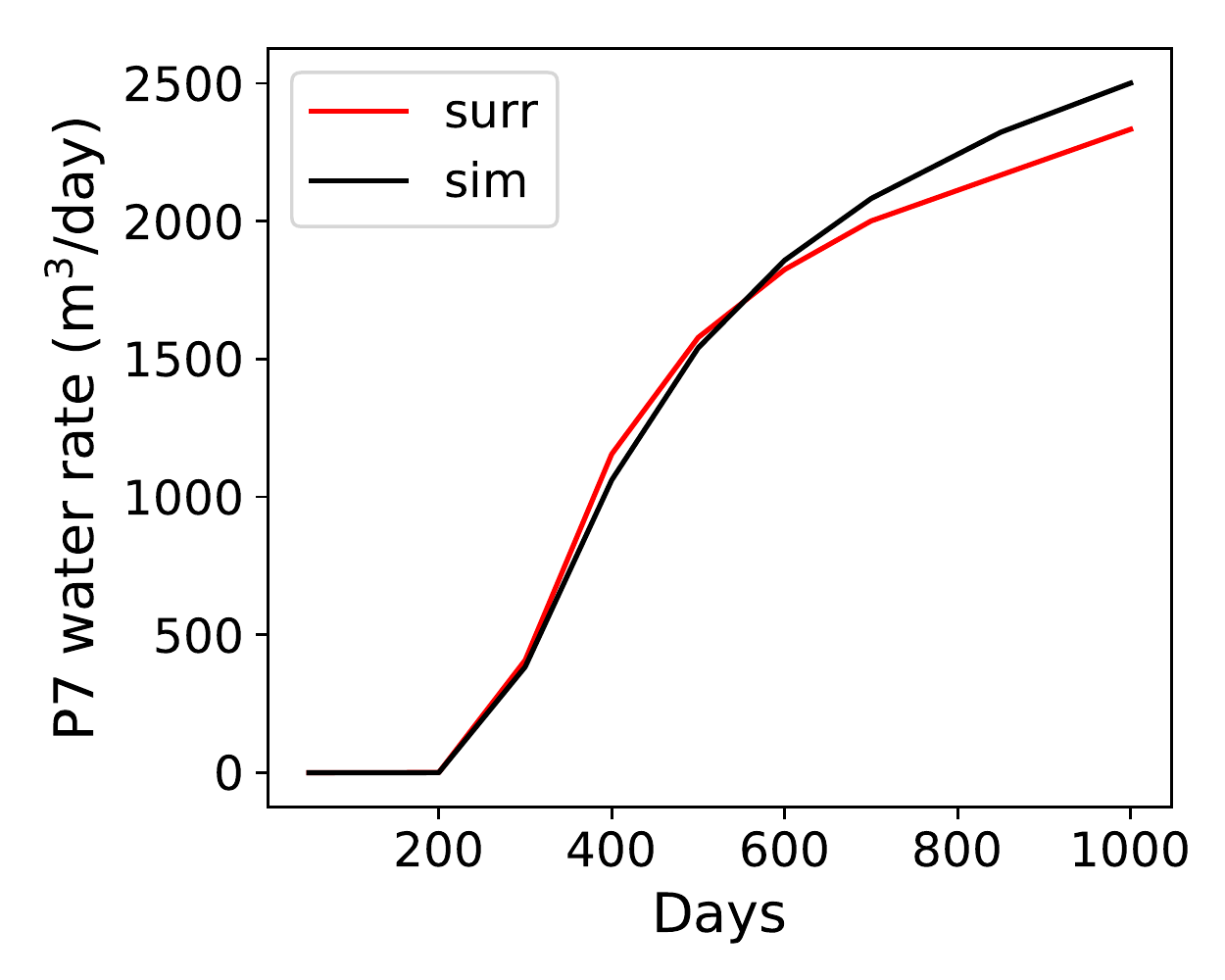}
         \caption{P7 water rate}
         \label{pfs-case3-wrate-w5}
     \end{subfigure}
     
    \begin{subfigure}[b]{0.45\textwidth}
         \centering
         \includegraphics[width=\textwidth]{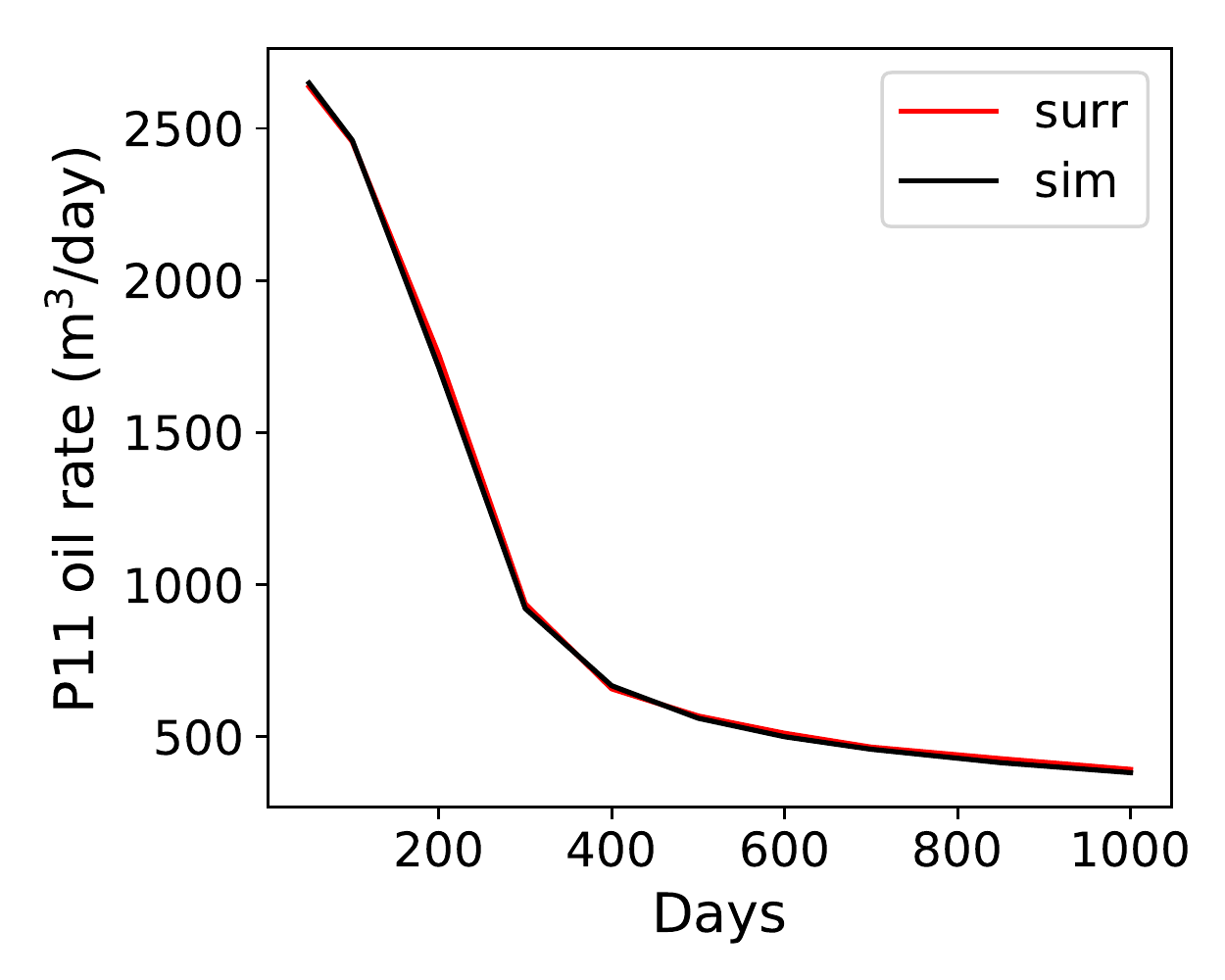}
         \caption{P11 oil rate}
         \label{pfs-case3-orate-w11}
     \end{subfigure}
          \begin{subfigure}[b]{0.45\textwidth}
         \centering
         \includegraphics[width=\textwidth]{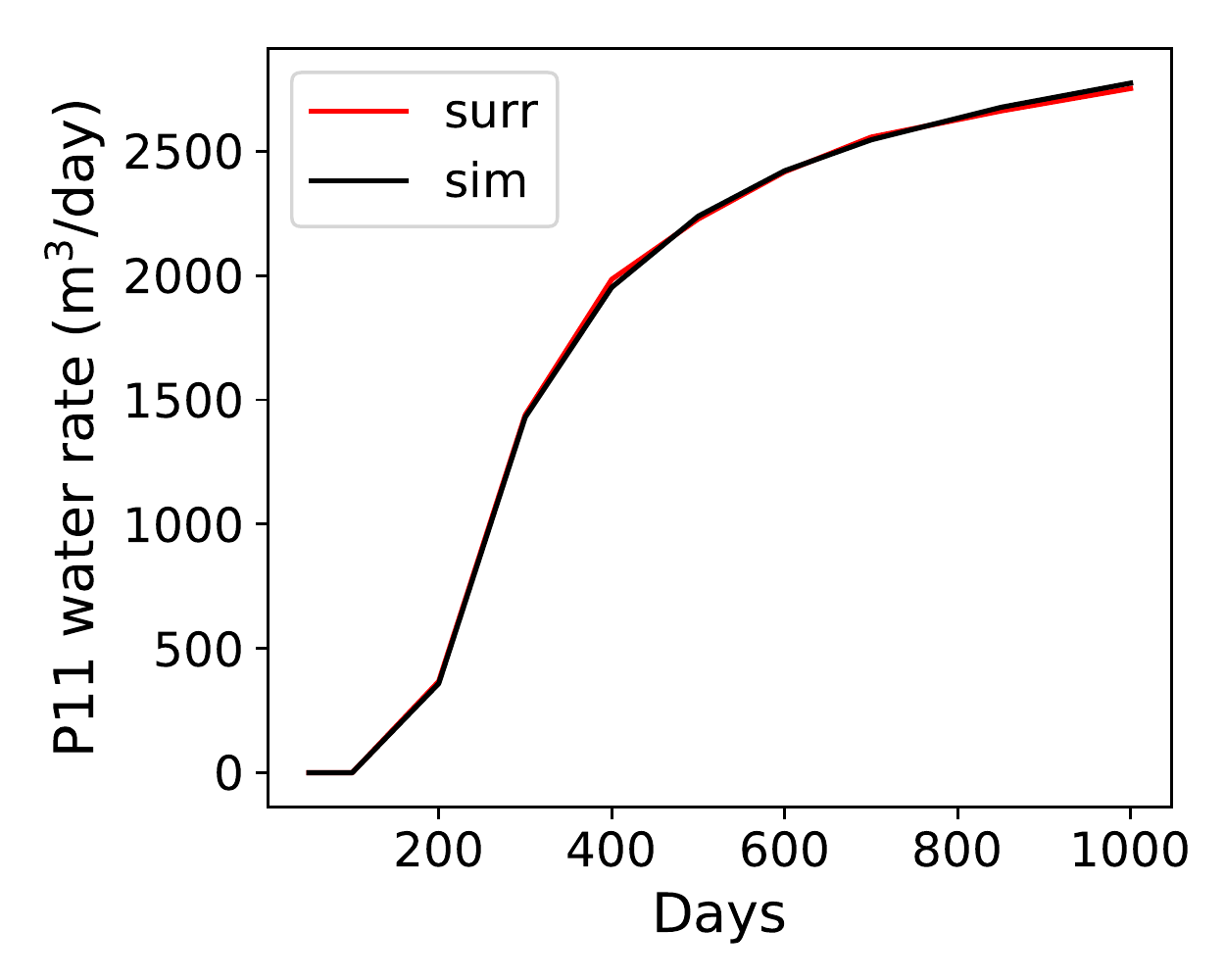}
         \caption{P11 water rate}
         \label{pfs-case3-wrate-w11}
     \end{subfigure}
    
    \caption{Comparison of oil (left) and water (right) production rates, for three different wells, for the geomodel considered in Section~\ref{sect:map_pred}. Red and black curves represent results from the recurrent R-U-Net surrogate model and the AD-GPRS high-fidelity simulator, respectively.}
    \label{fig:producer-flow-single-case}
\end{figure}

Fig.~\ref{fig:producer-flow-single-case} displays comparisons of oil and water production rates for the geomodel considered in Section~\ref{sect:map_pred} (shown in Fig.~\ref{fig:cnn-pca-reals}a). Well locations are as indicated on Fig.~\ref{fig:chap3-model-set-up}. Well rate results in Fig.~\ref{fig:producer-flow-single-case} are presented for the high-fidelity AD-GPRS run (black curves, designated `sim' in the figure legends) and for the recurrent R-U-Net predictions (red curves, designated `surr'), for three production wells. These rates are actually computed at only 10 particular times, but the results are presented as continuous curves. These results demonstrate that the surrogate model provides a high degree of accuracy in well flow rates, which is essential if this model is to be used for history matching. There are, however, small but noticeable discrepancies in water rate at late time.

In order to concisely evaluate recurrent R-U-Net performance for the full set of 500 test cases, we now present results, in terms of flow statistics, for the full ensemble. Specifically, for a given well (or for the entire field), we order the 500 oil and water production rate results at each time step, and then select the result corresponding to the 10th, 50th, and 90th percentile for each quantity. These are referred to as the P10, P50 and P90 responses. This type of assessment was also used in the evaluation of CNN-PCA for geomodel generation -- see \citep{liu2019deep} for details.

\begin{figure}
     \centering
     \begin{subfigure}[b]{0.45\textwidth}
         \centering
         \includegraphics[width=\textwidth]{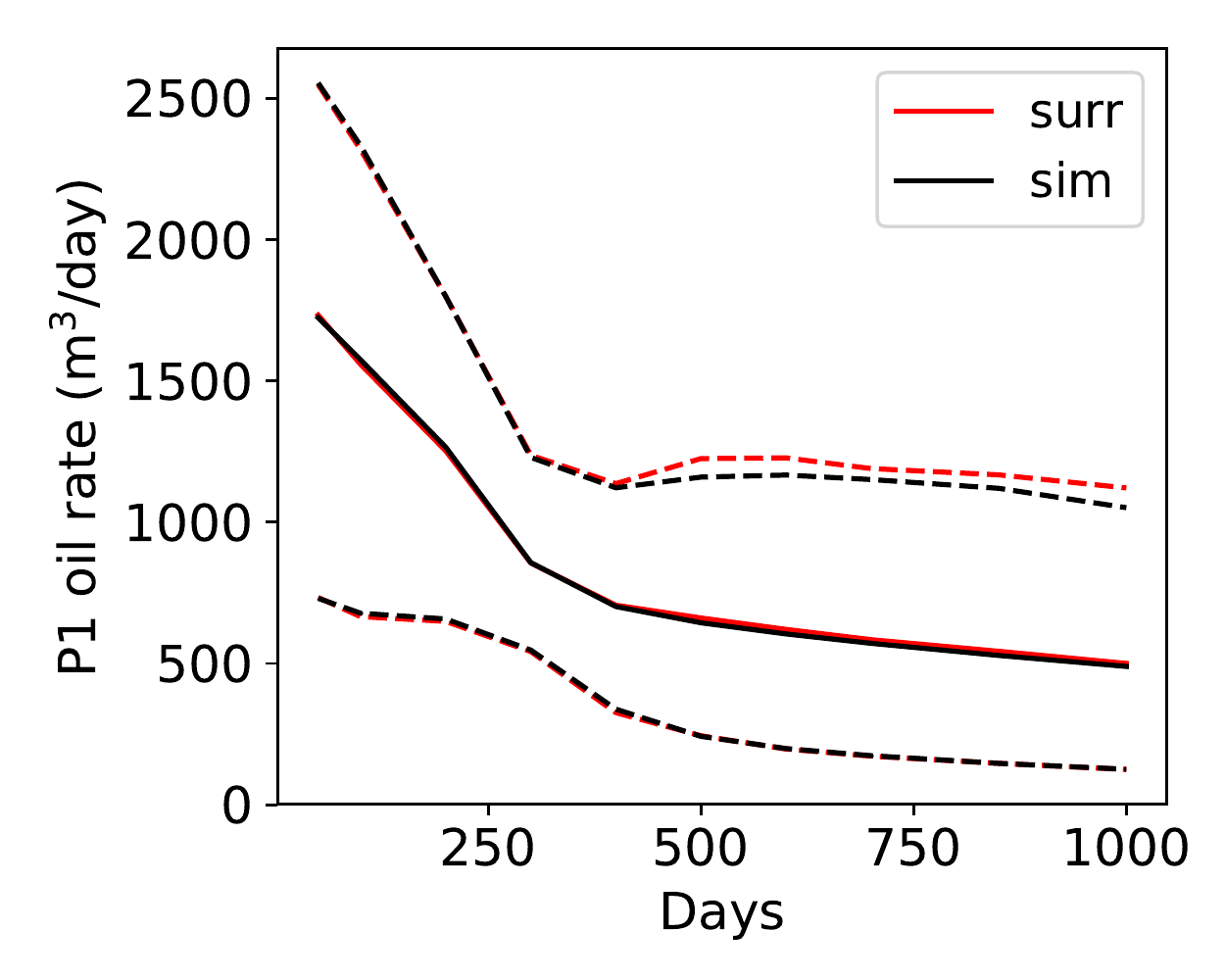}
         \caption{P1 oil rate}
         \label{pfs-orate-w1}
     \end{subfigure}
     \begin{subfigure}[b]{0.45\textwidth}
         \centering
         \includegraphics[width=\textwidth]{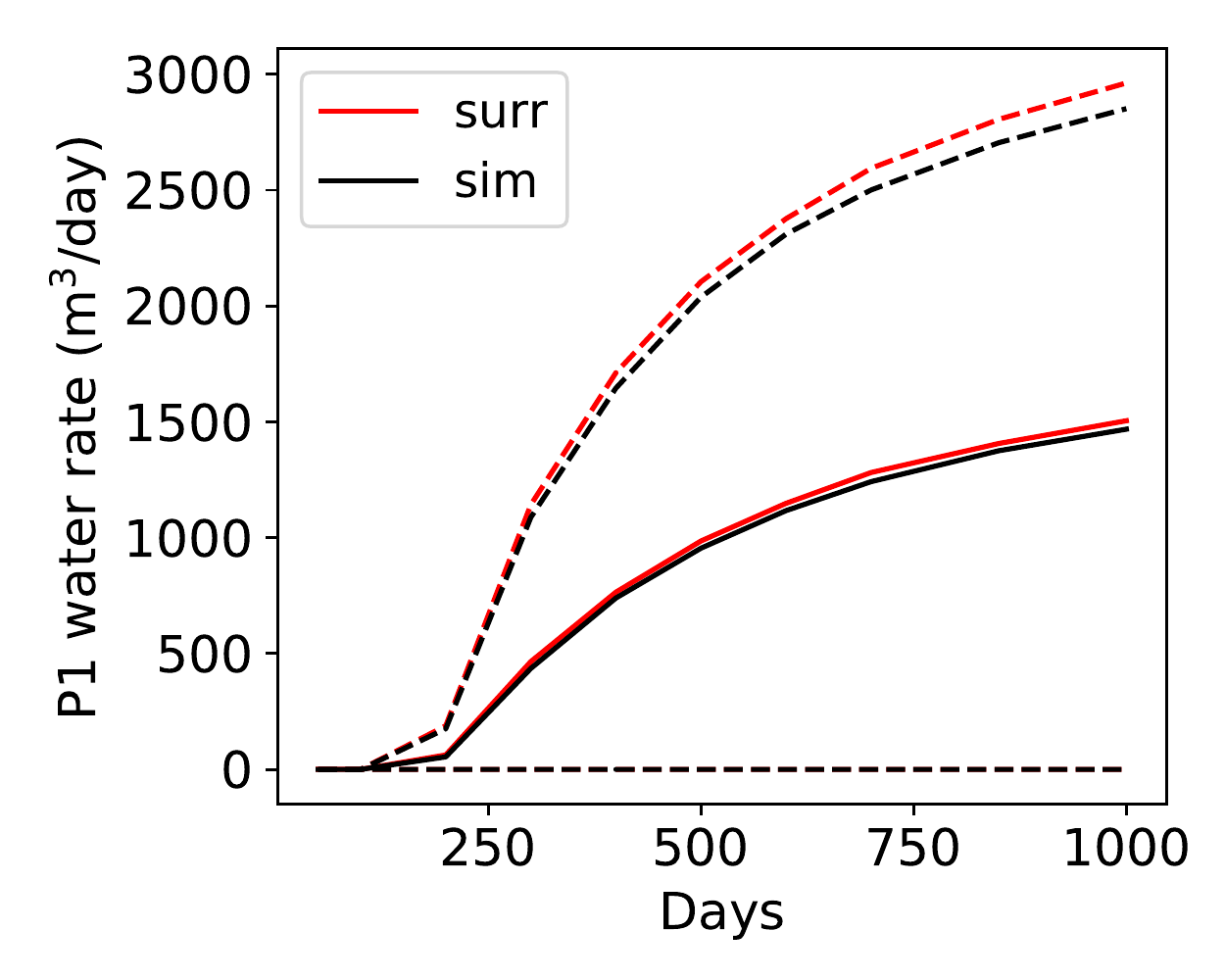}
         \caption{P1 water rate}
         \label{pfs-wrate-w1}
     \end{subfigure}
     
     \begin{subfigure}[b]{0.45\textwidth}
         \centering
         \includegraphics[width=\textwidth]{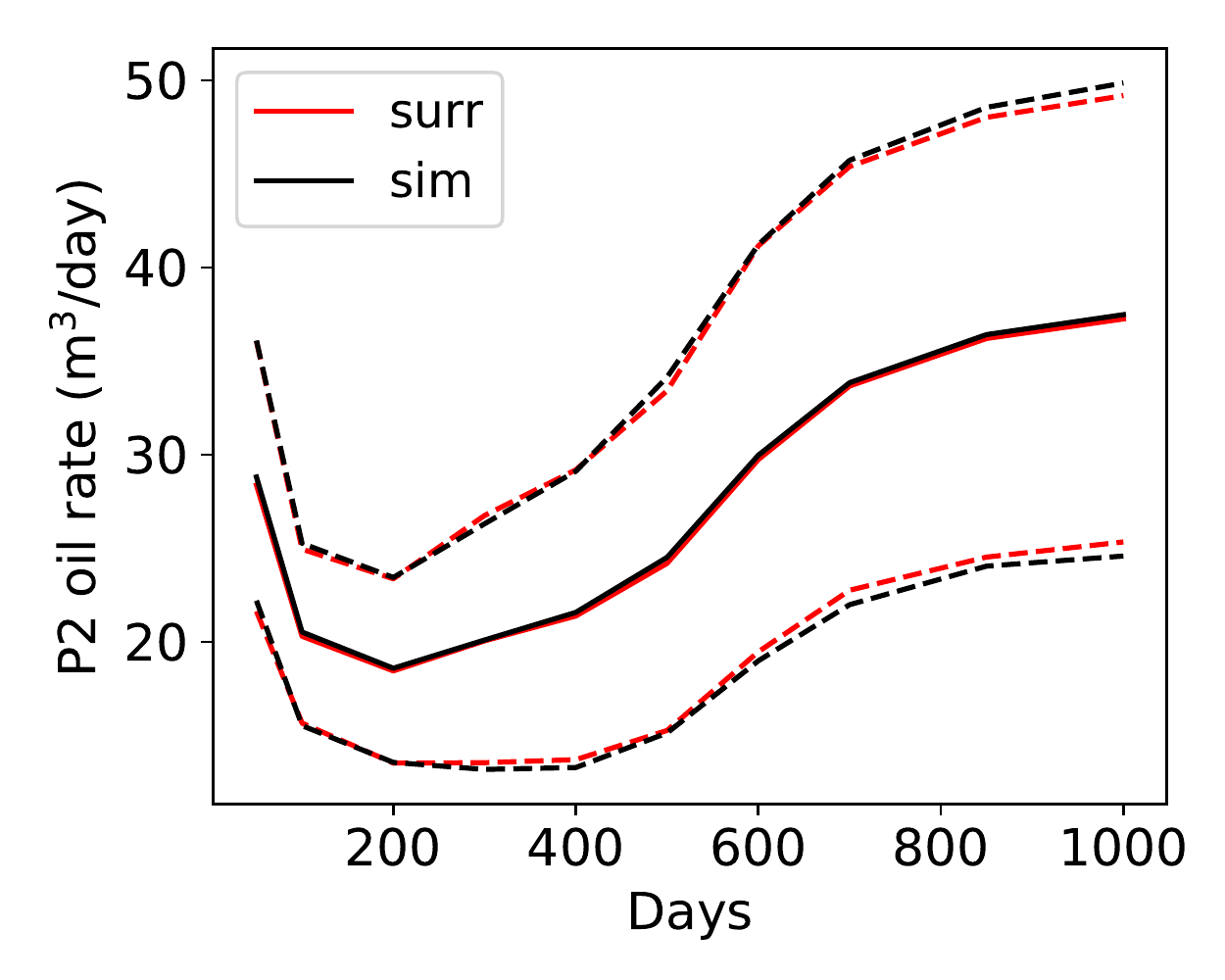}
         \caption{P2 oil rate}
         \label{pfs-orate-w2}
     \end{subfigure}
     \begin{subfigure}[b]{0.45\textwidth}
         \centering
         \includegraphics[width=\textwidth]{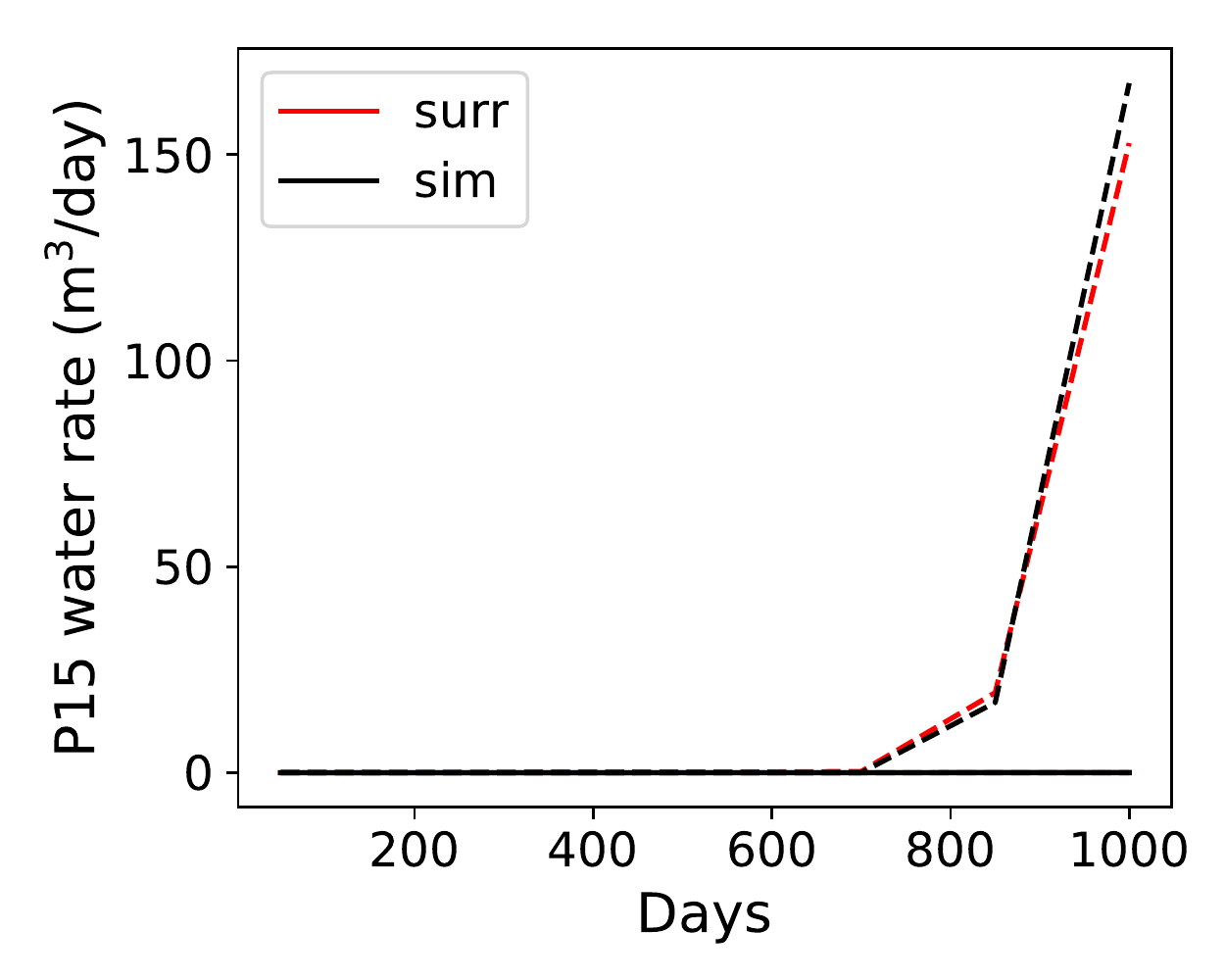}
         \caption{P15 water rate}
         \label{pfs-wrate-w15}
     \end{subfigure}
     
    \begin{subfigure}[b]{0.45\textwidth}
         \centering
         \includegraphics[width=\textwidth]{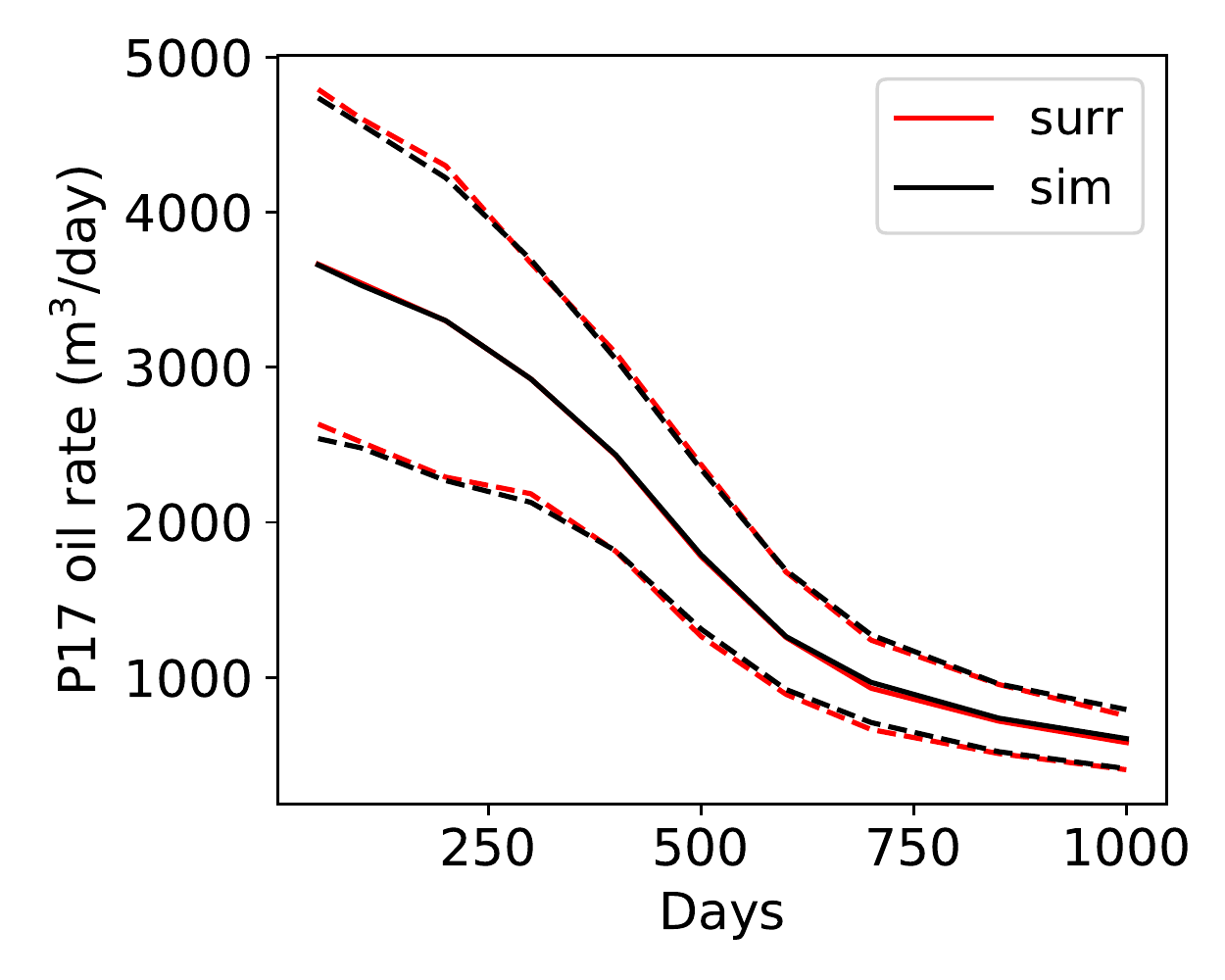}
         \caption{P17 oil rate}
         \label{pfs-orate-w17}
     \end{subfigure}
          \begin{subfigure}[b]{0.45\textwidth}
         \centering
         \includegraphics[width=\textwidth]{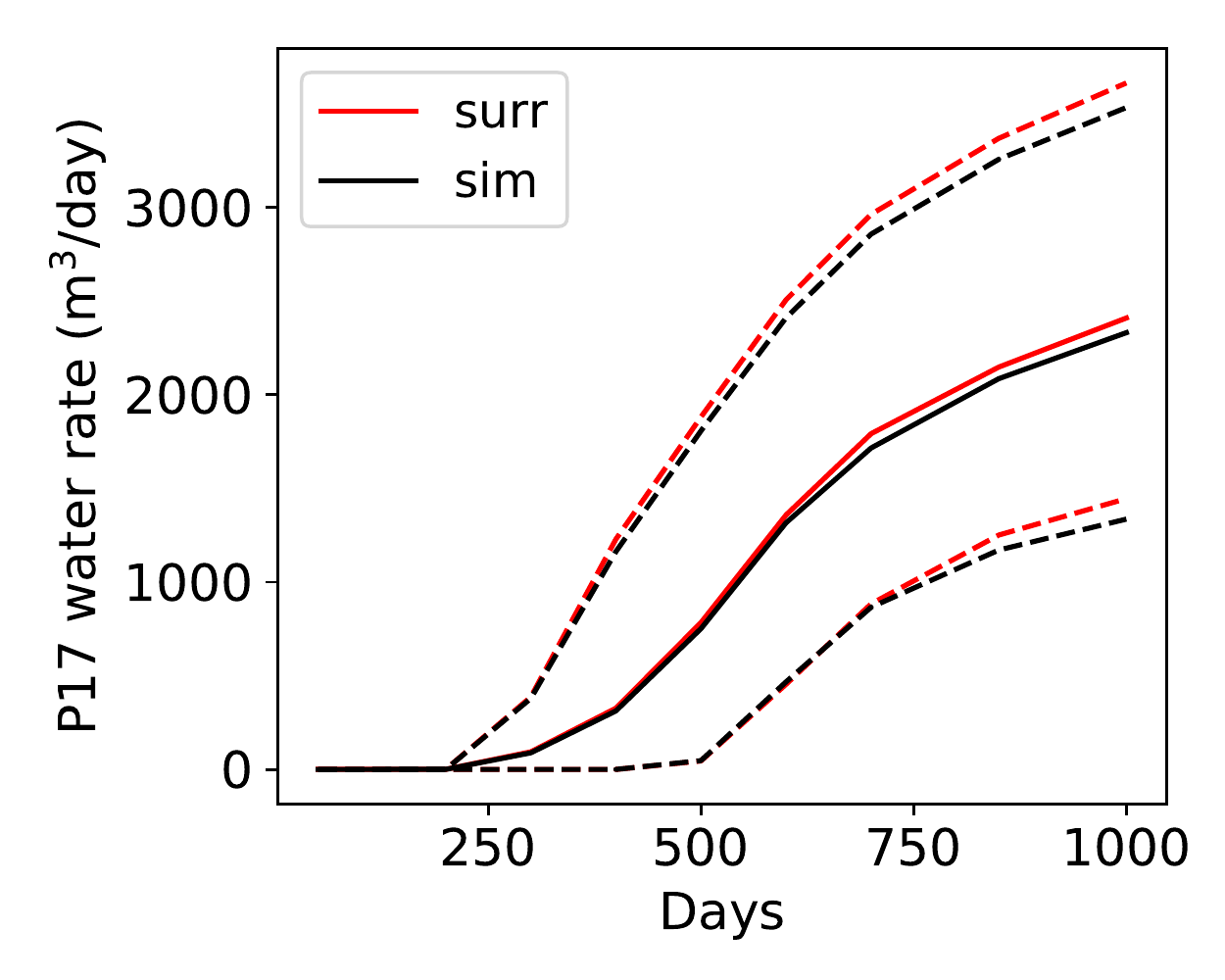}
         \caption{P17 water rate}
         \label{pfs-wrate-w17}
     \end{subfigure}
    
    \caption{Comparison of oil (left) and water (right) production rate statistics, for different wells, over the full ensemble of 500 test cases. Red and black curves represent results from the recurrent R-U-Net surrogate model and the AD-GPRS high-fidelity simulator, respectively. Solid curves correspond to P50 results, lower and upper dashed curves to P10 and P90 results. P10, P50 and P90 water production rates for well~P2 are all zero (thus results are shown for well~P15 instead).}
    \label{fig:producer-flow-statistics}
\end{figure}

Fig.~\ref{fig:producer-flow-statistics} displays the P10, P50 and P90 flow responses, for oil and water production rate, obtained from the surrogate model (red curves) and from AD-GPRS (black curves). Oil rate results for production wells P1, P2 and P17 are shown, while water rates for wells P1, P15 and P17 are displayed. We consider P2 for oil and P15 for water in order to display a range of behaviors. The P50 results are shown as solid curves, while the P10 and P90 responses are shown as the (lower and upper) dashed curves. Note that the model corresponding to the P50 result (or the P10 or P90 result) can differ from time step to time step. Agreement is clearly very close between the two sets of results. 

Note that, for well~P2, the oil rates evident in Fig.~\ref{fig:producer-flow-statistics}c are very low compared to those for the other wells. This is the case because this well is located in mud (see Fig.~\ref{fig:chap3-model-set-up} for the well locations). For this well even the P90 result corresponds to zero water production, so we do not show well~P2 water rate results. We instead show the water production statistics for well~P15. For this well it is apparent from Fig.~\ref{fig:producer-flow-statistics}d that more than half of the realizations do not produce any water over the entire 1000-day simulation time frame. It is encouraging to see that the recurrent R-U-Net results are consistent with those from AD-GPRS even in these more extreme situations.

\begin{figure}
     \centering
     \begin{subfigure}[b]{0.45\textwidth}
         \centering
         \includegraphics[width=\textwidth]{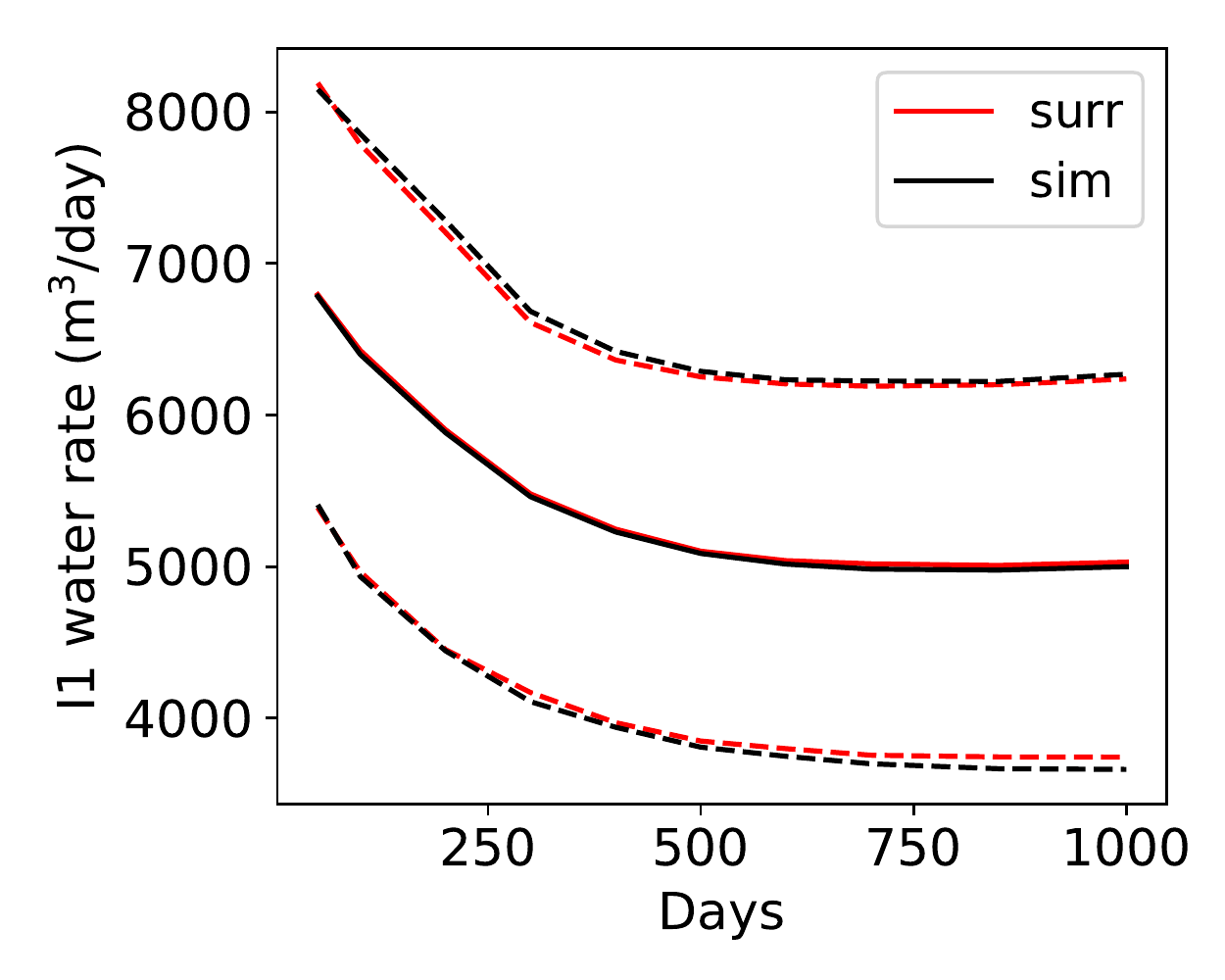}
         \caption{I1 water injection rate}
         \label{fifs-wrate1}
     \end{subfigure}
     \begin{subfigure}[b]{0.47\textwidth}
         \centering
         \includegraphics[width=\textwidth]{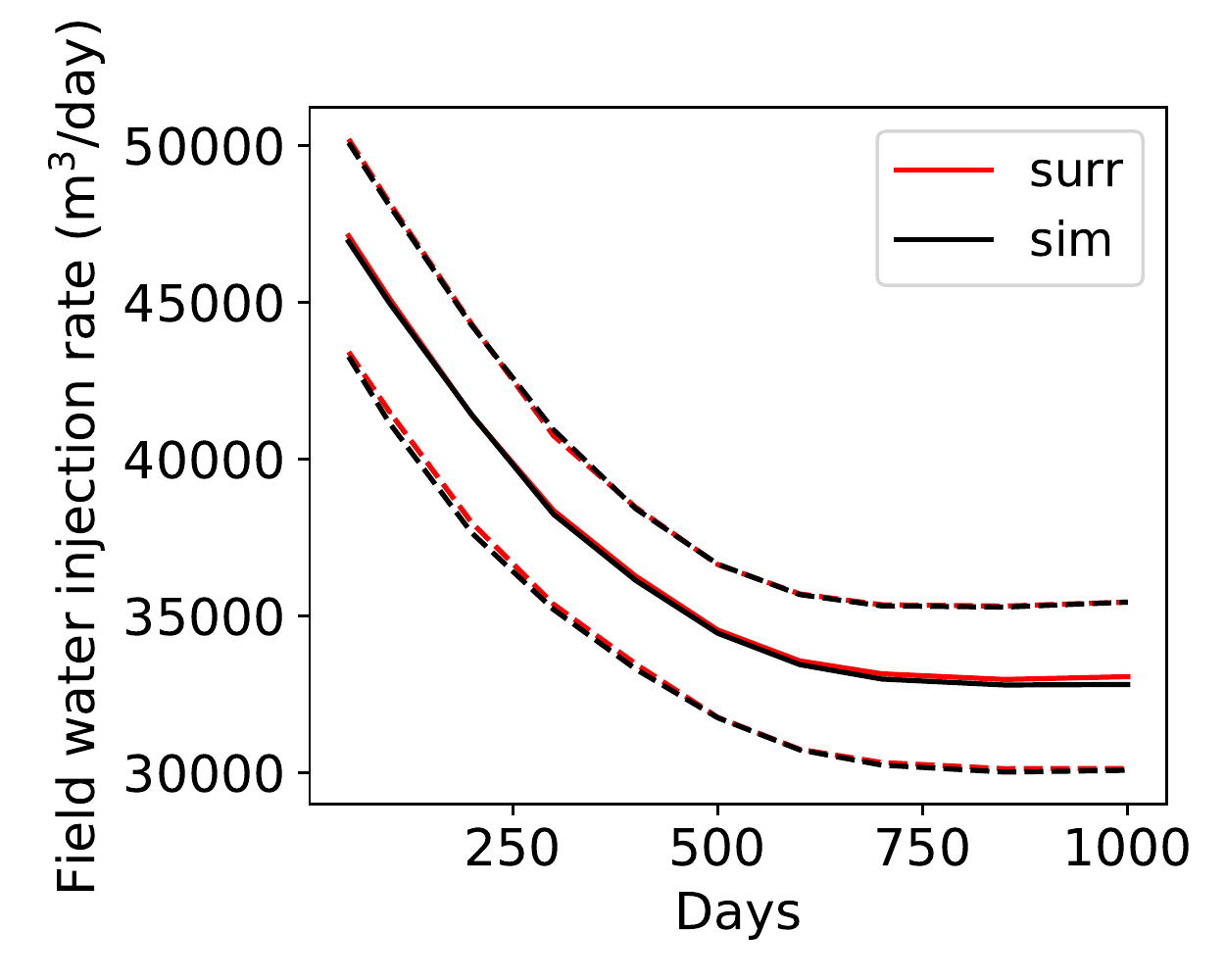}
         \caption{Field water injection rate}
         \label{fifs-inj-wrate}
     \end{subfigure}

    \caption{Comparison of water injection rate statistics for well~I1 (left) and for the entire field (right) over the full ensemble of 500 test cases. Red and black curves represent results from the recurrent R-U-Net surrogate model and the AD-GPRS high-fidelity simulator, respectively. Solid curves correspond to P50 results, lower and upper dashed curves to P10 and P90 results.}
    \label{fig:field-injector-flow-statistics}
\end{figure}

Although the well rate results presented thus far have been for production wells, it is important to demonstrate that the recurrent R-U-Net results for water injection are also accurate. Comparisons of water injection rate statistics between the surrogate model and the high-fidelity simulator, for well~I1 and for the entire field, are shown in Fig.~\ref{fig:field-injector-flow-statistics}. There we see very close agreement between the two sets of results, which demonstrates that R-U-Net predictions for water injection rate are indeed accurate.

Finally, we quantify the average relative errors in oil and water production rates, and water injection rate, over all production and injection wells, respectively. For oil and water production rates, error is given by:
\begin{equation}
\delta_{r,j} = \frac{1}{n_{e}n_{p}n_{t}}\sum_{i=1}^{n_{e}}\sum_{k=1}^{n_p}\sum_{t=1}^{n_t} \frac{\norm{r_{ i,k,t}^{surr,j} - r_{i,k,t}^{sim, j}}}{r_{i,k,t}^{sim, j} + \epsilon},
\label{eq:rate-rel-error}
\end{equation}
where $r_{i,k,t}^{surr, j}$ and $r_{i,k,t}^{sim, j}$ denote the phase ($j=o$ for oil and $w$ for water) production rate from the surrogate model and
the simulator for well $k$ at time step $t$ in test sample $i$. To avoid division by zero, a constant $\epsilon = 1$ is introduced in the denominator. Here $n_p$ is the number of production wells. For injection rate error, we replace $n_p$ by $n_i$, the number of injection wells, and compute $\delta_{r,inj}$. Note that there is no error cancellation from time step to time step in Eq.~\ref{eq:rate-rel-error}. The relative errors for oil and water production rates are $\delta_{r,o} = 6.4\%$ and $\delta_{r,w} = 5.8\%$, while the error for water injection rate is $\delta_{r,inj} = 3.5\%$. For the particular geomodel in Fig.~\ref{fig:cnn-pca-reals}a, these errors are comparable but slightly smaller (5.7\%, 5.4\%, 3.2\% for oil production, water production and water injection rate, respectively). In total, the recurrent R-U-Net well rate predictions are sufficiently accurate for use in history matching, as will be demonstrated in the next section.


\section{History Matching Using Deep-learning-based Surrogate Model}
\label{sect:hm}

The applicability of our surrogate model for estimating dynamic state evolution and well rates was demonstrated in Section~\ref{sect:flowstats}. We now apply the method for a challenging history matching problem.

\subsection{History matching procedure}
\label{sect:hm_method}

In this study, the randomized maximum likelihood (RML) procedure is applied to generate posterior geomodels. The RML method, originally developed in \citep{kitanidis1995quasi} and \citep{oliver1996multiple}, has been widely used both with high-dimensional geomodels (e.g., \citep{gao2005quantifying}) and parameterized models (e.g., \citep{vo2015data,liu2019deep}). Here we apply it with a CNN-PCA parameterization, so our treatment is analogous to that described in \citep{liu2019deep}. However, in place of high-fidelity simulations,  recurrent R-U-Net predictions are used. This means we (1) represent a geomodel $\mathbf{m}$ as $\mathbf{m} \approx \mathbf{m}_{cnn}(\boldsymbol\xi)$, where $\boldsymbol\xi \in \mathbb{R}^{n_\xi}$ is the low-dimensional (parameterization) variable, and (2) model the flow response as $f \approx \hat{f}(\mathbf{m}_{cnn}(\boldsymbol\xi))$, where the use of $\hat{f}$ (rather than $f$) means the recurrent R-U-Net is used in place of the numerical simulator.

RML is an optimization-based procedure in which a minimization problem is solved repeatedly to generate multiple ($N_r$) posterior samples. In our setting, each run provides a posterior sample designated $\mathbf{\boldsymbol\xi}_{i,rml}$. The minimization problem is expressed as:
\begin{equation}
        \begin{split}
            \mathbf{\boldsymbol\xi}_{i,rml} =& \argmin_{\boldsymbol\xi_i}[(\hat{f}(\mathbf{m}_{cnn}(\boldsymbol\xi_i)) - \mathbf{d}_{i, obs}^*)^{\intercal}C_D^{-1}(\hat{f}(\mathbf{m}_{cnn}(\boldsymbol\xi_i)) - \mathbf{d}_{i, obs}^*)\\
            &+ (\boldsymbol\xi_i - \boldsymbol\xi_i^*)^{\intercal}(\boldsymbol\xi_i - \boldsymbol\xi_i^*)], \ \ \ i=1, \ldots, N_r.
        \end{split}
        \label{eq:rml}
\end{equation}
Here $\mathbf{d}_{obs}$ denotes the observed data and $C_D$ is the covariance of the data measurement error. The superscript $*$ indicates that the quantity is sampled from a normal distribution. Specifically, the (perturbed) observation data $\mathbf{d}_{i, obs}^*$ is sampled from $\mathcal{N}(\mathbf{d}_{obs}, C_D)$, and $\mathbf{\boldsymbol\xi}_i^*$ is sampled from $\mathcal{N}(0, I)$. The observed and simulated data in our case include oil and water production rates and water injection rates. Though not considered here, time-lapse seismic data are available in some cases, and these data can be used to infer an approximate global saturation field. In this situation, the global saturation field provided by the recurrent R-U-Net would also enter into the formulation. We note finally that the $(\boldsymbol\xi_i - \boldsymbol\xi_i^*)^{\intercal}(\boldsymbol\xi_i - \boldsymbol\xi_i^*)$ term on the right-hand side of Eq.~\ref{eq:rml} acts as a regularization. The use of the CNN-PCA parameterization assures that posterior realizations are consistent (in a geostatistical sense) with prior realizations.

The minimization in Eq.~\ref{eq:rml} can be performed using various algorithms. Here we apply mesh adaptive direct search method (MADS), described in \citep{audet2006mesh}. MADS is a local-search (derivative-free) procedure that evaluates trial points determined by an underlying stencil. This stencil is centered around the current-best solution. Each iteration (in the MADS variant used here) entails $2l$ function evaluations, where $l$ denotes the number of optimization variables. In our case, $l=n_\xi=100$. MADS applies a set of strategies when no improvement is achieved with the current stencil. Convergence to a local minimum is guaranteed if particular problem criteria are satisfied. We note finally that a range of optimizers could be used for the minimization in Eq.~\ref{eq:rml}.



\subsection{Problem setup}
\label{sect:hm_setup}

The system considered here is consistent with that used in Section~\ref{sect:flowstats}. We again have a 2D channelized model defined on an $80\times 80$ grid. The model contains 25~wells (18~producers and seven injectors, 20~wells are in sand and five are in mud). All realizations are conditioned to facies-type at well locations. The `true' model, which is a (new) random SGeMS realization generated from the training image used in the construction of the CNN-PCA representation, is shown in Fig.~\ref{fig:hm-true}.

Oil-water flow is again considered, with fluid and rock-fluid properties as described in Section~\ref{sect:flowstats}. The total simulation time frame is 1000~days. The first 400~days are prescribed to be the history matching period (during which data are collected), and the following 600~days are the forecast period. Oil and water production rates for the 18~producers, and water injection rates for the seven injectors, at five time steps, comprise the observed data (there are thus $18 \times 5 \times 2 + 7 \times 5 = 215$ measurements to be matched). Random Gaussian noise, consistent with $C_D$, is added to the simulated flow response for the `true' model to provide the observed data vector. The mean and standard deviation of the random Gaussian noise are set to zero and 5\% of the corresponding true data, respectively. 


In this work we generate $N_r=100$ posterior models. In each RML run, 200 MADS iterations are performed, and at each iteration 200 function evaluations are required. Consequently, the overall history matching procedure entails $100\times200\times 200 = 4\times 10^6$ flow model evaluations. The average runtime for AD-GPRS for a single flow problem is around 10~seconds on a single CPU, while that of the deep-learning-based surrogate model is about 0.01~seconds (for batch predictions on a single GPU). Thus, in terms of total simulation time, the use of AD-GPRS for all function evaluations would correspond to $4\times 10^7$~seconds (about 1.3~years of computation) for this problem, while the use of the surrogate model entails 40,000~seconds, or about 11~hours. The surrogate model also requires 1500 (training) AD-GPRS simulations and recurrent R-U-Net training, which adds several more hours of computation. The speedup using the surrogate model is still extremely large.

It is important to point out that these timings are for data assimilation using RML and MADS, with $N_r=100$, 200 MADS iterations, and $l=n_\xi=100$. The use of ensemble-based history matching approaches will reduce these timings dramatically. Savings could also be achieved by reducing the values of $N_r$ and $l$, or by using a gradient-based optimization procedure. However, one advantage of the surrogate model in this setting is that it may enable the use of history matching algorithms that would be otherwise intractable.

\subsection{History matching results}
\label{hm_results}

\begin{figure}
\centering
 \begin{subfigure}[b]{0.4\textwidth}
     \centering
     \includegraphics[width=\textwidth]{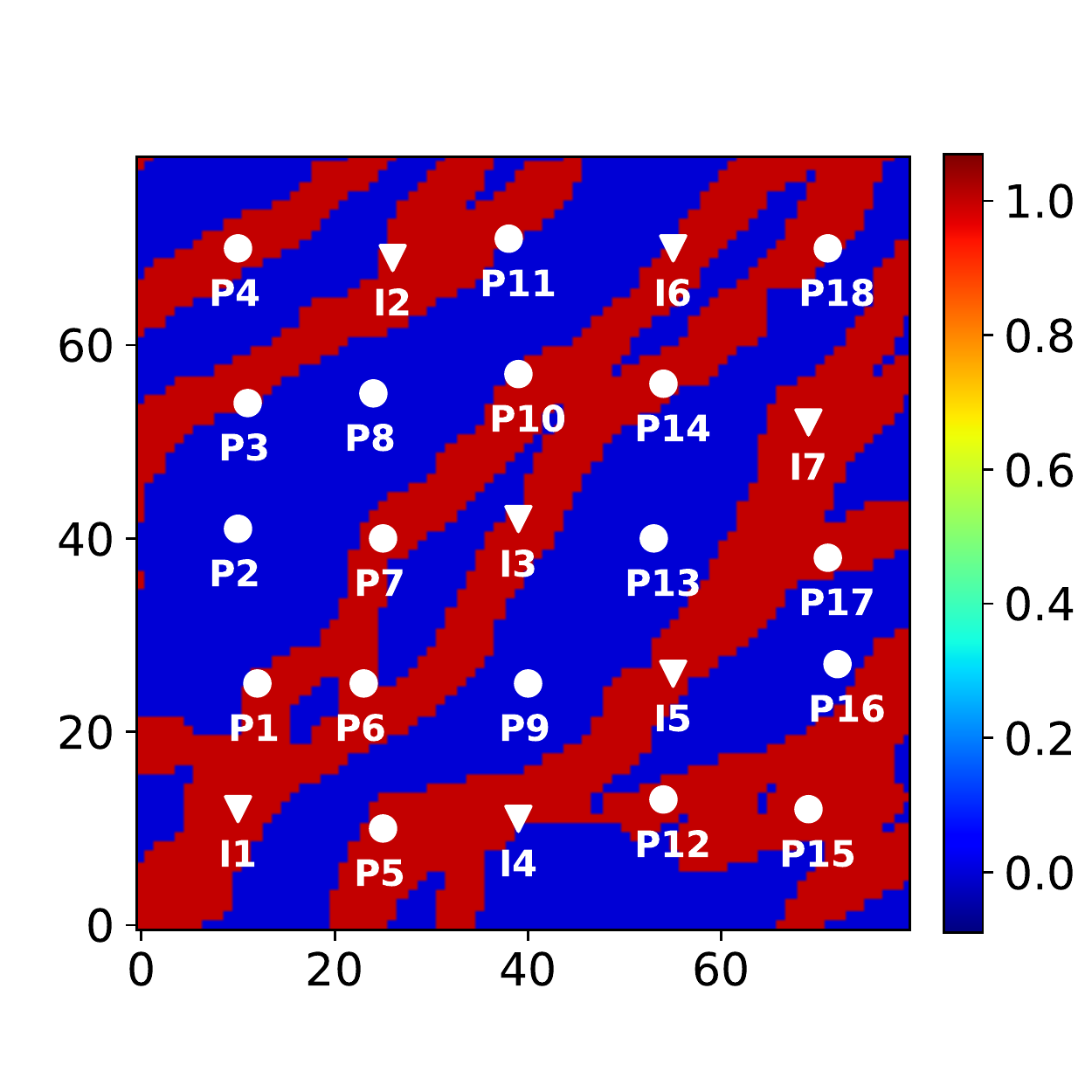}
     \caption{True model}
     \label{fig:hm-true}
 \end{subfigure}
 
     \centering
     \begin{subfigure}[b]{0.3\textwidth}
         \centering
         \includegraphics[width=\textwidth]{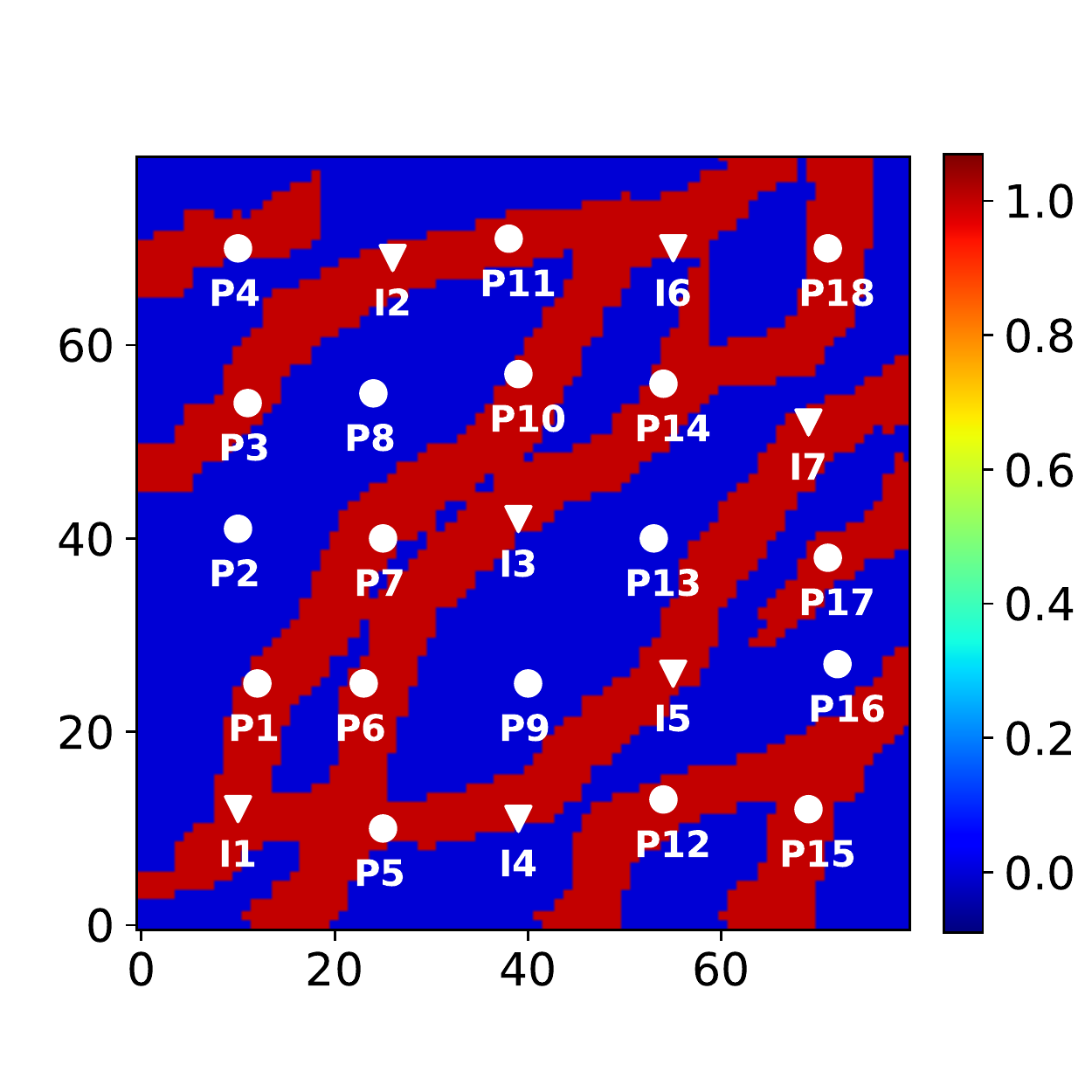}
         \caption{Prior model 1}
         \label{fig:hm-prior-1}
     \end{subfigure}
     \begin{subfigure}[b]{0.3\textwidth}
         \centering
         \includegraphics[width=\textwidth]{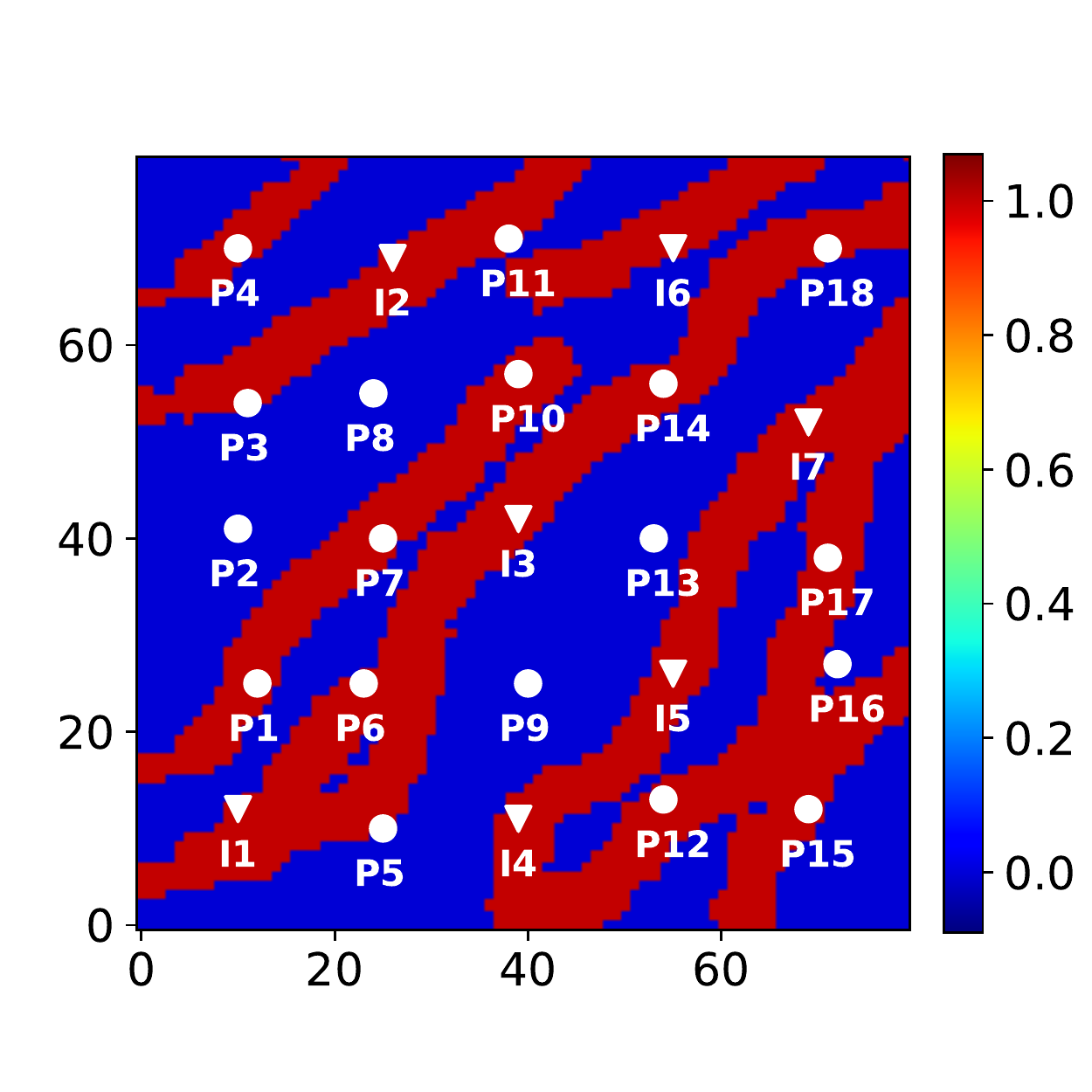}
         \caption{Prior model 2}
         \label{fig:hm-prior-2}
     \end{subfigure}
     \begin{subfigure}[b]{0.3\textwidth}
         \centering
         \includegraphics[width=\textwidth]{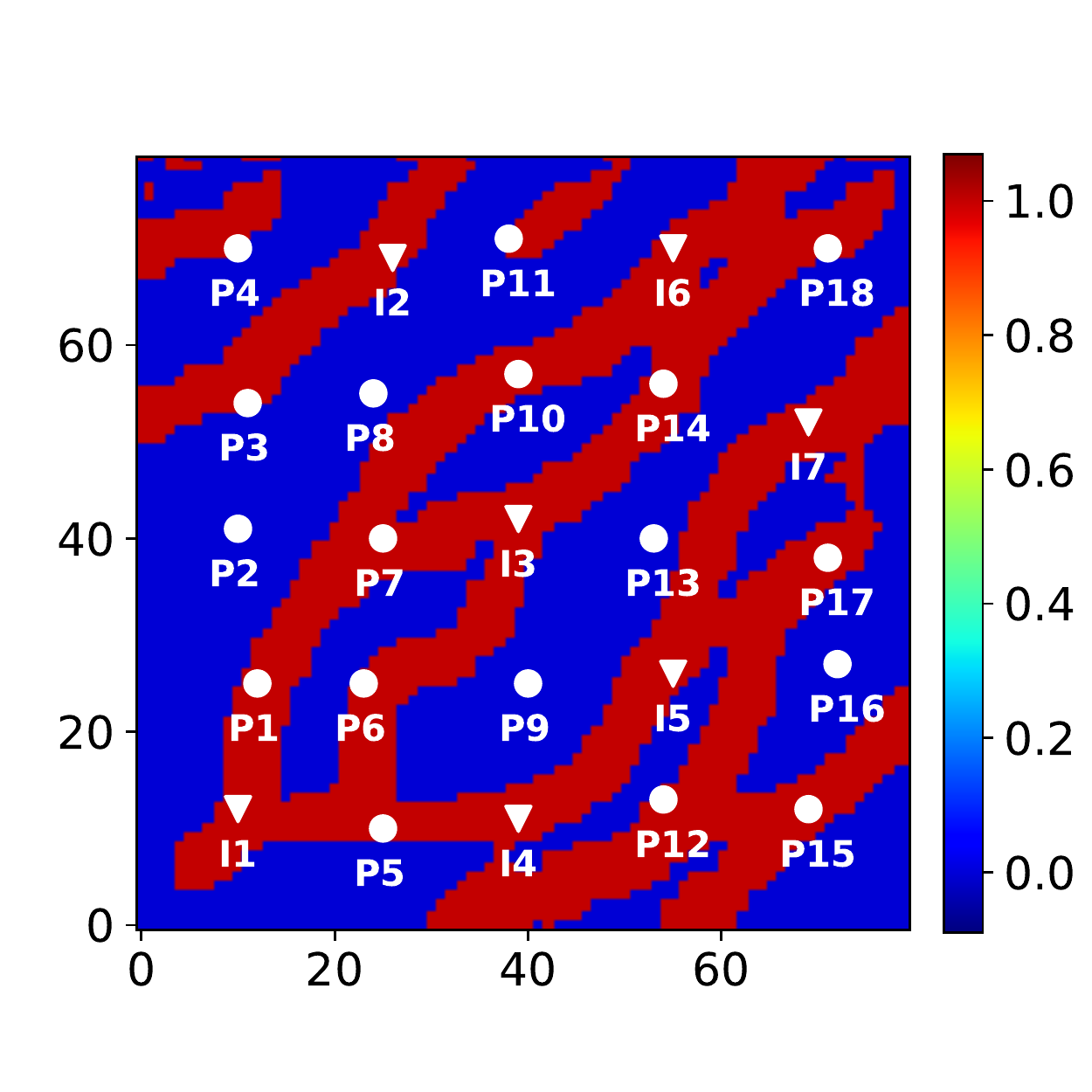}
         \caption{Prior model 3}
         \label{fig:hm-prior-3}
     \end{subfigure}

     \begin{subfigure}[b]{0.3\textwidth}
         \centering
         \includegraphics[width=\textwidth]{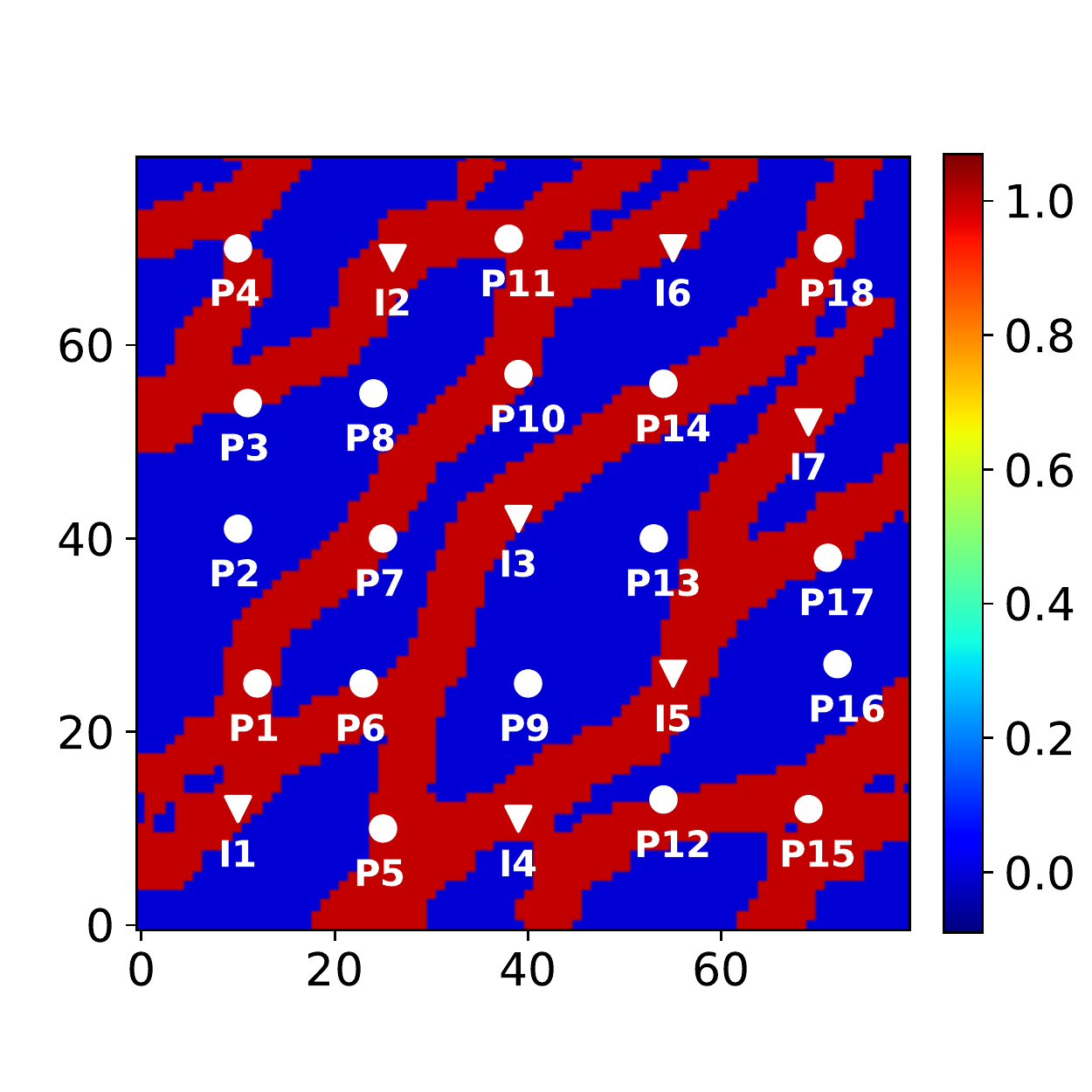}
         \caption{Posterior model 1}
         \label{fig:hm-posterior-1}
     \end{subfigure}
     \begin{subfigure}[b]{0.3\textwidth}
         \centering
         \includegraphics[width=\textwidth]{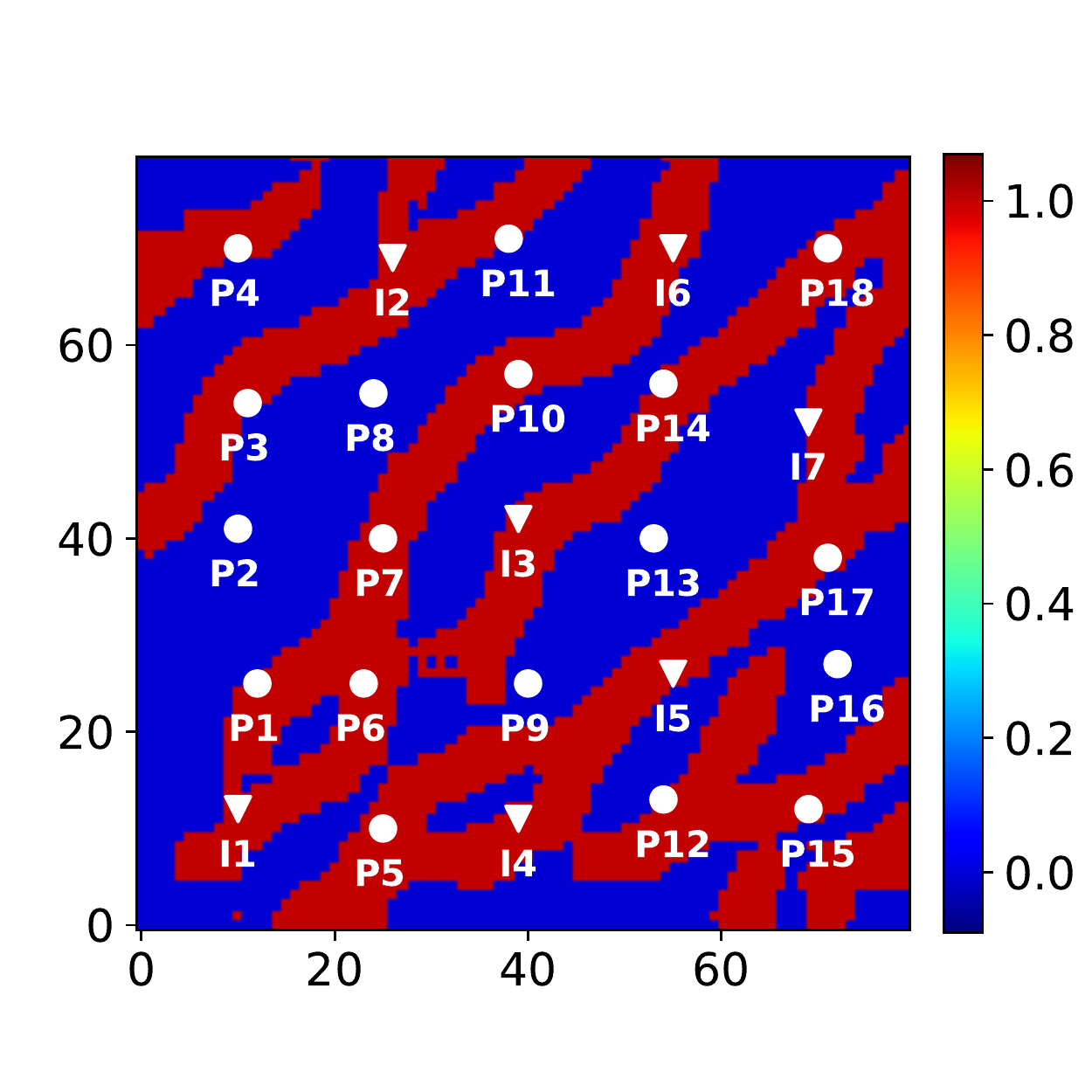}
         \caption{Posterior model 2}
         \label{fig:hm-posterior-2}
     \end{subfigure}
     \begin{subfigure}[b]{0.3\textwidth}
         \centering
         \includegraphics[width=\textwidth]{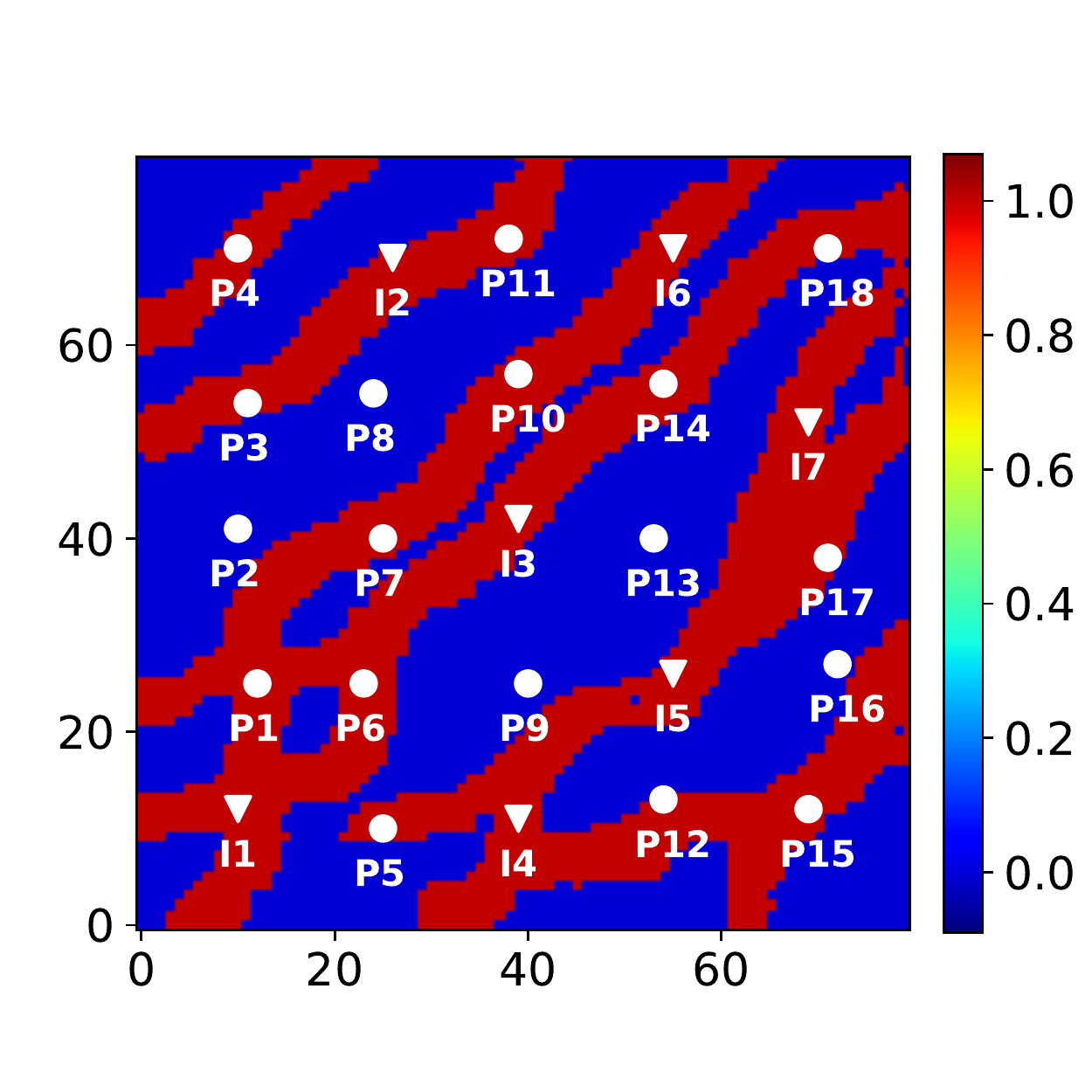}
         \caption{Posterior model 3}
         \label{fig:hm-posterior-3}
     \end{subfigure}
    \caption{Binary facies models for the 2D channelized system. All models conditioned to facies type at the 25~wells. True SGeMS model, along with three prior and (corresponding) posterior CNN-PCA models are shown.}
    \label{fig:hm-facies-models}
\end{figure}

We now present history matching results using the recurrent R-U-Net as a surrogate for the flow simulator. Three randomly selected prior models are shown in Fig.~\ref{fig:hm-facies-models}b, c and d. These models are used as initial guesses in the subspace RML procedure (Eq.~\ref{eq:rml}), and the corresponding posterior models appear in Fig.~\ref{fig:hm-facies-models}e, f and g. Both the prior and posterior models clearly resemble the true model (Fig.~\ref{fig:hm-facies-models}a) in terms of their general channelized appearance. This is because CNN-PCA is able to maintain the channelized structure apparent in the original training image.

There are, however, important differences between the prior and posterior models in terms of how particular sets of wells are connected (or not connected) through sand. For example, in the true model, injector~I1 and producer~P5 (both are in the lower left portion of the model) are not connected through sand, while in all three prior models they are connected. In the three posterior models, wells~I1 and P5 are (correctly) not connected via sand. In addition, wells~I2 and P11 are connected through sand in the true model, while in prior model~3 (Fig.~\ref{fig:hm-facies-models}d) they are not connected. In the corresponding posterior model (Fig.~\ref{fig:hm-facies-models}g) they are properly connected. Sand connections are also established between wells~I6 and P10, and between wells~I4 and P5, in posterior model~2 (compare Fig.~\ref{fig:hm-facies-models}c and f).

\begin{figure}
     \centering
     \begin{subfigure}[b]{0.45\textwidth}
         \centering
         \includegraphics[width=\textwidth]{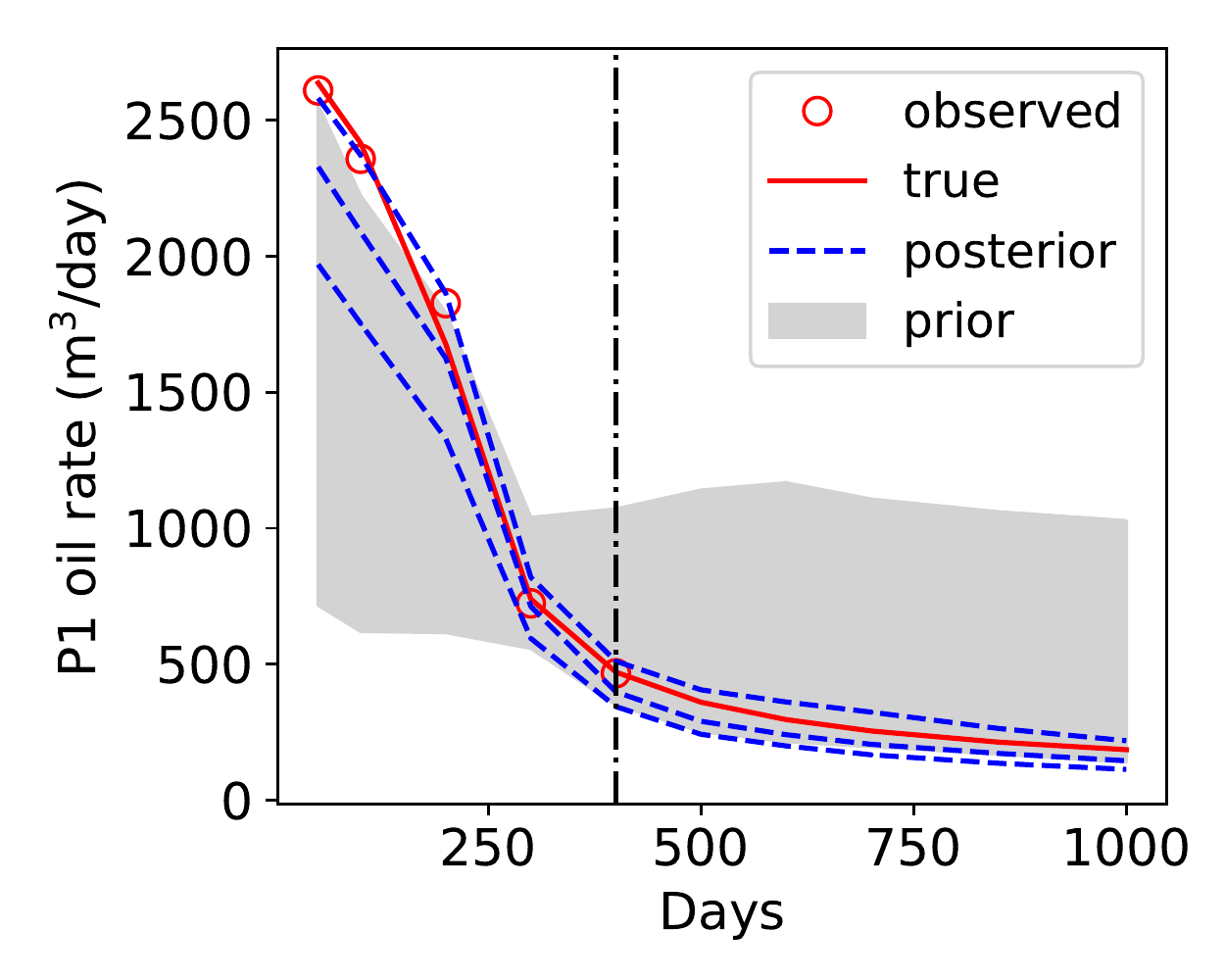}
         \caption{P1 oil rate}
         \label{hm-orate-w1}
     \end{subfigure}
     \begin{subfigure}[b]{0.45\textwidth}
         \centering
         \includegraphics[width=\textwidth]{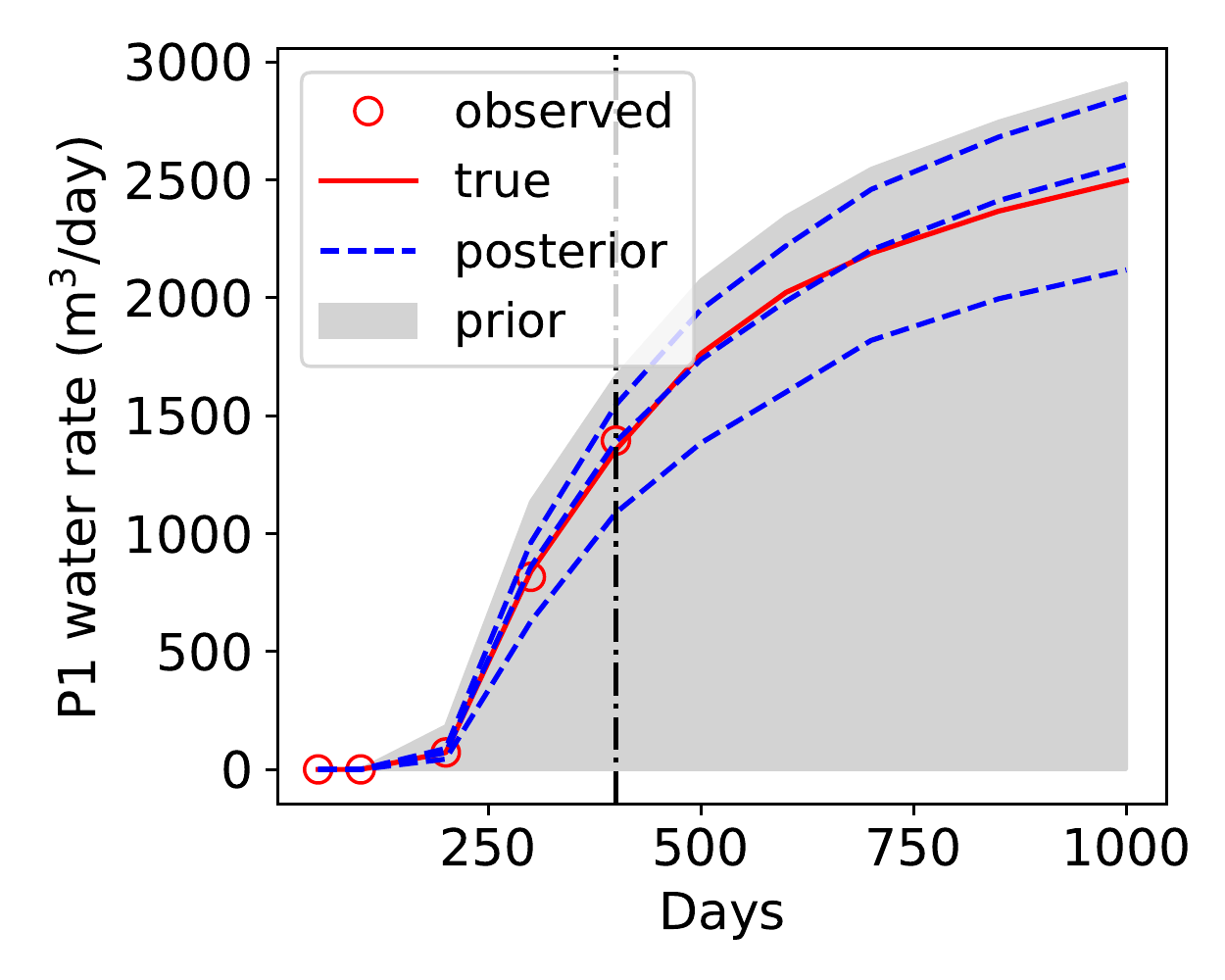}
         \caption{P1 water rate}
         \label{hm-wrate-w1}
     \end{subfigure}
     
     \begin{subfigure}[b]{0.45\textwidth}
         \centering
         \includegraphics[width=\textwidth]{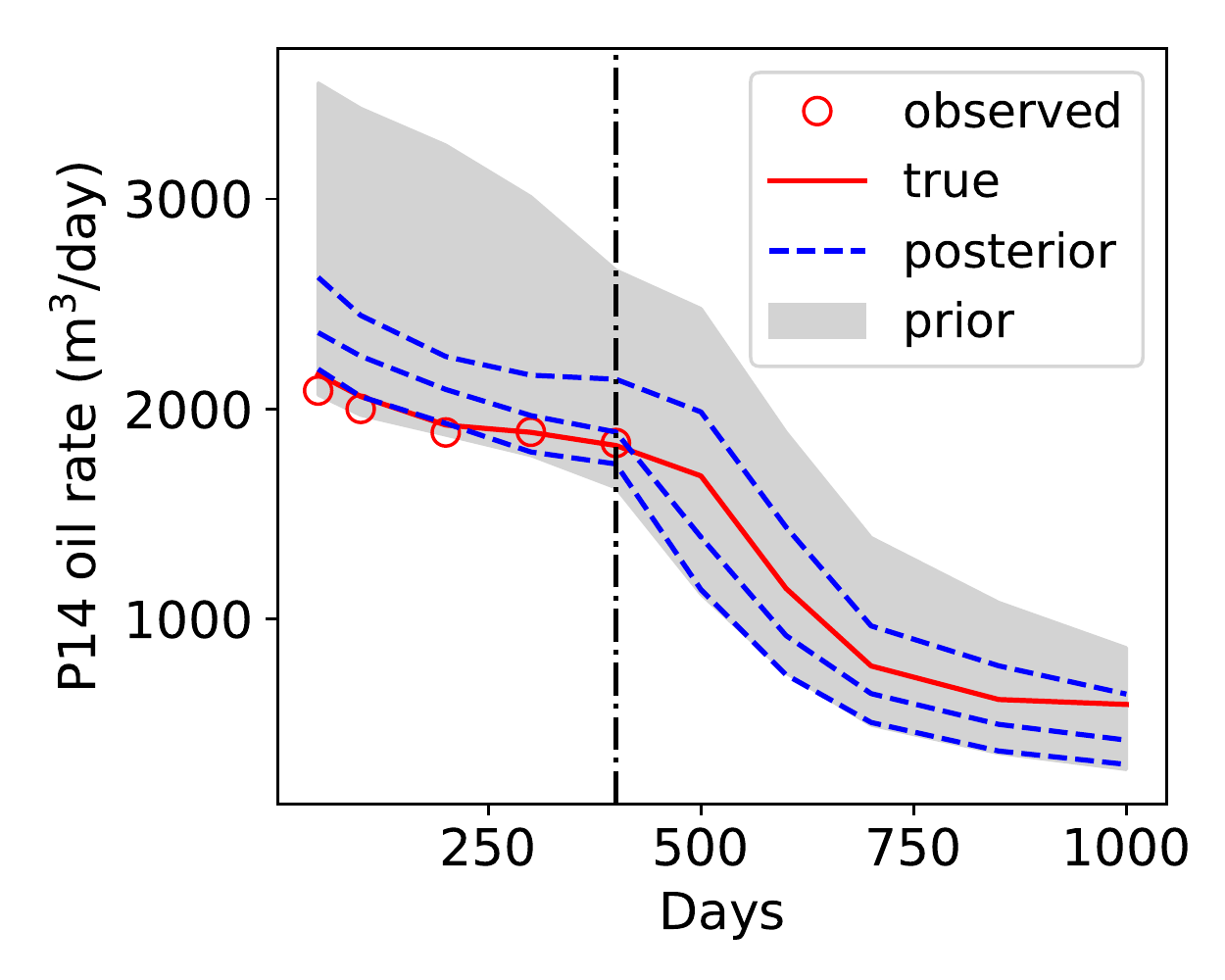}
         \caption{P14 oil rate}
         \label{hm-orate-w14}
     \end{subfigure}
     \begin{subfigure}[b]{0.45\textwidth}
         \centering
         \includegraphics[width=\textwidth]{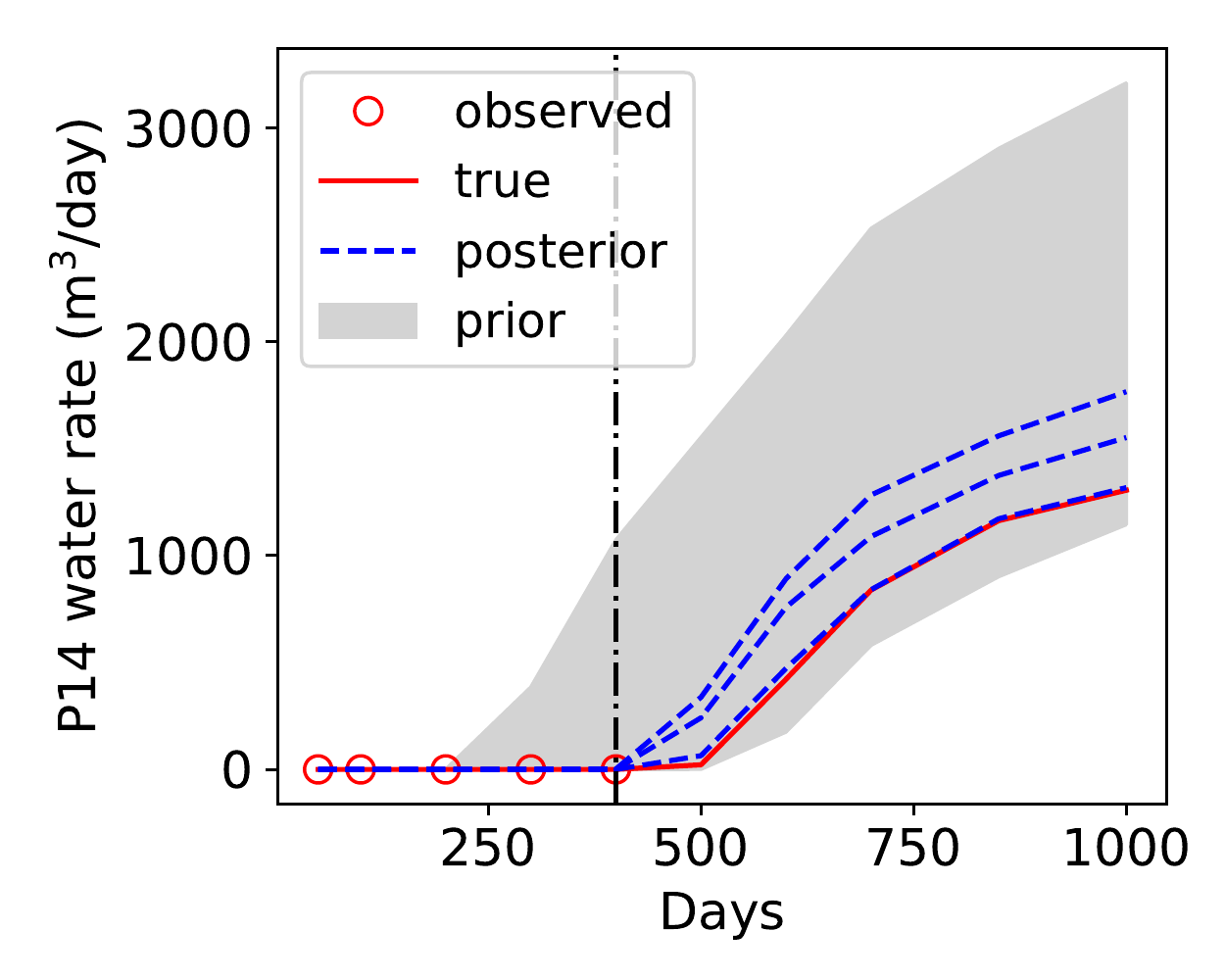}
         \caption{P14 water rate}
         \label{hm-wrate-w14}
     \end{subfigure}
     
    \begin{subfigure}[b]{0.45\textwidth}
         \centering
         \includegraphics[width=\textwidth]{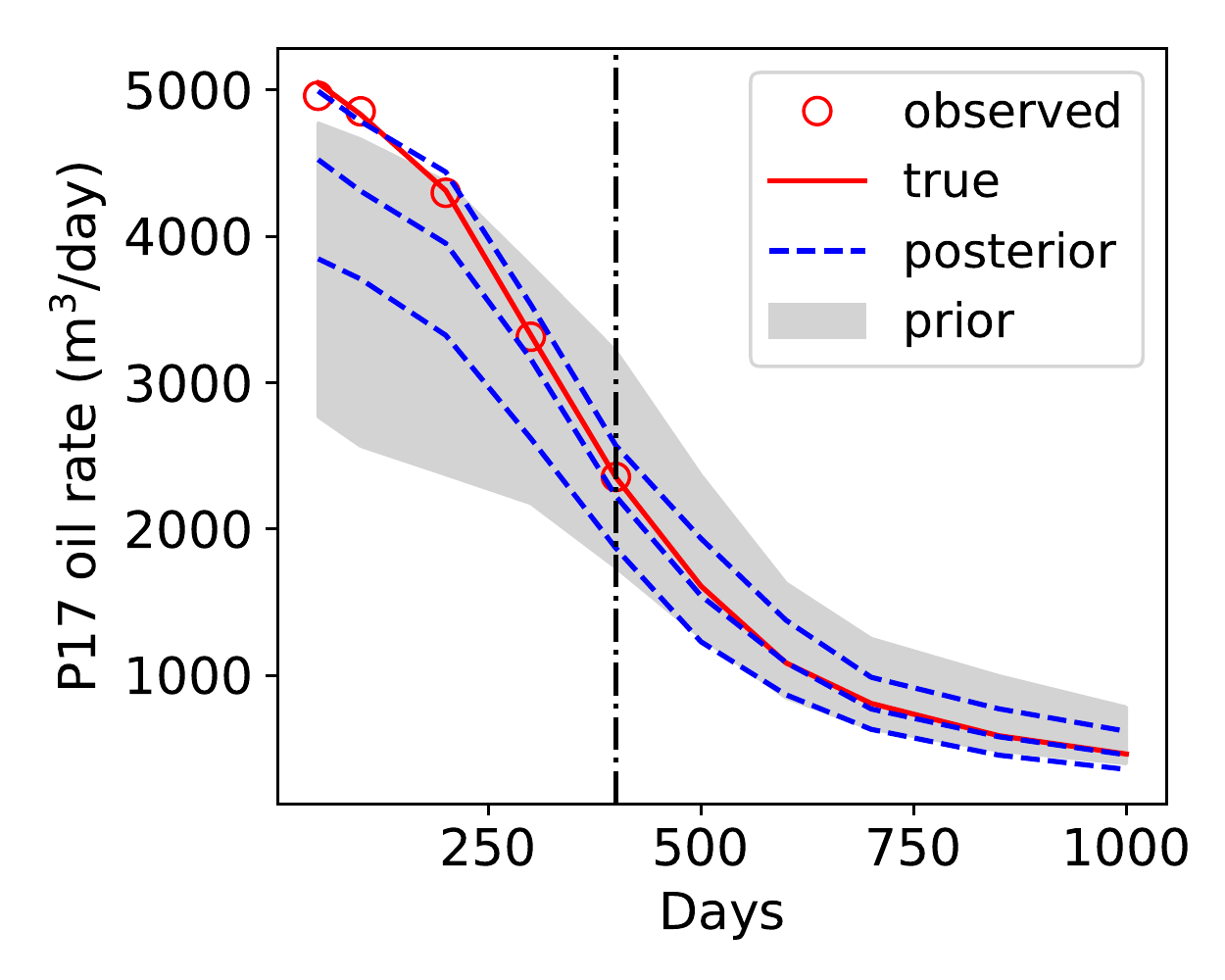}
         \caption{P17 oil rate}
         \label{hm-orate-w17}
     \end{subfigure}
          \begin{subfigure}[b]{0.45\textwidth}
         \centering
         \includegraphics[width=\textwidth]{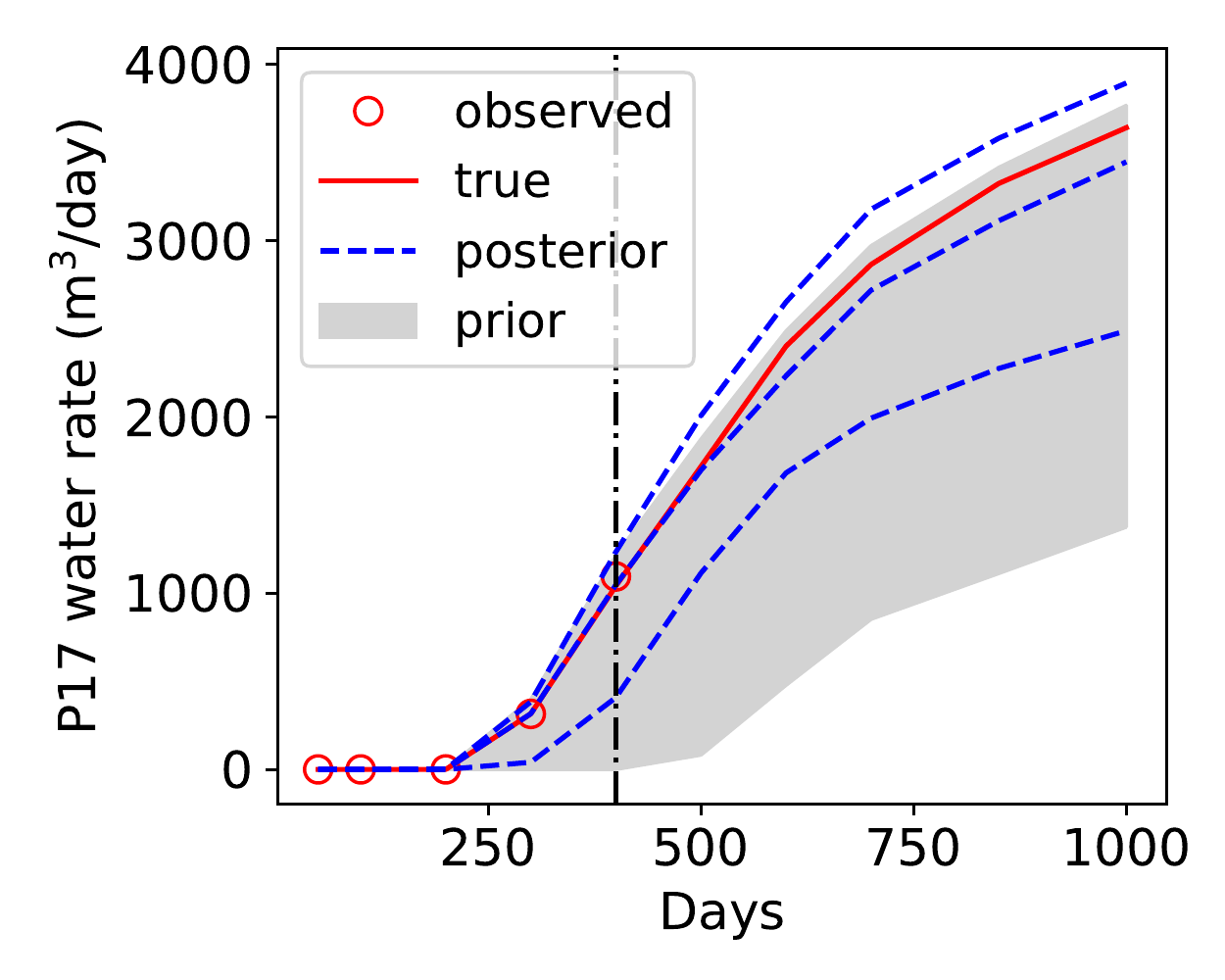}
         \caption{P17 water rate}
         \label{hm-wrate-w17}
     \end{subfigure}
    
    \caption{Oil (left) and water (right) production for producers P1, P14 and P17. Gray regions represent the prior P10--P90 range, red points and red lines denote observed and true data, and blue dashed curves denote the posterior P10 (lower), P50 (middle) and P90 (upper) predictions. Vertical dashed line divides simulation time frame into history match and prediction periods.}
    \label{fig:hm-well-flow}
\end{figure}

Fig.~\ref{fig:hm-well-flow} displays results for oil and water production rates for producers P1, P14 and P17. In these figures, the gray region displays the P10--P90 interval for the prior models, the red curve denotes the true model flow response, the red circles indicate the observed data (generated by perturbing from the true model response, as described in Section~\ref{sect:hm_setup}), and the blue dashed curves depict the P10, P50 and P90 posterior results generated using RML with the recurrent R-U-Net surrogate model. The vertical black dashed line at 400~days divides the simulation time frame into the history matching period (to the left of the line) and forecast or prediction period.

It is immediately evident that the history matching procedure leads to a significant reduction in uncertainty; i.e., the posterior P10--P90 ranges are clearly smaller than the prior P10--P90 ranges. The posterior P10--P90 interval generally captures the observed (and true) data, even when these data fall toward the edge of the prior P10--P90 interval. This is evident in all three oil rate plots in Fig.~\ref{fig:hm-well-flow}. It is also interesting to note that, even when water breakthrough has not yet occurred in a particular well during the history match period (e.g., well~P14 in Fig.~\ref{fig:hm-well-flow}d), there is still significant uncertainty reduction in the water rate prediction.

\begin{figure}
     \centering
     \begin{subfigure}[b]{0.45\textwidth}
         \centering
         \includegraphics[width=\textwidth]{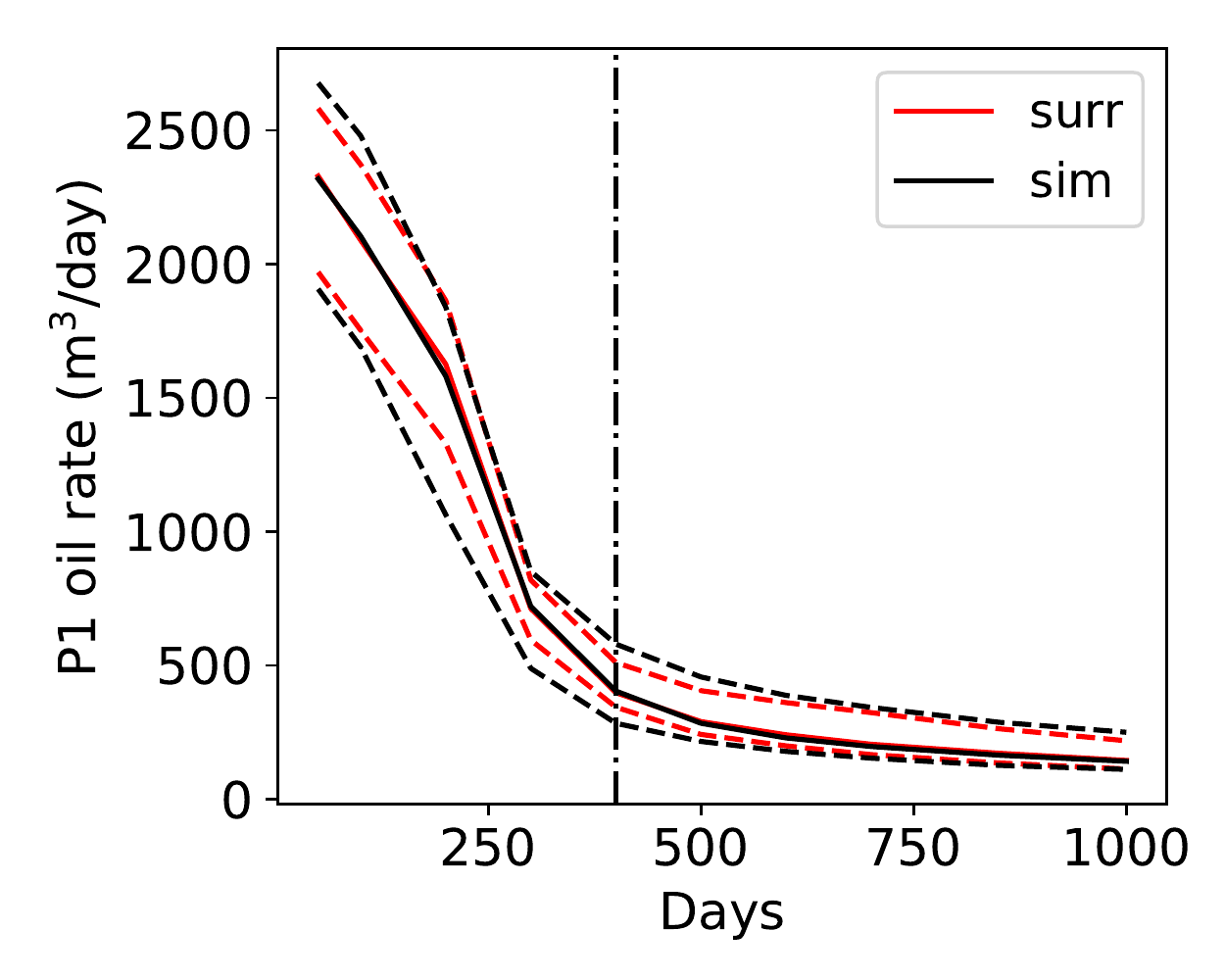}
         \caption{P1 oil rate}
         \label{hm-val-orate-w1}
     \end{subfigure}
     \begin{subfigure}[b]{0.45\textwidth}
         \centering
         \includegraphics[width=\textwidth]{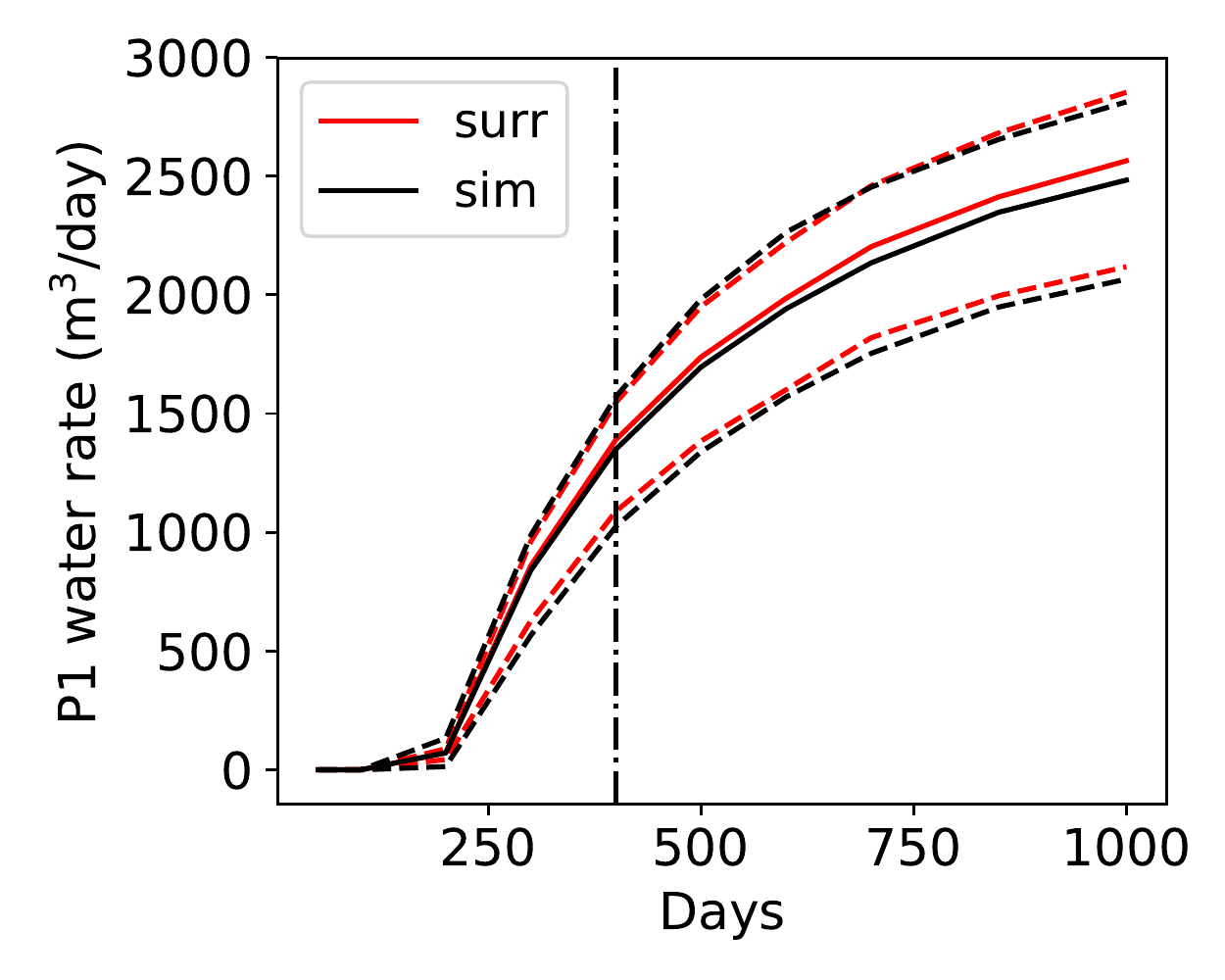}
         \caption{P1 water rate}
         \label{hm-val-wrate-w1}
     \end{subfigure}
     
     \begin{subfigure}[b]{0.45\textwidth}
         \centering
         \includegraphics[width=\textwidth]{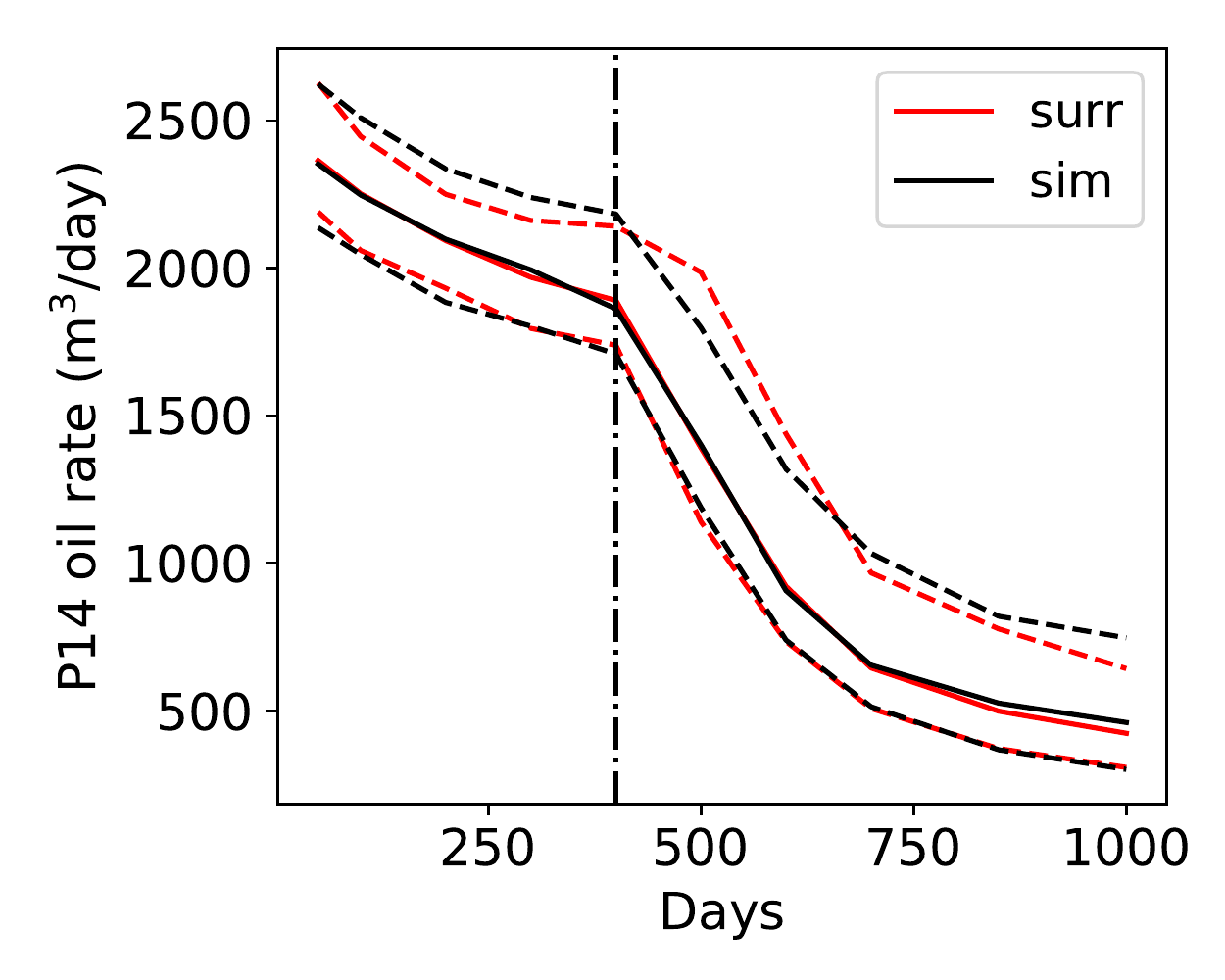}
         \caption{P14 oil rate}
         \label{hm-val-orate-w14}
     \end{subfigure}
     \begin{subfigure}[b]{0.45\textwidth}
         \centering
         \includegraphics[width=\textwidth]{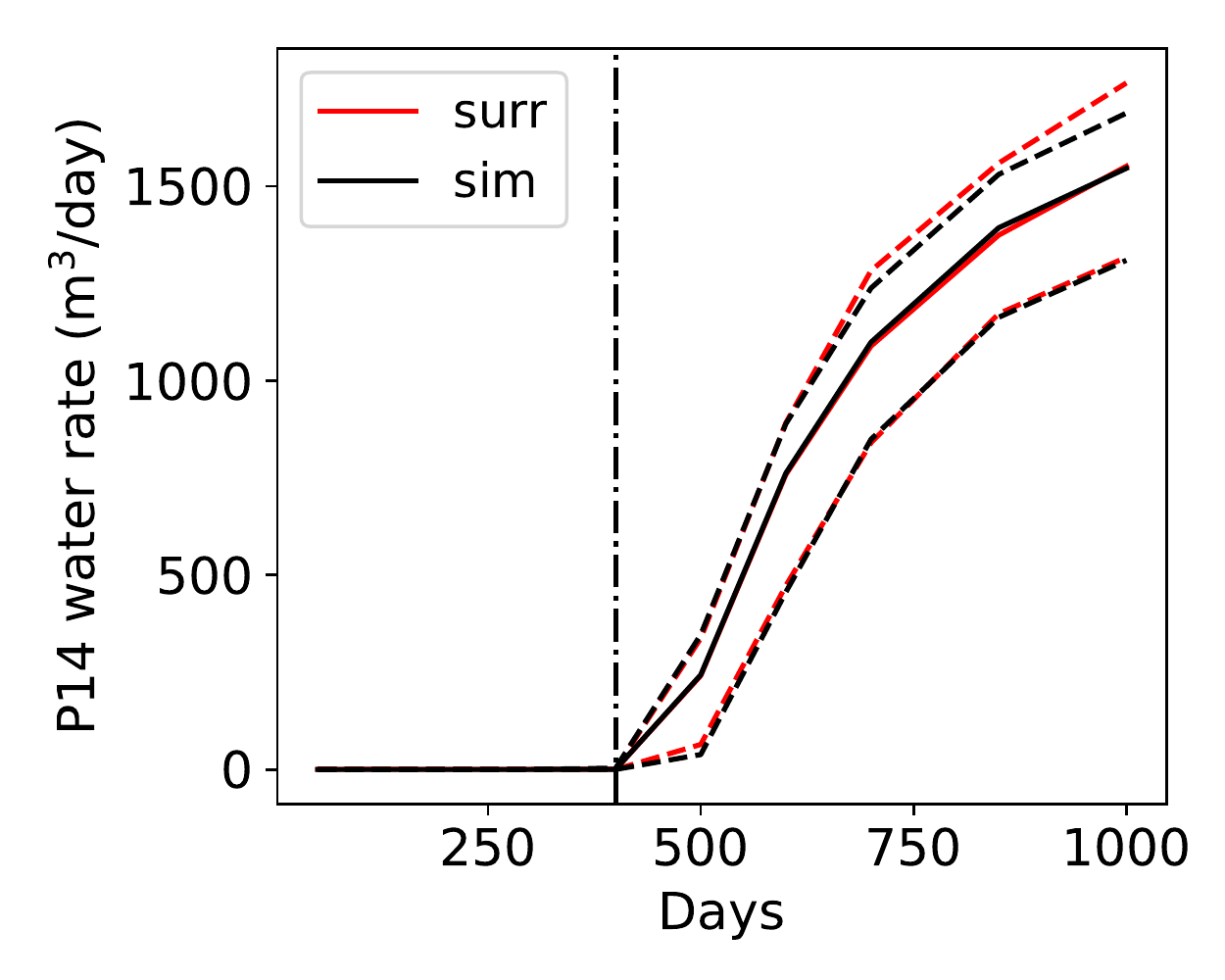}
         \caption{P14 water rate}
         \label{hm-val-wrate-w14}
     \end{subfigure}
     
    \begin{subfigure}[b]{0.45\textwidth}
         \centering
         \includegraphics[width=\textwidth]{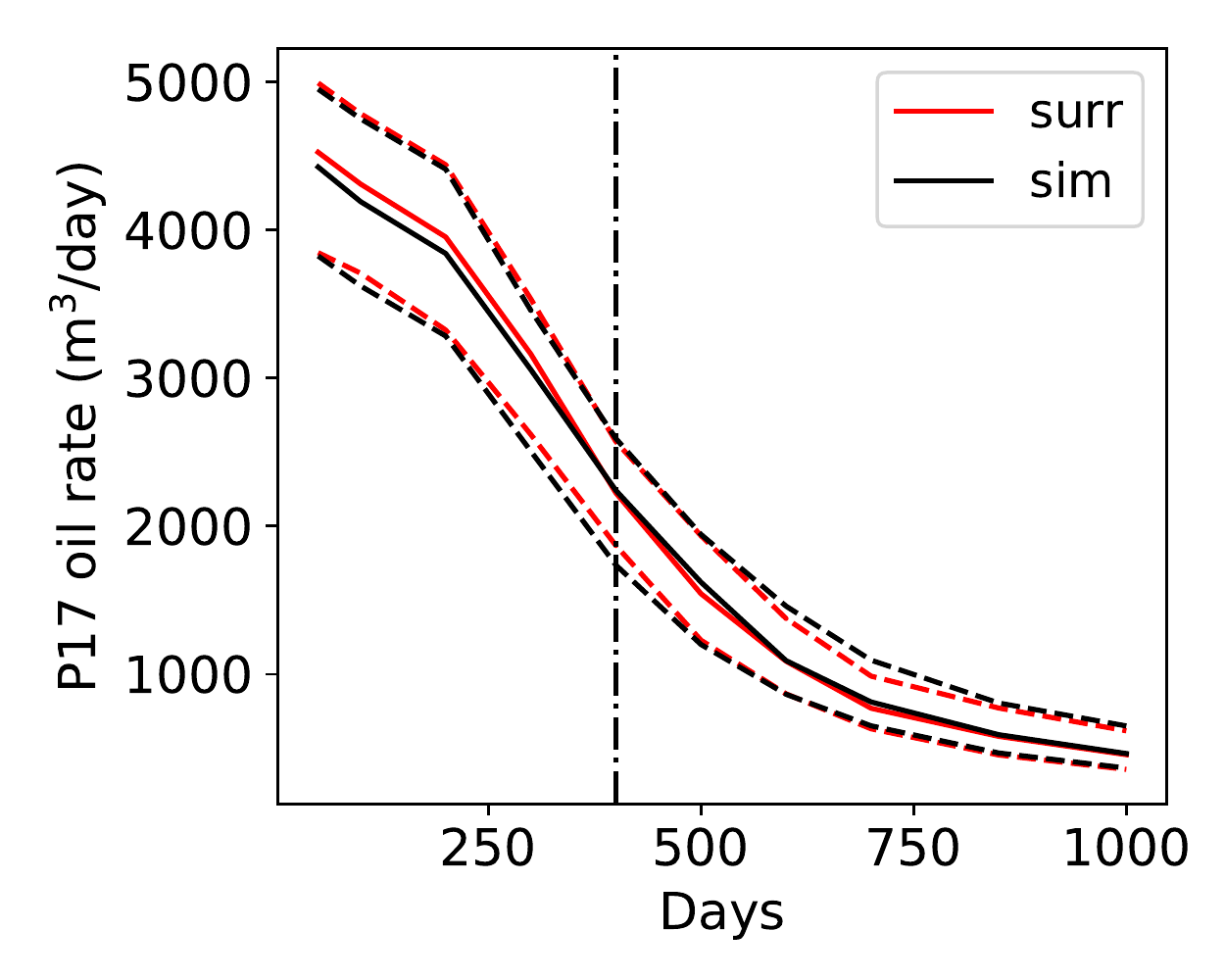}
         \caption{P17 oil rate}
         \label{hm-val-orate-w17}
     \end{subfigure}
          \begin{subfigure}[b]{0.45\textwidth}
         \centering
         \includegraphics[width=\textwidth]{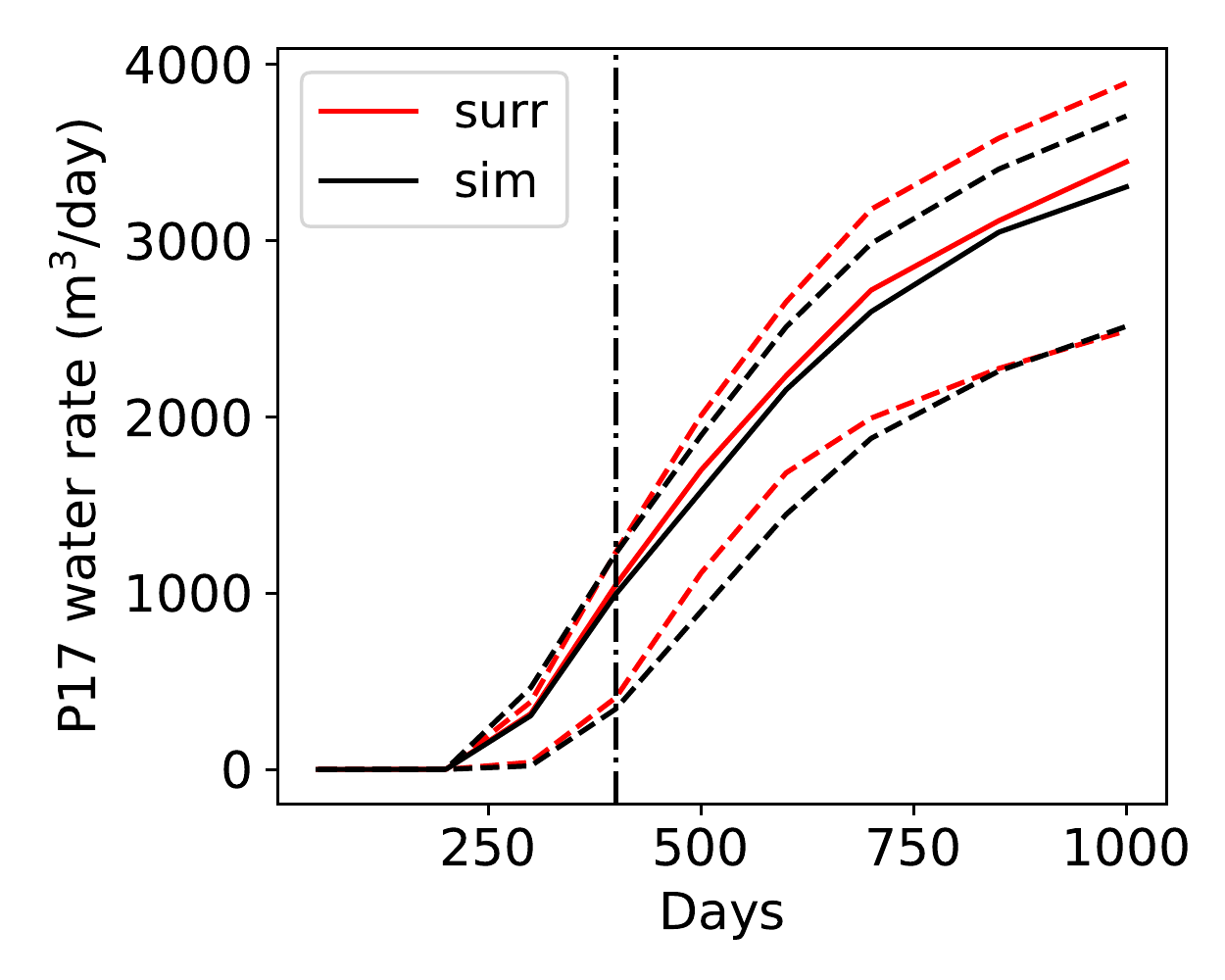}
         \caption{P17 water rate}
         \label{hm-val-wrate-w17}
     \end{subfigure}
    
   \caption{Comparison of oil (left) and water (right) posterior production forecasts generated by AD-GPRS numerical simulator (black curves) and recurrent R-U-Net surrogate model (red curves). Solid curves correspond to P50 response, and dashed curves to P10 (lower curves) and P90 (upper curves) responses.}
    \label{fig:hm-val-well-flow}
\end{figure}

The inversion procedure provides 100 posterior geomodels, and all of the posterior flow responses for these models (in Fig.~\ref{fig:hm-well-flow}) were generated using the recurrent R-U-Net surrogate model. It is therefore reasonable to verify the predicted flow responses for these models by simulating them using the numerical simulator. The results of such an assessment are shown in Fig.~\ref{fig:hm-val-well-flow}. The black and red solid curves denote the P50 flow response using AD-GPRS and the surrogate model, respectively. The black and red dashed curves depict the P10 and P90 flow responses from AD-GPRS and the surrogate model. The red (surrogate model) curves in Fig.~\ref{fig:hm-val-well-flow} are the same as the blue curves in Fig.~\ref{fig:hm-well-flow}, but the scales in some plots differ because the prior P10--P90 ranges are not shown in Fig.~\ref{fig:hm-val-well-flow}.

From Fig.~\ref{fig:hm-val-well-flow}, we see that there is generally very close correspondence between the AD-GPRS and recurrent R-U-Net P10, P50, P90 posterior results. Agreement in the P50 curves is consistently close, though small discrepancies are observed in some of the P10 and P90 curves (e.g., P90 curve in Fig.~\ref{fig:hm-val-well-flow}c and P10 and P90 curves in Fig.~\ref{fig:hm-val-well-flow}f). These errors are, however, very small compared to the amount of uncertainty reduction achieved by the inversion procedure.

\section{Concluding Remarks}
\label{sect:concl}

In this work, a deep-learning-based surrogate model was developed to capture temporal dynamics in a high-dimensional nonlinear system, specifically an oil-water subsurface flow model. The surrogate model entails a residual U-Net architecture linked to a convolutional long short term memory recurrent network. 
We refer to the overall network as a recurrent R-U-Net. The recurrent R-U-Net is trained on numerically simulated dynamic saturation and pressure maps, generated for different geological realizations drawn from a particular geological scenario. In our case 2D channelized models were considered, and 1500 training simulations were performed. The recurrent R-U-Net was trained to predict dynamic saturation and pressure in every grid block at 10 particular time steps. Additional weighting at well blocks was introduced in the loss function in order to accurately capture well flow responses, since these represent the primary data used in history matching. 

The recurrent R-U-Net was evaluated for oil-water reservoir simulation problems involving flow through new (test) channelized geological realizations. The ability to predict flow responses for new geomodels is essential if the recurrent R-U-Net is to be used for history matching. Detailed comparison of the dynamic pressure and saturation maps generated by the surrogate model and the reference numerical simulator, for a particular geomodel, demonstrated that the recurrent R-U-Net can accurately predict these high-dimensional state variables. Well flow responses -- specifically time-varying oil and water production rates and water injection rates -- were also shown to be in close agreement. The surrogate model was then evaluated for a test ensemble of 500 new geomodels. In this assessment the P10, P50 and P90 flow responses from the recurrent R-U-Net were compared to those from the numerical simulator. The high level of accuracy of the surrogate model in predicting these flow statistics demonstrates its applicability for uncertainty quantification. 

The recurrent R-U-Net was next applied for history matching. This is a challenging application area, particularly when channelized (rather than multi-Gaussian) geomodels are considered. Posterior model generation was accomplished using a randomized maximum likelihood (RML) procedure, and geomodels were represented concisely (in terms of 100 parameters) using the recently developed CNN-PCA parameterization. RML is an optimization-based method, and the minimization was accomplished using mesh adaptive direct search. Significant uncertainty reduction was achieved, and the posterior (surrogate-model) predictions for oil and water production rates were shown to be reasonably accurate through comparison to numerical simulation results (for the posterior models). The speedup obtained using the surrogate model relative to high-fidelity numerical simulation was dramatic in this example, suggesting that the use of the recurrent R-U-Net may enable the application of more rigorous inverse modeling procedures for realistic problems. This will be considered in future work.

There are many other promising directions for future research in this general area. The surrogate model developed in this work is for 2D problems, and we plan to investigate extensions of the recurrent R-U-Net to 3D systems. The surrogate model can also be extended to treat larger and more complicated systems. Our specific interest is in multiphysics problems involving coupled flow and geomechanics, which entail additional governing equations and output state variables such as surface deformations. Surrogate models can have a very large impact for such problems since high-fidelity numerical simulations are often extremely expensive. Finally, the  deep-learning-based surrogate model can potentially be extended to handle varying well control and well location variables. This could be accomplished by encoding these variables as additional input maps during training. If successful, this capability could enable the surrogate model to be used for field development optimization in addition to history matching.

\bigskip
\noindent\textbf{Acknowledgements} We are grateful to the Stanford Smart Fields Consortium and to Stanford--Chevron CoRE for partial funding of this work. We also thank the Stanford Center for Computational Earth \& Environmental Science (CEES) for providing the computational resources used in this study.

\section*{Appendix: Recurrent R-U-Net Architecture}
The detailed architecture of the recurrent R-U-Net is shown in Table~\ref{table:r-u-net architecture}. In the table, `Conv' represents a convolutional layer followed by batch normalization and ReLU nonlinear activation, while `Transposed conv' denotes a transposed (upsampling) convolutional layer followed by batch normalization and ReLU. A stack of two convolutional layers with 128 filters of size $3\times3\times128$ constitute a `Residual block,' in which the first convolutional layer has skip connections with the output of the second convolutional layer. The `ConvLSTM2D block,' which also employs 128 filters of size $3\times3\times128$, performs all of the LSTM gate operations. Note that the convLSTM net generates $(N_x/4, N_y/4, 128)$ activation maps for all $n_t$ time steps. The decoder layers process these $n_t$ activation maps separately to produce the state maps.

\begin{table}
\begin{center}
  \begin{tabular}{ c | c | c  }
    \hline
    Net &  Layer  & Output size\\ 
     \hline
    \multirow{8}{4em}{Encoder}& Input  & $(N_x, N_y, 2)$ \\
    & Conv, 16 filters of size $3\times3\times2$, stride 2  & $(N_x/2, N_y/2, 16)$ \\
    & Conv, 32 filters of size $3\times3\times16$, stride 1  & $(N_x/2, N_y/2, 32)$ \\
    &Conv, 64 filters of size $3\times3\times32$, stride 2  & $(N_x/4, N_y/4, 64)$ \\
    &Conv, 128 filters of size $3\times3\times64$, stride 1  & $(N_x/4, N_y/4, 128)$ \\
    &Residual block, 128 filters & $(N_x/4, N_y/4, 128)$ \\
    &Residual block, 128 filters & $(N_x/4, N_y/4, 128)$ \\
    &Residual block, 128 filters & $(N_x/4, N_y/4, 128)$ \\ \hline
    ConvLSTM&ConvLSTM2D block, 128 filters & $(N_x/4, N_y/4, 128, N_t)$ \\ \hline
    \multirow{8}{4em}{Decoder}&Residual block, 128 filters & $(N_x/4, N_y/4, 128, N_t)$ \\
    &Residual block, 128 filters & $(N_x/4, N_y/4, 128, N_t)$ \\
    &Residual block, 128 filters & $(N_x/4, N_y/4, 128, N_t)$ \\
    &Transposed conv, 128 filters of size $3\times3\times128$, stride 1  & $(N_x/4, N_y/4, 128, N_t)$ \\
    &Transposed conv, 64 filters of size $3\times3\times128$, stride 2  & $(N_x/2, N_y/2, 64, N_t)$ \\
    &Transposed conv, 32 filters of size $3\times3\times64$, stride 1  & $(N_x/2, N_y/2, 32, N_t)$ \\
    &Tranposed conv, 16 filters of size $3\times3\times32$, stride 2  & $(N_x, N_y, 16, N_t)$ \\
    &Conv, 1 filter of size $3\times3\times16$, stride 1  & $(N_x, N_y, 1, N_t)$ \\
   
    \hline
  \end{tabular}
  \caption{Recurrent R-U-Net architecture}
  \label{table:r-u-net architecture}
\end{center}
\end{table}

\newpage

\bibliographystyle{elsarticle-num-names}
\bibliography{reference}

\end{document}